\documentclass[conference]{IEEEtran}
\usepackage{times}

\usepackage[numbers]{natbib}
\usepackage{multicol}

\usepackage[table,xcdraw,dvipsnames,usenames]{xcolor}
\usepackage{hyperref}
\hypersetup{
    colorlinks=true,
    linkcolor=MidnightBlue,
    filecolor=magenta,      
    urlcolor=MidnightBlue,
    citecolor=MidnightBlue,
}
\usepackage{url}
\usepackage[capitalise, nameinlink]{cleveref}
\usepackage[subtle,title]{savetrees}  
\usepackage{comment}

\let\labelindent\relax
\usepackage{enumitem}

\usepackage{subcaption}
\usepackage{inconsolata}
\usepackage[T1]{fontenc}
\usepackage{float}
\usepackage{bbm}
\usepackage{graphicx}
\usepackage{booktabs}
\usepackage{balance}

\usepackage[breakable]{tcolorbox}
\usepackage{listings}

\definecolor{prompt-gray}{HTML}{a7a7a7}
\definecolor{comment-green}{rgb}{0.435, 0.576, 0.106}
\definecolor{code-syntax}{HTML}{0060b1}
\definecolor{code-constant}{HTML}{d86001}
\definecolor{code-highlight}{HTML}{e3eeff}
\definecolor{code-function}{HTML}{693da8}
\definecolor{border-gray}{rgb}{0.8, 0.8, 0.8}
\definecolor{plot-gray}{HTML}{999999}
\definecolor{plot-blue}{HTML}{1f76b4}
\definecolor{plot-purple}{HTML}{9467bd}
\definecolor{prompt-purple}{HTML}{674ea7}
\definecolor{prompt-orange}{HTML}{e69138}
\definecolor{prompt-green}{HTML}{6aa84f}
\definecolor{prompt-blue}{HTML}{4a86e8}

\renewcommand\fbox{\fcolorbox{border-gray}{white}}
\newcommand{\codecomment}[1]{\textcolor{comment-green}{#1}}

\newcommand{\codetab}{\hspace*{4mm}}

\newcommand{\ie}{i.e., }
\newcommand{\eg}{e.g., }

\begin{document}

\title{\normalfont\Huge{Learning to Learn Faster from Human Feedback\\
\vspace{0.2em}
\Huge{with Language Model Predictive Control}}
\vspace{0.1em}}

\author{
\small
Jacky Liang$^*$, Fei Xia$^*$, Wenhao Yu$^*$, Andy Zeng$^*$\\
\small Montserrat Gonzalez Arenas, Maria Attarian, Maria Bauza, Matthew Bennice, Alex Bewley, Adil Dostmohamed, Chuyuan Kelly Fu\\
\small Nimrod Gileadi, Marissa Giustina, Keerthana Gopalakrishnan, Leonard Hasenclever, Jan Humplik, Jasmine Hsu, Nikhil Joshi,\\
\small Ben Jyenis, Chase Kew, Sean Kirmani, Tsang-Wei Edward Lee, Kuang-Huei Lee, Assaf Hurwitz Michaely, Joss Moore, Ken Oslund\\
\small Dushyant Rao, Allen Ren, Baruch Tabanpour, Quan Vuong, Ayzaan Wahid, Ted Xiao, Ying Xu, Vincent Zhuang\\
\small Peng Xu$^\dagger$, Erik Frey$^\dagger$, Ken Caluwaerts$^\dagger$,Tingnan Zhang$^\dagger$, Brian Ichter$^\dagger$, Jonathan Tompson$^\dagger$, Leila Takayama$^\dagger$, Vincent Vanhoucke$^\dagger$\\
\small Izhak Shafran$^\dagger$, Maja Mataric$^\dagger$, Dorsa Sadigh$^\dagger$, Nicolas Heess$^\dagger$, Kanishka Rao$^\dagger$, Nik Stewart$^\dagger$, Jie Tan$^\dagger$, Carolina Parada$^\dagger$\\
{\scriptsize $^*$corresponding authors in alphabetical order, $^\dagger$advising leads, all other authors in alphabetical order}
}



%

\makeatletter
\let\@oldmaketitle\@maketitle%
\renewcommand{\@maketitle}{\@oldmaketitle%
    \centering
    \vspace{-1em}
    \captionsetup{type=figure}
    \includegraphics[width=0.99\linewidth]{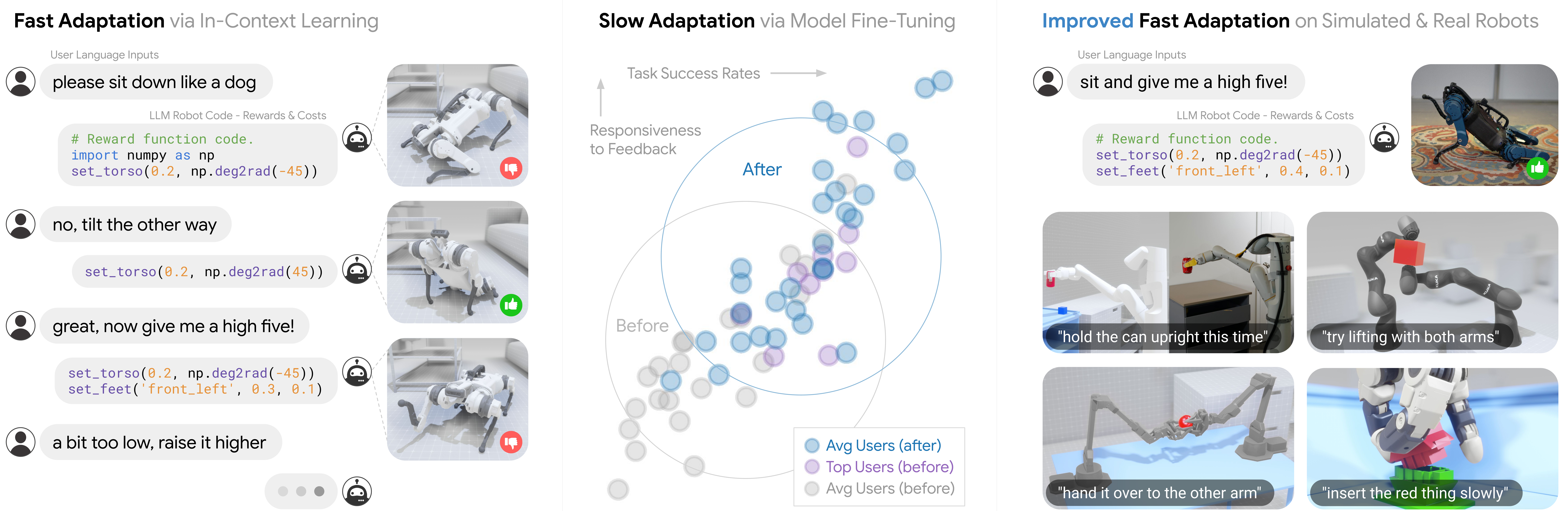}
    \captionof{figure}{
    Code-writing large language models (LLMs) present opportunities for non-experts to teach robots new tasks with language -- enabled by fast adaptation via in-context learning (left). In this work, we fine-tune the underlying LLMs to further accelerate fast adaptation and improve their teachability (right). Results with human-robot interactions from non-experts teaching 5 robot embodiments on 78 tasks (\textcolor{plot-gray}{gray}) show that our framework (middle$^*$) can identify top performing users (\textcolor{plot-purple}{purple}), and leverage their interactions (only 14\% of task coverage) to drive LLM performance improvements for all users (\textcolor{plot-blue}{blue}) -- measured in terms teaching success rates on unseen tasks, responsiveness to user feedback, and number of user corrections. Experiments show that these improvements generalize to new robot embodiments and APIs.}
    {\scriptsize $^*$visualizations from real data: before and after circles are centered on mean good rating rates (\ie responsiveness to feedback metric, described in Sec. \ref{sec:experiment-results}) vs task success rates over all users.}
    \label{fig:teaser}
    \vspace{-1em}
}
\makeatother

\maketitle

\begin{abstract}
Large language models (LLMs) have been shown to exhibit a wide range of capabilities, such as writing robot code from language commands -- enabling non-experts to direct robot behaviors, modify them based on feedback, or compose them to perform new tasks.
However, these capabilities (driven by in-context learning) are limited to short-term interactions, where users' feedback remains relevant for only as long as it fits within the context size of the LLM, and can be forgotten over longer interactions.
In this work, we investigate fine-tuning the robot code-writing LLMs, to remember their in-context interactions and improve their \textit{teachability} \ie how efficiently they adapt to human inputs (measured by average number of corrections before the user considers the task successful).
Our key observation is that when human-robot interactions are viewed as a partially observable Markov decision process (in which human language inputs are observations, and robot code outputs are actions), then training an LLM to complete previous interactions is training a transition dynamics model – that can be combined with classic robotics techniques such as model predictive control (MPC) to discover shorter paths to success.
This gives rise to Language Model Predictive Control (LMPC), a framework that fine-tunes PaLM 2 to improve its teachability on 78 tasks across 5 robot embodiments – improving non-expert teaching success rates of unseen tasks by $26.9\%$ while reducing the average number of human corrections from $2.4$ to $1.9$. 
Experiments show that LMPC also produces strong meta-learners, improving the success rate of in-context learning new tasks on 
unseen 
robot embodiments and APIs by $31.5\%$.
See videos, code, and demos at:
\href{https://robot-teaching.github.io/}{https://robot-teaching.github.io/}.
\end{abstract}

\IEEEpeerreviewmaketitle

\section{Introduction}

Natural language provides a rich and accessible interface for teaching robots -- with the potential to enable anyone with minimal training to direct behaviors, express preferences, and provide feedback. 
Recent works show that large language models (LLMs), pretrained on Internet-scale data, exhibit out-of-the-box capabilities that can be applied to robotics -- from planning a sequence of steps given language commands \cite{ahn2022can,huang2022language}, to writing robot code \cite{liang2023code,singh2023progprompt,yu2023language,ma2023eureka}. 
Language inputs can also be sequenced in a multi-turn setting for example, to generate and modify reward function code from human feedback to compose new quadruped behaviors via real-time motion control \cite{yu2023language} (example in \cref{fig:teaser}).

LLM-based robot teaching (as shown in \cref{fig:teaser}) can be driven by in-context learning \cite{brown2020language} (\eg on code and dialogue data), where previous interactions are kept as input context for subsequent ones. 
In-context learning occurs during inference without gradient updates to model weights, enabling fast adaptation to language instructions (via exemplar-based compositional generalization \cite{chan2022data,hupkes2020compositionality}). 
However, this adaptation is limited to short-term reactive interactions where the users' feedback remains relevant for only as long as it fits within the context size of the LLM.
As a result, if human instructions accumulate over longer multi-step interactions that fall outside the receding context horizon, previous instructions can simply be forgotten.

We are interested in improving LLMs' \textit{teachability} for robot tasks, \ie how efficiently they adapt to human feedback, by enabling LLMs to remember their in-context interactions.
Teachability in multi-turn language-based human-robot interaction (HRI) can be measured as the average number of human inputs (\eg corrections) $n$ before the robot succeeds at the task. 
For instance, $n=1$ refers to the standard zero-shot instruction following setting \cite{jang2022bcz,lynch2021language}.
Prior works propose to improve teachability by generating linguistic summaries of human feedback \cite{zha2023distilling} or preferences \cite{wu2023tidybot} that can be indexed into memory and later retrieved in-context to guide future interactions.
However, such methods are often constrained by in-context learning generalization (observed to be more ``exemplar-based'' \ie on the basis of similarity to in-context examples \cite{chan2022data,shepard1963stimulus}), as opposed to generalization from in-weights learning via fine-tuning (which tends to be more ``rule-based'' \ie on the basis of minimal features that support category boundaries in the training data \cite{chan2022data,ashby1986varieties}). 
Subsequently, prior methods excel at overfitting to training tasks, but offer limited generalization (\eg domain-level adaptation) to unseen tasks. 
Is it possible to leverage both forms of learning to address these shortcomings?

In this work, we investigate improving the teachability of robot code-writing LLMs via \textit{in-context learning} (fast adaptation) by day, and model \textit{fine-tuning} (slow adaptation) by night, to accelerate fast adaptation the next day.
Given a setting where non-experts teach robots new tasks with language, our goal is to study which methods of improvement (\eg via fine-tuning) can best leverage data collected from in-context learning to improve future teachability (as measured on unseen tasks).  
%
Our key observation is that when human-robot interactions are formulated as a partially observable Markov decision process (POMDP -- in which human language inputs are observations, and robot code outputs are actions), then training an LLM to autoregressively complete previous interactions can be viewed as training a transition dynamics model – that can be combined with classic robotics techniques such as model predictive control (MPC) to discover shorter paths to success.
%
This gives rise to Language Model Predictive Control (LMPC), where we train the LLM to predict imagined future rollouts of human-robot interactions -- and at inference time, sample multiple futures (with non-zero decoding temperature) to search for the best one and take the next action (\ie receding horizon control as a decoding strategy). 
Classically challenging HRI problems (such as modeling individual user preferences) become more straightforward \eg by simply conditioning LMPC rollouts on usernames (``user \_\_ might say...''), with the intuition that different users cover different areas of the POMDP.

Extensive experiments (via blind A/B evaluations) show that fine-tuning with LMPC improves the teachability of PaLM 2 \cite{anil2023palm} on 78 tasks across 5 robot embodiments (on simulated and real platforms) -- enabling non-experts to teach robots to achieve higher success rates on unseen tasks by $26.9\%$, and reduces average number of human corrections from $2.4$ to $1.9$. %
In particular, LMPC produces strong meta-learners -- teachability improvements generalize
to unseen embodiments, improving the success rate of in-context learning new tasks with new robot APIs by $31.5\%$.
Interestingly, we observe substantial gains from top-user conditioned LMPC, which (i) autonomously identifies top users (by performance on training tasks), (ii) groups their data together with a special username ``top-user,''  then (iii) conditions inference-time LMPC rollouts on this special username (\ie assume everyone is a top-user). 
Despite top users having seen only 14\% of tasks, experiments show this conditioning mechanism drives performance improvements for all users on all tasks, including unseen ones by $10.5\%$. 
LMPC also outperforms retrieval baselines \cite{zha2023distilling}, and user studies affirm that performance improvements are likely the result of changes in model capability, rather than user teaching proficiency.
Our approach is not without limitations -- we discuss these and areas for future work in \cref{sec:discussions}. 
\section{Related Work}
\label{sec:related}


\smallskip \noindent \textbf{Language and Robotics.}
A large body of work integrates language and robotics, including mapping language to planning primitives~\cite{tellex2011understanding,kollar10directions,matuszek2012learning,artzi2013weakly,karamcheti2017draggns}, imitation learning from demonstrations along with language instructions~\cite{jang2022bcz,lynch2021language,shridhar2022cliport,stepputtis2020language, mees2022calvin}, learning language-conditioned reward functions~\cite{mirchandani2021ella,jiang2019language,goyal2020pixl2r,cideron2019self,misra2017mapping,akakzia2020grounding}, and using language as corrective feedback to adapt or define new behaviors~\cite{zha2023distilling,cui2023lilac,coreyes2019guiding}. We refer the reader to comprehensive surveys for a more complete review of prior work in this area \cite{2020tellex-robots-use-language,luketina2019survey}.

Recently, LLMs trained on Internet-scale data have been shown to exhibit profound capabilities ranging from step-by-step planning \cite{ahn2022can,huang2022language,driess2023palme,xie2023translating,ding2023task,liu2023llm+,wu2023tidybot,ren2023robots}, writing robot code \cite{liang2023code,singh2023progprompt,zelikman2023parsel,yoneda2023statler,arenas2023prompt,mirchandani2023large}, commonsense reasoning~\cite{talmor2022commonsenseqa,kwon2023toward}, and acting as a proxy reward function capturing human preferences~\cite{kwon2023reward,hu2023language,yu2023language}. 
In this work, we are also interested in leveraging the power of LLMs for adapting and teaching new behaviors via language feedback \cite{sharma2022correcting,zeng2022socratic,ren2023robots,zha2023distilling,huang2022inner} -- but in contrast to prior work, we focus on not only evaluating online adaptation via in-context learning (\eg prompting LLMs), but also on how we can improve that adaptation via offline model fine-tuning.

\smallskip \noindent \textbf{In-Context Learning for Robot Adaptation.}
In-context learning is a form of supervised meta-training \cite{brown2020language}, where multiple examples and instructions \cite{ouyang2022training} from the same dataset are packed sequentially into a context buffer that is fed as input to an LLM with an unsupervised autoregressive completion objective \cite{vaswani2017attention}. 
The instructions and examples specify tasks (extending the concept of ``task prefixes'', \ie predefined token sequences \cite{raffel2020exploring,LLMs-are-multitask-learners}), where the model is expected to complete further instances of the task by predicting what comes next. 

In robotics, in-context learning (via prompting) has been used to elicit a wide-range of capabilities -- responding to feedback \cite{zeng2022socratic,huang2022inner}, modifying low-level behaviors \cite{yu2023language,sha2023languagempc,mirchandani2023large,arenas2023prompt,kwon2023reward}, remembering and applying user preferences \cite{wu2023tidybot}, and asking for help \cite{ren2023robots}. 
Most related to our work is~\citet{zha2023distilling}, which investigates robot teaching by summarizing human feedback, and indexing it in memory to be used again as in-context examples for similar future interactions via retrieval (\eg retrieval augmented generation (RAG) \cite{lewis2020retrieval}). 
In contrast, we focus on directly fine-tuning the underlying LLM to improve in-context learning from human language inputs (which can be multi-round contextual).
We find finetuning exceeds the performance of retrieval-based methods for teaching unseen tasks -- without additional external modules.

\smallskip \noindent \textbf{Improving LLM Alignment to User Feedback.}
Our work builds on an active area of research to \emph{align} LLMs with user intent~\cite{ouyang2022training}. 
A common approach is to supervised fine-tune (SFT) the model on expert human inputs and outputs, then use non-expert labeled rankings (preferences) from model outputs to train reward models for reinforcement learning from human feedback (RLHF)~\cite{christiano2023deep,stiennon2022learning,bai2022constitutional}. 
However, these works often focus on mapping from single user inputs to preferred outputs (\eg single-turn dialogue). 
In this work, we also investigate learning from human feedback, but we focus on improving the teachability of LLMs that write and improve robot code based on multi-turn, interactive human feedback.
Our LMPC approach uses SFT to model human-robot interaction dynamics, and uses inference-time search and receding horizon control to discover shorter paths (with fewer rounds of corrections) to task success.
\section{Language Model Predictive Control}

We investigate teachability in the context of language-based human-robot interactions, where users communicate with robots via text messages through a chat-like interface next to a simulated visualization of the robot and its surroundings using the MuJoCo simulation engine \cite{todorov2012mujoco} (see \cref{fig:data_collection}, more details in Appendix \ref{app:MJPC}).
User messages are free-form and up to users' discretion; they may include instructions, preferences, feedback, etc. 
In response to each message, the system outputs robot code, which is directly sent to a real-time motion controller on a simulated or real robot (\cref{sec:mjpc}).
Users then provide subsequent feedback based on the observed robot behavior.

Each human-robot conversation (\ie chat session) is goal-driven: users are asked to teach one task per session and at the end of each session label ``success'' or ``failure'' conditioned on whether they believe the robot to have completed the task. 
Chat sessions can consist of multiple chat turns (\ie human-robot input-output pairs) before success. 
On average, successful sessions run for 2-3 chat turns, while failure sessions run for 5-6 chat turns (see \cref{fig:data_collection}; bar plot shown in bottom left). 
User messages can be corrections or broken-up step-by-step sub-tasks to piece together more complex ones, and they are usually multi-round contextual.
During data collection, users rate individual robot responses as `good' or `bad' --- good if the robot responded correctly to the most recent human feedback (although it may not be successful at completing the entire task yet), and bad otherwise.
We find that the ratio of good chat turn ratings correlates with task success (\cref{fig:data_collection}, bottom right).




\begin{figure*}[!t]
    \centering
    \vspace{0.8em}
    \includegraphics[width=0.99\linewidth]{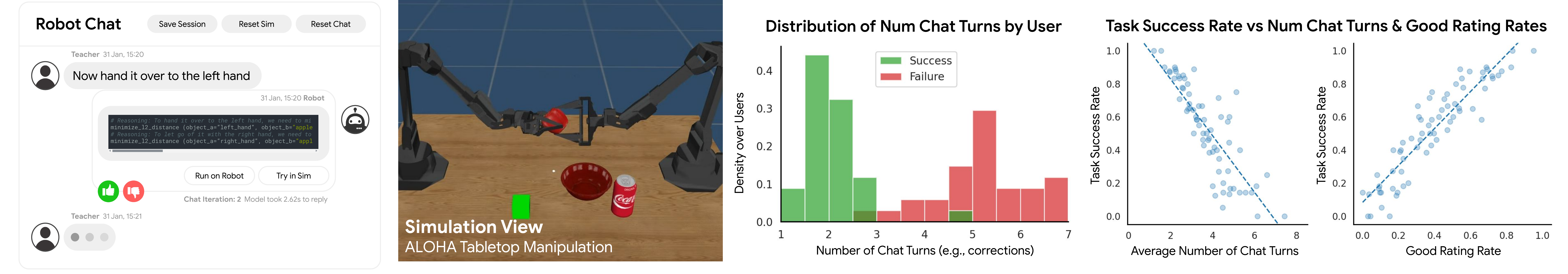}
    \caption{
        Our chat interface (left) allows non-experts to use language to teach robots new behaviors (visualized in simulation). Our LLM responds with reward code, to drive real-time motion control of a simulated or real robot. Statistics (right) show that base model data meets expectations: successful teaching sessions take fewer chat turns than failures, and task success rates correlate with fewer chat turns ($r=-0.85$) and higher good rating rates (\ie responsiveness to feedback, $r=0.92$).
    }
    \label{fig:data_collection}
    \vspace{-1.5em}
\end{figure*}

\subsection{Problem Statement}
\label{sec:problem-statement}

Our goal is to improve the teachability of LLMs that follow human instructions and feedback to write robot code.
Teachability is defined as the average number of human inputs (chat turns) $n$ before the robot succeeds at the task. 
This metric measures how efficiently the robot adapts to human inputs, and $n=1$ is equivalent to a standard zero-shot instruction following setting \cite{jang2022bcz,lynch2021language}. 
To \textit{improve} teachability is to reduce the number of chat turns $n$ before a desired success rate, and can be viewed as a meta-learning objective -- i.e., learning to learn faster from human feedback \cite{hospedales2021meta}. 
Intuitively, improving teachability of a model should encourage its responsiveness to feedback, as a means to maximize the likelihood of generating the correct behavior (according to the user). 
Teachability can also reflect how well a model adapts to preferences.
For instance, user input ``move a bit to the left'' might yield different robot behavior modifications depending on the user -- a strong meta-learner (with respect to teachability) is one that can learn this difference to minimize the number of interactions $n$, conditioned on with whom it interacts.

During our language-based human-robot interaction, the LLM interacts with the human teacher through code that is executed via motion control on the robot, and the human gives natural language feedback and indicates the success of the teaching session. The LLM’s goal is to produce code that leads the robot to behave as intended by the human, however, this target behavior has to be inferred from the human feedback. This is analogous to a partially observable Markov decision process (POMDP), where a policy (the LLM) is trained to generate actions (robot code) from observations (natural language feedback) in order to maximize reward (human indicated success). Improving teachability here can then be considered as an additional time-penalty term in the reward that encourages the model to achieve task success with as few interactions as possible.

Our approach is driven by two complementary forms of LLM improvement: 
(i) in-context learning (fast adaptation) for users to teach the model new tasks \textit{online} (\cref{sec:fast-adaptation}), 
and (ii) Language Model Predictive Control (LMPC) fine-tuning (slow adaptation) to update the model weights \textit{offline} (\cref{sec:slow-adaptation}). 
Our main contribution is developing a slow adaptation method (LMPC) that improves fast adaptation (measured via teachability of the model on unseen tasks).
To that end, we developed a system that enables fast adaptation of robot behaviors from natural language feedback --- we first explain how this system converts language feedback to robot actions through LLM in-context learning of robot reward code; then we explain how we use the collect feedback data to fine-tune and improve LLM teachability.


\subsection{Fast Adaptation with In-Context Learning}
\label{sec:fast-adaptation}

Fast Adaptation involves: 1) an LLM converting multi-turn language inputs to robot reward code and 2) converting robot reward code to robot actions.

\smallskip \noindent \textbf{Language to Robot Reward Code.}
In this work, fast adaptation is driven by in-context learning, where the language model is conditioned on a prompt that provides the initial tokens in the sequence $x_{1:k} = (x_1, \dots, x_k)$ and uses the model to complete $x_{k+1:n}$. 
Our in-context prompt uses PromptBook formatting \cite{arenas2023prompt}, which contains a description of the embodiment, the available robot APIs, as well as 1-2 example episodes (chat sessions) between the user and LLM, followed by the current chat session (full prompts in Appendix \ref{app:prompts}):

\vspace{0.3em}
\noindent\fbox{\parbox{0.97\linewidth}{\footnotesize{\texttt{{
\codecomment{\# You are a stationary robot arm with a 3-fingered hand.}\\
{\color{code-syntax}class} Robot:\\
\codetab {\color{code-syntax}def} {\color{code-function}reach}(self, obj):\\
\codetab {\color{code-syntax}def} {\color{code-function}min\_L2\_dist}(self, obj1, obj2):\\
\codecomment{\# Example Session.}\\
\codecomment{\# Chat Turn \#1: move the red and green things together.}\\
{\color{code-function}reach}(obj={\color{code-constant}`red'}, weight={\color{code-constant}1.0})\\
{\color{code-function}min\_L2\_dist}(obj1={\color{code-constant}`red'}, obj2={\color{code-constant}`green'}, weight={\color{code-constant}1.0})\\
\textbf{$\cdots$}
}}}}}\\

Users interact with the LLM in an interactive process -- provide feedback based on observing robot behaviors online, rather than labeling offline LLM data.
We use an existing pre-trained LLM PaLM 2 \cite{anil2023palm}, with which using the above prompts yields non-zero initial task success rates given feedback from the user.
The code generated within each turn can either be a single reward function, or a sequence of multiple reward functions.
Upon terminating a chat session, the interaction data is saved into a cached dataset to be used for slow adaptation.
Note that even with human inputs, the model may struggle to perform certain tasks -- experiments in \cref{sec:experiment-results} show that slow adaptation is needed to unlock fast adaptation on these tasks.

Fast adaptation requires fast LLM inference runtime speeds, so that latencies do not negatively influence the human-robot interactions. 
Our model inference runs at $~100$ tokens per second, and returns robot reward code expressed with 200 - 300 tokens on average (which amounts to roughly 10-15 lines of code). 
The median duration for each chat turn is $56$s, and the majority of user time is spent observing the robot performing the task in simulation.

\smallskip \noindent \textbf{Reward Code to Robot Motions.}
\label{sec:mjpc}
We use robot reward code as an interface between LLM and robot actions. 
This leverages the effective high-level reasoning capabilities of LLMs to translate user intent into semantically meaningful reward functions, which are then used to drive low-level motion control for the robot in real-time, providing immediate visual feedback to the user. 
To realize low-level robot actions from robot rewards, we build on \citet{yu2023language}, where given a reward function generated by the LLM, MuJoCo Model Predictive Control (MJPC)\footnote{https://github.com/google-deepmind/mujoco\_mpc}\cite{howell2022predictive} is used to synthesize robot motions. 
Using MuJoCo simulations as the robot dynamics model, MJPC implements a receding horizon trajectory optimization algorithm to find an action sequence that maximizes a given reward in real-time (simultaneously optimizing and executing robot actions).
With MJPC, after the LLM outputs a reward code, the reward code can be immediately executed on the robot.
This allows users to quickly observe changes in robot behavior as a result of their language feedback, enabling an interactive robot teaching experience.


%
Our code format extends \citet{yu2023language} with 2 notable changes to expand the expressiveness of behaviors across embodiments:
(i) \citet{yu2023language} relied on two prompts to respond to task commands - one to generate high-level motion descriptions in natural language, and another to convert those into reward code. 
In our approach, we only use one prompt that embeds motion descriptions as comments interspersed between the lines of the reward code. This Chain-of-Thought style prompting simplifies reward code writing and enables more flexible code generation.
(ii) \citet{yu2023language} can only specify one reward function (robot behavior) at a time. 
In our approach, the LLM can sequence multiple reward functions together by writing condition functions that signify when the robot should transition from one reward function to the next.
Here is an example of an LLM response to a task that involves transferring an object from one arm to another:


\vspace{0.3em}
\noindent\fbox{\parbox{0.97\linewidth}{\footnotesize{\texttt{{
\codecomment{\# To pick up the apple, bring it close to the left gripper.}\\
{\color{code-function}min\_L2\_dist}(obj1={\color{code-constant}`left\_hand'}, obj2={\color{code-constant}`apple'}, weight={\color{code-constant}5.0})\\
\codecomment{\# To lift up the apple, get its position and increment along z.}\\
pos = {\color{code-function}get\_obj\_pos}(obj={\color{code-constant}`apple'})\\
{\color{code-function}set\_target\_pos}(obj={\color{code-constant}`apple'}, (pos[{\color{code-constant}0}], pos[{\color{code-constant}1}], pos[{\color{code-constant}2}] + {\color{code-constant}0.25}))\\
\codecomment{\# Wait until the apple is in the air.}\\
{\color{code-syntax}def} condition\_fn():\\
\codetab {\color{code-syntax}return} {\color{code-function}get\_obj\_pos}(obj={\color{code-constant}`apple'})[{\color{code-constant}2}] >= {\color{code-constant}0.25}\\
{\color{code-function}wait\_until\_condition}(condition\_fn)\\
\codecomment{\# To hand over the apple, bring it close to the right gripper.}\\
{\color{code-function}min\_L2\_dist}(obj1={\color{code-constant}`apple'}, obj2={\color{code-constant}`right\_hand'}, weight={\color{code-constant}5.0})\\
\codecomment{\# Now let go of the apple with the left gripper.}\\
{\color{code-function}min\_L2\_dist}(obj1={\color{code-constant}`left\_hand'}, obj2={\color{code-constant}`apple'}, weight={\color{code-constant}0.0})
}}}}}\\

\noindent Functions such as \texttt{\small\color{code-function}min\_L2\_dist} and \texttt{\small\color{code-function}set\_target\_pos} directly set reward terms for real-time MJPC, which returns high-rate low-level action trajectories that maximize rewards.

\subsection{Slow Adaptation with Model Fine-Tuning}
\label{sec:slow-adaptation}

Gathering interaction data from in-context learning (fast adaptation) allows us to fine-tune the underlying LLM (slow adaptation) to improve its ability to both write useful robot reward code and respond to human feedback, and subsequently improve teachability.
In this work, we propose Language Model Predictive Control (LMPC)~\footnote{Not to be confused with MJPC, the MPC-based algorithm that uses a MuJoCo simulation to generate robot actions from robot reward code}, a supervised fine-tuning (SFT) technique that improves LLMs' teachability for robot tasks via modeling and optimizing over human-robot teaching sessions.
We further improve performance of fine-tuned models by conditioning LLM response on users during training, and on top-users during inference.

\smallskip \noindent \textbf{Language Model Predictive Control.}
We are interested in learning the human-robot interaction process using LMPC. Denoting the system prompt as $P$, human text inputs as $h_t$, robot code outputs as $c_t$ at chat turn $t$, and final human indicated success as $r$, we can represent an entire chat session as $[P, h_0, c_0, h_1, c_1, \dots, h_T, c_T, r]$.


Given the current chat session (system prompt and current chat history), the LLM is trained to autoregressively predict the rest of the chat session (sequence of $h_t$ and $c_t$'s, until receiving a reward $r$ at the end of the episode).
For training, the input to the LLM is the system prompt $P$ with the initial user instruction $h_0$; both are included because different robot embodiments have different system prompts (robot APIs), and this allows the LLM generation to support different robot APIs at inference time.
The target is for the LLM to predict any remaining portions of a chat session, conditioned on the current portion.
We only train LMPC-Rollouts on successful trajectories (training on both successes and failures yielded much worse performance. 
See Appendix \ref{app:additional_results}).
Training Transformers with causal attention on entire chat sessions is analogous to training a sequence-conditioned transition dynamics model of a POMDP, which is used for search during inference.

A key aspect of LMPC is that at inference time, the fine-tuned LLM is used as a transition model together with model predictive control (MPC) to discover optimal paths to success. 
MPC can be thought of as a sequence-level decoding strategy \cite{freitag2017beam}, but differs from standard ones used in modern language models as it generates multiple episodic rollouts to search for the next best action, and repeats the process at every decision-making step.
To do so, we sample from the LLM $8$ rollouts with non-zero temperature sampling (next-token decoding) for a max token length of $4096$.
If a sampled trajectory reaches termination within the max token length, we treat it as successful, since LMPC-Rollouts is only trained on successful data.
From these terminated samples, we choose the trajectory with the fewest predicted timesteps (\ie chat turns) and return its first action, as shown in Fig~\ref{fig:lmpc-rollout-vs-skip} (center). 
This strategy can be derived from optimizing a cumulative cost in the trajectory (assuming a sparse reward of 1 for success and a constant time penalty), which is inspired by existing trajectory optimization works in control field.
If no sample terminates, then we randomly pick a trajectory and return its $a_{t+1}$.
This process is then repeated given new human input for every chat turn. 
Intuitively, LMPC-Rollouts can be thought of as training the LLM via human-robot interaction as a form of chain-of-thought \cite{wei2022chain} during both training and inference –- rather than cloning successful code, LMPC learns the process of getting to the correct code, and accelerating it via search at inference time.

\smallskip \noindent \textbf{Top-User Conditioning.}
\label{sec:expert-conditioning}
To further improve LLM teachability with fine-tuning, we propose conditioning LLM generations, during both training and inference, on the user. 
For training, we modify the input prompt to include which user generated the following chat session using a unique ID label.
Top-users are autonomously identified from the training dataset and are given a special ID ``top-user."
During inference, we always condition LLM generations on the ``top-user" label.
We identify top-users as the top $25\%$ of users by their user performance score.
This score is the average of a user's task success rate weighted by task difficulty, which is the task's failure rate across all users.
See Appendix \ref{app:top_users} for more details.

Top-user conditioning, in the context of LMPC, can be interpreted as conditioning the LLM to generate the distribution of observations $o_t$ (expected human inputs) and actions $a_t$ (expected code outputs) closest to the top $75$th percentile of users. 
Intuitively, if observations are viewed as a partial noisy representation of the true (user) state (or intent, during teaching), then different user proficiency levels can correspond to a varying amount of noise (\ie higher proficiency is less noise), to which conditioning on top-users prompts the LLM to generate rollouts with less noise. 
Top-user conditioning draws similarity to performance conditioning with Decision Transformers \cite{chen2021decision}, albeit (in the absence of dense rewards) using inference-time search via MPC. 
Note that top-user conditioning can broadly index distributions that represent a wide range of user-related attributes (\eg preferences, user-specific styles, etc.), expanding beyond the scope of what performance conditioning on rewards alone can provide.

\section{Experiments}

Our experiments evaluate how much the various proposed finetuning strategies (slow adaptation) improve online in-context learning (fast adaptation) for humans interactively teaching robots via natural language feedback.
Evaluations are performed on $78$ robot tasks, across $5$ robot embodiments in simulation and $2$ on real hardware. 
We specifically explore the following questions:
\begin{itemize}[leftmargin=12pt]
    \item How much does fine-tuning improve teachability, especially on test tasks?
    \item How do LMPC-Rollouts and LMPC-Skip compare?
    \item What are the benefits of Top-User Conditioning?
    \item Does finetuning enable cross-embodiment generalization?
    \item Can iterative finetuning further improve teachability?
\end{itemize}

All data collection and most evaluations were performed in simulations.
All models were trained on data obtained with simulation.
We separately evaluate finetuned models on real robots, but we have not experimented with training on data from teaching real robots.

\subsection{Data Collection and Evaluation}

To collect human teaching data and evaluate performance, we worked with $35$ non-expert users, who collected $350$ chat sessions per day. 
These users are not researchers or engineers, and they are not familiar with the underlying LLMs or robot code.
We instruct users to give natural language feedback on the behavior of the robot for each chat turn, instead of giving technical feedback or giving feedback on the robot code.
The user can give at most 7 rounds of feedback before the session is considered unsuccessful.
Data collection protocol details are in Appendix \ref{app:data_collect}.
When a user starts a new chat session, a random robot embodiment and task is sampled, and the user is asked to teach the robot that task.
Data collection is separated into two phases: 1) initial data collection with the base model and 2) subsequent data collection (evaluations) with finetuned models.
In phase 2, we randomly sample which model the user interacts with, and the user does not know which model they are currently engaging with.
This allows for blind A/B evaluations.

Out of the $78$ tasks, $51$ are train tasks ($65\%$), while $27$ are test ($35\%$).
While separating tasks into train and test splits allows us to measure model generalization performance, it also means there are less data available for training.
To address this and also to make the data distribution robust to user teaching noise, data collection and evaluation of models are typically aggregated across 2 days.
Additional data filtering were performed to remove invalid and incorrect data ($<4\%$).
In total, $299$ successful chat sessions from the initial data collection were made available for fine-tuning.
Across chat sessions, the max total token length is $3900$, with $1800$ as the median.
Given the limited amount of data, and to make LLM responses more robust to small differences in user feedback, we perform data augmentation on the collected data.
This is done by generating $5$ variations of user instructions (as well as intermediate feedbacks for training LMPC-Rollouts) using PaLM 2-L.
We do not generate variations of the robot code.
Combining the augmented and the original data, the training set contains about $3$M tokens.

For evaluations, we collect approximately $350$ chat sessions for all model variants we evaluate, split across all platforms and tasks.
We observe minimal user performance drift over time (see Appendix ~\ref{app:user_drift}), so differences in model performance are likely due to changes in model capabilities, and not in users' teaching proficiency.

\subsection{Robot Embodiments and Tasks}
\label{sec:experiments-robot-embodiments-and-tasks}

In this section we give a brief overview of the $5$ robot embodiments in our experiments.
We chose these embodiments to explore teaching a diverse set of robot capabilities, from tasks that require a single arm, to bi-manual tasks, and to dexterous and locomotion tasks.
See~\cref{fig:teaser} for illustrations.
We include the full list of tasks each embodiment in the Appendix \ref{app:tasks}.

\smallskip \noindent\textbf{1. Robot Dog.}
This is a small custom quadruped robot~\cite{caluwaerts2023barkour}.
Robot Dog has a total of 12 actuated degrees-of-freedom (DoF), 3 on each leg. 
Tasks range from stationary posing tasks, like sitting and high-five, to more dynamic tasks, like trotting and door opening.
We perform Robot Dog experiments in both simulation and the real world.

\smallskip \noindent\textbf{2. Mobile Manipulator}
In this embodiment, we use a mobile manipulator~\cite{herzog2023deep}
with a 7 DoF arm and parallel jaw grippers.
We explore tabletop manipulation tasks with rigid objects, such as flipping and stacking objects.
The Mobile Manipulator is also available both in simulation and the real world.

\smallskip \noindent\textbf{3. Aloha.}
This is a bi-manual robot with two 6 DoF arms, each attached with a parallel jaw gripper ~\cite{zhao2023learning}. 
The two arms sit directly opposite of each other on a table that has a set of rigid household objects.
With Aloha, we explore tasks that require coordination with both arms, such as object transfers.

\smallskip \noindent\textbf{4. Bi-arm Kuka.}
THis is a bi-manual robot with two 7 DoF Kuka LBR IIWA14 arms without end-effectors.
The omission of end-effectors allows us to explore whole-body manipulation tasks (e.g. manipulating objects with any part of the robot arm) with this embodiment.
The workspace has boxes of different sizes and colors, and the robot needs to manipulate individual or sets of objects to desired goal locations (which may be on the workspace surface or in the air) and in a given order.

\smallskip \noindent\textbf{5. Kuka+Hand.}
This has a custom three-fingered hand attached to one 7 DoF Kuka arm.
All DoFs are controlled via torque control.
Along with the arm, a set of rigid objects is provided in the workspace.
With Kuka+Hand, we explore dexterous manipulation tasks that are difficult to perform with the other manipulation embodiments, such as lifting multiple objects in-hand and plug insertion.

\subsection{Compared Methods}

\begin{figure}[!t]
    \centering
    \vspace{0.5em}
    \includegraphics[width=\linewidth]{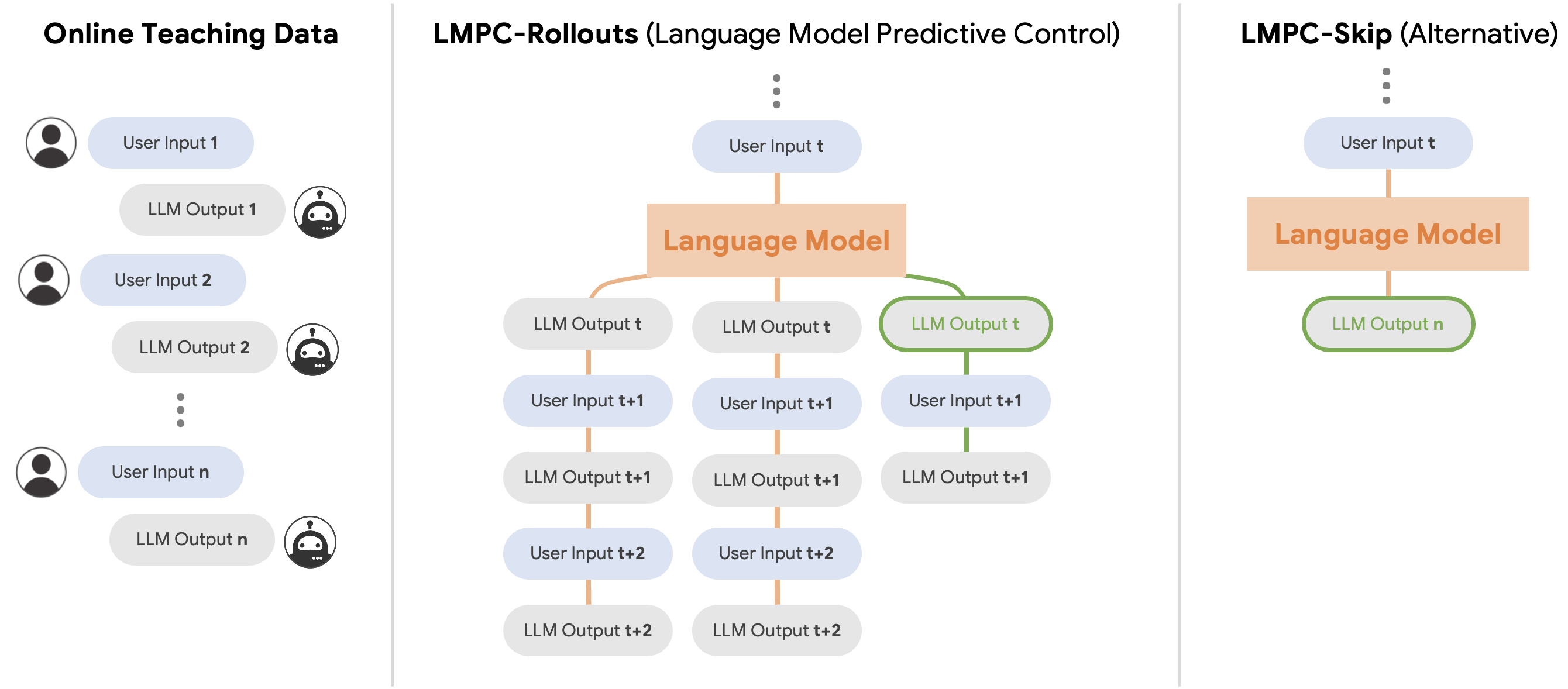}
    \caption{
    Given a dataset of users teaching robots new tasks with language (represented as text inputs and code outputs from online in-context learning -- left), LMPC-Rollouts is trained to predict subsequent inputs and outputs conditioned on the current chat history (middle), and uses MPC (receding horizon control) for inference-time search to return the next best action (with fewest expected corrections before success). LMPC-Skip is an alternate variant that is trained to directly predict the last action (right). Both LMPC variants accelerate fast robot adaptation via in-context learning.
    }
    \label{fig:lmpc-rollout-vs-skip}
    \vspace{-1em}
\end{figure}

We compare performances across the base model (PaLM 2-S), finetuned models, and a Retrieval-Augmented Generation (RAG)~\cite{lewis2020retrieval} baseline.
For finetuned models, we compare two variants: 1) \textbf{LMPC-Rollouts} --- our proposed method that learns to simulate chat session rollouts and performs MPC online as, and 2) \textbf{LMPC-Skip} --- a model that is trained to directly predict the last, correct code, skipping predictions of the interim trajectory (see Fig~\ref{fig:lmpc-rollout-vs-skip}, right). 
LMPC-Skip is encouraged to predict the final correct code as soon as possible \eg optimizing for 1-turn success.
However, because LMPC-Skip is not trained on nor does it model intermediate interactions with the user, it may be less responsive to corrective feedback.
During inference, we condition LMPC-Skip's generation on the system prompt with the chat session so far, and only query the model once to generate a response.
Like LMPC-Rollouts, LMPC-Skip is only trained on successful chat sessions.

Comparing LMPC-Rollouts and LMPC-Skip captures the difference between finetuning the LLM to leverage and predict the entire human-robot chat interaction, versus skipping to predicting the final robot-code response.
Comparing these finetuned methods to RAG captures if the LLM's improvement in our domain is possible if we do not have access to model weights or the resources needed for finetuning.
For RAG, we use a pretrained embedding model to retrieve relevant examples from the training data then inserting them into the LLM context, similar to other RAG applications for adapting robot behavior~\cite{zha2023distilling}.
See implementation details in the Appendix \ref{app:rag}.

\subsection{Experiment Results}
\label{sec:experiment-results}

\begin{table*}[!t]
    \centering
        \begin{tabular}{llcccccc}
        \toprule
        Tasks & Model & Success Rate & Num Chat Turns & Good Rating Rate & Successful Tasks Rate & 1 Turn Success Rate & 2+ Turn Success Rate \\
        \midrule
        Train & PaLM 2-S & 34.8\% & 2.3 & 16.7\% & 74.0\% & 13.0\% & 21.7\% \\
        & RAG & 46.4\% & 2.2 & 21.4\% & \textbf{83.3\%} & 25.1\% & 21.2\% \\
        & LMPC-Skip & \textbf{56.0\%} & \textbf{1.7} & \textbf{25.6\%} & \textbf{83.3\%} & \textbf{34.6\%} & 21.4\% \\
        & LMPC-Rollouts & 51.9\% & 2.2 & 21.8\% & 74.0\% & 23.5\% & \textbf{28.4\%} \\
        \midrule
        Test & PaLM 2-S & 39.4\% & 2.4 & 18.1\% & 81.5\% & 17.5\% & 21.9\% \\
        & RAG & 51.9\% & 2.0 & 20.9\% & 75.0\% & 27.9\% & 24.0\% \\
        & LMPC-Skip & 59.4\% & \textbf{1.6} & 24.7\% & \textbf{88.9\%} & \textbf{41.7\%} & 17.8\% \\
        & LMPC-Rollouts & \textbf{66.3\%} & 1.9 & \textbf{26.5\%} & \textbf{88.9\%} & 34.8\% & \textbf{31.5\%} \\
        \bottomrule
        \end{tabular}
    \caption{Comparing base and finetuned models across all embodiments. \emph{Success}: overall success rate on all tasks. \emph{Num Chat Turns}: mean number of chat turns for successful chat sessions. \emph{Good Rating}: proportion of positively rated chat turns after the turn. \emph{Successful Tasks}: proportion of tasks with at least one successful chat session. \emph{1 turn Success}: the proportion of chat sessions that were successful with just one chat turn. \emph{2+ turn Success}: the proportion of chat sessions that were successful with two or more chat turns.}
    \label{tab:model_metrics}
    \vspace{-1.5em}
\end{table*}

\begin{figure}[!t]
    \centering
    \vspace{0.5em}
    \includegraphics[width=\linewidth]{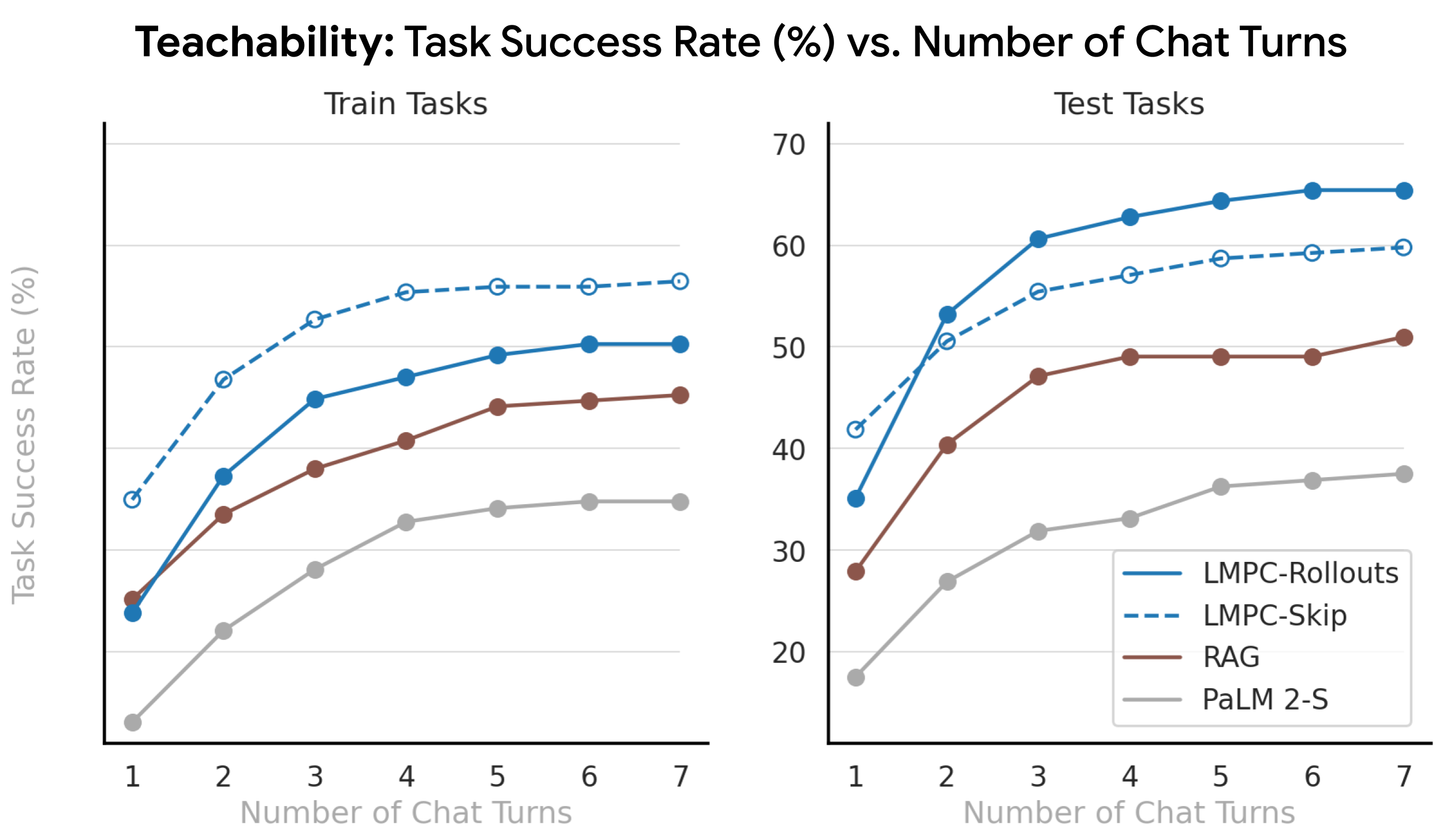} 
    \caption{Our fine-tuned LLMs with LMPC-Rollouts and LMPC-Skip improve the teachability of the base model (PaLM 2-S), and outperforms a RAG~\cite{lewis2020retrieval} baseline across all embodiments. LMPC-Skip overfits to train tasks (left), while LMPC-Rollouts generalizes better (\ie more teachable and responsive to feedback) on unseen test tasks (right) for multi-turn sessions (with more than one chat turn). }
    \label{fig:task_success_nci_all}
    \vspace{-1em}
\end{figure}

We evaluate the LLM's \emph{teachability} as task-success for $<N$ user interactions or ``chat turns".
This is visualized in a curve in~\cref{fig:task_success_nci_all}, where each point indicates the proportion of chat sessions that achieved success (y-axis) with equal to or less than a certain number of chat turns (x-axis).
Models that have better teachability would have a curve that is higher and to the left.

\cref{fig:task_success_nci_all} reports the main teachability results aggregated over all embodiments for the base PaLM 2-S model, the finetuned models LMPC-Rollouts and LMPC-Skip, and the base model with RAG.
Through finetuning, models are able to exceed teachability performance of the base model.
On train tasks, LMPC-Skip performs the best.
On test tasks, LMPC-Rollouts perform the best, improving success rate over the base model by $27\%$.
Both models also reach high success rates faster than the base model - matching or exceeding the final success rate of the base model after just one chat turn.
While LMPC-Skip achieves the higher 1-turn success rate than LMPC-Rollouts on test tasks, the order flips starting at 2 chat turns.
This suggests that LMPC-Rollouts is more amenable to improvements from user feedback.
RAG performs competitively over the base model, but it trails behind the finetuned methods in both train and test tasks.
See these results separated by embodiment in~\cref{fig:task_success_nci} in Appendix \ref{app:additional_results}.

\cref{tab:model_metrics} provides additional quantitative comparisons across all models evaluated, including:
\begin{itemize}[leftmargin=12pt]
    \item \emph{Success Rate}: overall success rate on all tasks and embodiments
    \item \emph{Num Chat Turns}: mean number of chat turns for successful chat sessions
    \item \emph{Good Rating Rate}: proportion of positively rated chat turns after the first chat turn (captures responsiveness to corrective feedback). Note this rating is given per chat turn, not the entire chat session.
    \item \emph{Successful Tasks Rate}: the proportion of tasks with at least one successful chat session
    \item \emph{1 turn Success Rate}: proportion of chat sessions that were successful with just one chat turn (1st instruction)
    \item \emph{2+ turn Success Rate}: proportion of chat sessions that were successful with $>1$ chat turns. This is the difference between the overall success rate and 1 turn Success Rate
\end{itemize}
For both train and test tasks, LMPC-Skip achieves the lowest Num Chat Turns for successful chat sessions, as well as the highest 1-turn Success Rate.
These reflect how LMPC-Skip is trained to predict the final code as fast as possible.
However, LMPC-Rollouts has the highest 2+ turn Success Rate, suggesting it is most amenable to corrective feedback given an incorrect first response.
To maximize performance in practice, these results suggest that one should use LMPC-Skip for responding to the initial user instruction, then LMPC-Rollouts for responding to subsequent user feedback.
For RAG, while the method improves upon the base model on overall success rate, it achieves lower Successful Task Rate than the base model on test tasks.
This suggests that while RAG may be proficient at increasing the success rate of tasks similar to the retrieved examples, it struggles to perform well on novel tasks.

\begin{table}[!t]
    \centering
    \begin{tabular}{llcc}
    \toprule
    Data & Model & Train Tasks & Test Tasks \\
    \midrule
    All Users & LMPC-Rollouts & -8.4\% & -10.5\% \\
    & LMPC-Skip & -16.3\% & -26.1\% \\
    \midrule
    Only Top Users & LMPC-Rollouts & -23.8\% & -21.7\% \\
    & LMPC-Skip & -9.6\% & -13.6\% \\
    \bottomrule
    \end{tabular}
    \caption{
        Changes in success rate without Top-User Conditioning.
        We evaluate two variants of LMPC-Rollouts and LMPC-Skip that do not apply top-user conditioning: training on data from all users and training on data from only top users.
        Success rates degrade significantly for both variants, suggesting that 1) focusing LLM generation on the style of top-users is important and 2) top-user data alone is insufficient, and training on the wider data distribution of all users is still important.
    }
    \label{tab:expert_conditioning}
    \vspace{-3em}
\end{table}

\smallskip \noindent \textbf{Effects of Top-User Conditioning.}
In~\cref{tab:expert_conditioning}, we show the change in task success when training without top-user conditioning on 1) data from all users and 2) data from only top users.
These ablations were only performed on the Robot Dog and Mobile Manipulator embodiments due to time constraints.
From the initial data collected on the base model, $10$ out of $35$ users were identified as top-users, and they only covered $11$ out of $50$ train tasks.
However, despite this small coverage, top-user conditioning significantly outperforms both variants of no top-user conditioning, across model types (LMPC-Rollouts and LMPC-Skip) and task types (train and test).
This suggests that with top-user conditioning, models can learn to transfer the style of responses induced by top-users teaching to novel tasks.
It also highlights the importance of training the LLM to mimic generations from a high-quality data distribution as well as across a diverse data distribution.
See Appendix \ref{app:top_users} for analysis on the teaching styles of top-users.

\begin{table*}[!th]
\centering
    \begin{tabular}{llcccc}
        \toprule
        & & \multicolumn{2}{c}{PaLM 2-S} & \multicolumn{2}{c}{LMPC-Rollouts} \\
        \cmidrule(lr){3-6}
        Embodiment & Task & Success & Num Chat Turns & Success & Num Chat Turns \\ 
        \midrule
        Robot Dog & ``downward dog''  & $100\%$ & $1.3$ & $100\%$ & $2.8$ \\ 
        & ``hop'' & $25\%$ & $2.0$ & $100\%$ & $2.3$ \\ 
         & ``high-five with left hand''$^*$  & $75\%$ & $2.3$ & $75\%$ & $3.0$ \\ 
        & ``walk forward in a trotting gait''$^*$ & $25\%$ & $2.0$ & $100\%$ & $2.8$ \\ 
        & ``hop while turning counterclockwise''$^*$ & $25\%$ & $5.0$ & $25\%$ & $4.0$ \\ 
        \midrule
        Mobile Manipulator & ``knock over coke can'' & $20\%$ & $5.0$ & $20\%$ & $3.0$ \\ 
        & ``open top drawer half-way''$^*$ &  $100\%$ & $3.4$ & $100\%$ & $3.2$ \\
        & ``push coke can from right to left''$^*$ & $60\%$ & $2.0$ & $80\%$ & $2.0$ \\ 
        \midrule
        \multicolumn{2}{c}{Average} & $53.8\%$ & $\mathbf{2.9}$ & $\mathbf{75\%}$ & $\mathbf{2.9}$ \\
        \bottomrule
    \end{tabular}
    \caption{
        LMPC-Rollouts has higher success than PaLM 2-S on real robots.
        Test tasks are starred$^*$. Robot Dog tasks are performed $4$ times, Mobile Manipulator tasks $5$ times.
    }
    \vspace{-1.5em}
\label{table:real}
\end{table*}

\begin{table}[!th]
    \centering
    \begin{tabular}{lccc}
    \toprule
    & \multicolumn{2}{c}{Train Embodiments} & Test Embodiments \\
    \cmidrule(lr){2-3}
    Model & Train Tasks & Test Tasks &  \\
    \midrule
    LMPC-Skip & $+28.8\%$ & $+19.0\%$ & $+18.6\%$ \\
    LMPC-Rollouts & $+17.2\%$ & $+23.8\%$ & $+31.5\%$ \\
    \bottomrule
    \end{tabular}
    \caption{
        Finetuned models can generalize to new robot embodiments and APIs not seen during training.
        Higher improvements in test tasks and embodiments are caused by the train:test split not being explicitly selected for uniform task difficulty and baseline performance; doing so is infeasible as the split needs to be chosen before starting evaluations, when task difficulty and baseline performance were unknown.
    }
    \label{tab:cross_embodiment}
    \vspace{-0.5em}
\end{table}

\smallskip \noindent \textbf{Cross-Embodiment Generalization.}
Beyond evaluating generalization towards test tasks, we also evaluate whether training on a subset of embodiments would lead to improved performance on new embodiments that the finetuned models were not trained on.
To the LLM, the difference in embodiment is captured through the prompt, which contains different robot descriptions and APIs for each embodiment.
We performed an experiment where we train the LMPC models on data from Robot Dog, Mobile Manipulator, and Aloha, omitting Bi-arm Kuka and Kuka+Hand.
See results in~\cref{tab:cross_embodiment}, where we report success rate differences between the finetuned models and the base model.
We see improvements in test embodiments of $18.6\%$ for LMPC-Skip and $31.5\%$ for LMPC-Rollouts, suggesting that finetuned models generalize not only to test tasks, but test robot embodiments as well.
This generalization is non-trivial as the embodiments have very different APIs from each other, and the test embodiments require writing robot reward code that can induce complex dexterous manipulation behaviors.

\begin{figure}[!t]
    \centering
    \includegraphics[width=\linewidth]{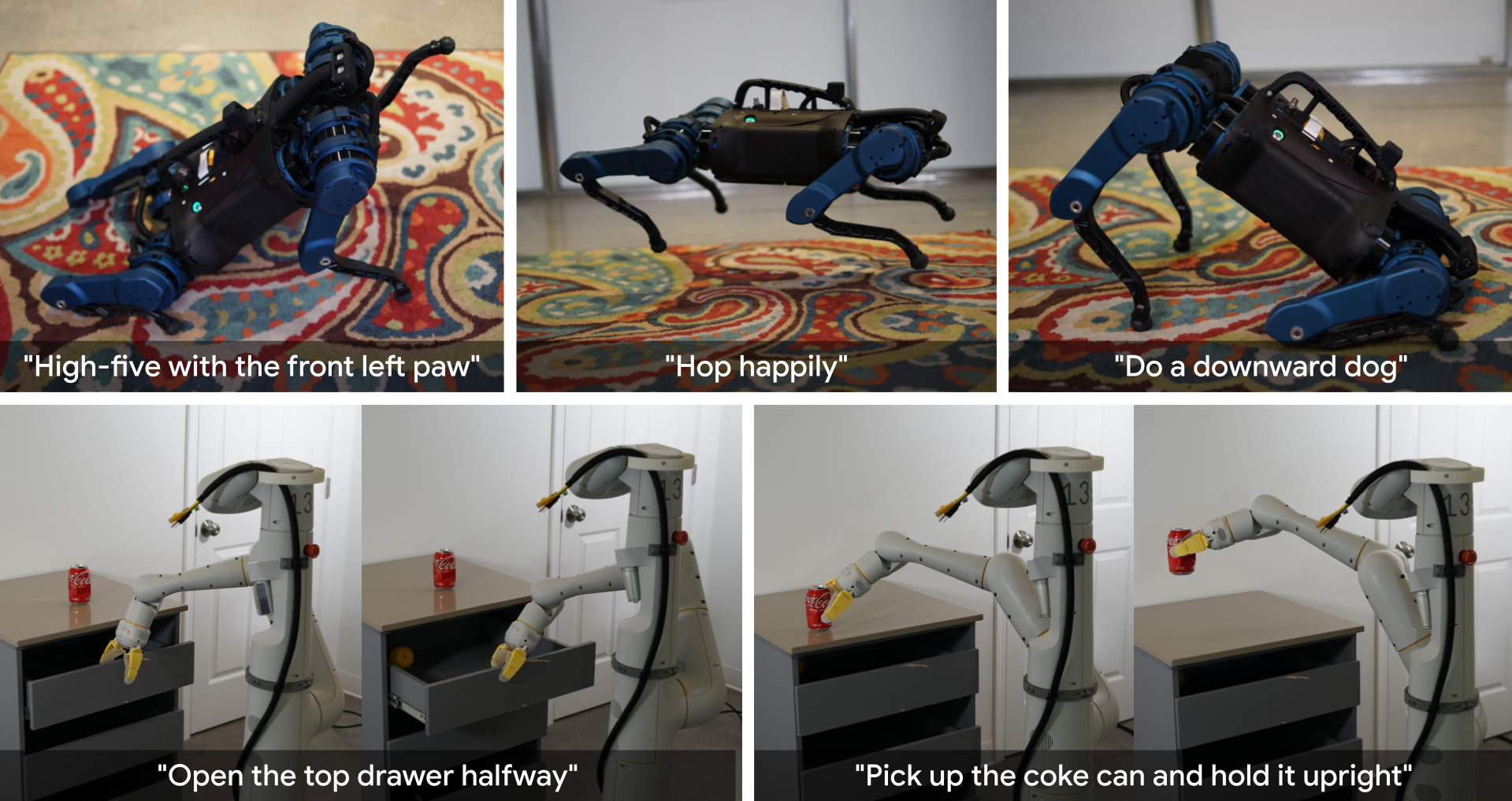}
    \caption{
        Tasks evaluated in the real-world Mobile Manipulator and Robot Dog.
    } 
    \label{fig:ameta-real}
    \vspace{-2em}
\end{figure}

\smallskip \noindent \textbf{Real-world Evaluations.}
While we did not perform data collection with real robots, we evaluated our approach on a subset of tasks for the Mobile Manipulator and Robot Dog in the real world (\cref{fig:ameta-real}).
Deploying our approach in the real world requires deploying MJPC on real robots.
For the real Mobile Manipulator, we used the method from \cite{yu2023language}, where MJPC is used as a motion planner with object state estimation done via an external vision pipeline.
This method however, does not work for robots that require high-frequency control feedback, such as quadrupeds.
To enable real-world teaching on the Robot Dog, we developed a policy distillation pipeline to train low-level end-to-end policies that are conditioned on reward terms from the LLM-generated reward code.
This can be thought of as training multi-task policies conditioned on latent task descriptors, which serve as an interface between high and low level control.
Please see additional real-world implementation details in \cref{app:real_robots}.

Real-world evaluations are done on $8$ tasks, and we collect $4$ real-world teaching sessions per task (where the generated reward code is executed on the real robot, instead of in simulation).

See results that compare PaLM 2-S and LMPC-Rollouts in \cref{table:real}.
LMPC-Rollouts achieves higher success rate than PaLM 2-S across all tasks.
While Num Chat Turns for successful sessions is about the same for PaLM 2-S and LMPC-Rollouts on these tasks, LMPC-Rollouts achieves much higher success rates.
See more detailed comparisons between sim and real executions in the Appendix \ref{app:real_robots}.

\begin{table}[!ht]
    \centering
    \begin{tabular}{lcc}
    \toprule
    & \multicolumn{2}{c}{Success Rate Diff from Iter 1} \\
    \cmidrule(lr){2-3}
    Model & Train Tasks & Test Tasks \\
    \midrule
    LMPC-Skip Iter 2 & +5.1\% & -4.7\% \\
    LMPC-Rollouts Iter 2 & -5.5\% & -1.9\% \\
    \bottomrule
    \end{tabular}
    \caption{
        Further finetuning on data generated from both the base model and the first finetuned models models does not yield performance improvements.
    }
    \vspace{-1em}
    \label{tab:day2}
\end{table}

\smallskip \noindent \textbf{Multiple Fine-tune Iterations.}
Given that the finetuned models exhibit improved teachability performance over the base model, additional, iterative training with data collected with the finetuned models could potentially further improve performance.
We tested this hypothesis by training Iteration 2 LMPC models with data collected by the Iteration 1 models.
Results are shown in~\cref{tab:day2}.
Currently, we do not observe further improvements from the second iteration of finetuning.
This implies that the data distribution or data amount used to train the second iteration of models do not differ significantly from that of the first iteration, so the resultant model behaviors remain largely unchanged.
While recent works have demonstrated iterative self-improving finetuning for LLMs~\cite{yuan2024self, honovich2022unnatural, gulcehre2023reinforced}, enabling LLM iterative improvement with human feedback and grounded on robot code executions remain promising but under-explored, and we defer this topic to future research.

\section{Discussions}
\label{sec:discussions}
We introduce a method that improves the teachability of LLMs by 
performing Language Model Predictive Control with LLMs finetuned to predict the dynamics of human-robot interactions.
LMPC can learn to learn faster from human feedback, and it generalizes to test tasks and test robot embodiments.
Despite the promising results, there are several limitations to our work that can point to potential future research.
We assume access to sufficient computational resources both for MJPC (e.g. 128 CPU cores) 
and for LLM finetuning.
More efficient MPC and finetuning techniques (e.g. LoRA~\cite{hu2021lora}) would help.
We also assume that the base LLM can generate some positive chat sessions for bootstrapping the learning process.
Our work only uses language models; future work on using multimodal models (e.g. to video/audio inputs) can expand the richness of feedback as well as improve finetuned models' ability to predict human reactions to robot behavior.
Lastly, we observed no benefit from multiple data collection and fine-tuning cycles.
Adapting the data distribution, with methods like active task exploration or synthetic data generation, may unlock additional performance gains.

\clearpage

\footnotesize
\bibliographystyle{plainnat}
\balance
\bibliography{references}

\begin{thebibliography}{76}
\providecommand{\natexlab}[1]{#1}
\providecommand{\url}[1]{\texttt{#1}}
\expandafter\ifx\csname urlstyle\endcsname\relax
  \providecommand{\doi}[1]{doi: #1}\else
  \providecommand{\doi}{doi: \begingroup \urlstyle{rm}\Url}\fi

\bibitem[Ahn et~al.(2022)Ahn, Brohan, Brown, Chebotar, Cortes, David, Finn, Fu, Gopalakrishnan, Hausman, et~al.]{ahn2022can}
Michael Ahn, Anthony Brohan, Noah Brown, Yevgen Chebotar, Omar Cortes, Byron David, Chelsea Finn, Chuyuan Fu, Keerthana Gopalakrishnan, Karol Hausman, et~al.
\newblock Do as i can, not as i say: Grounding language in robotic affordances.
\newblock \emph{arXiv preprint arXiv:2204.01691}, 2022.

\bibitem[Akakzia et~al.(2020)Akakzia, Colas, Oudeyer, Chetouani, and Sigaud]{akakzia2020grounding}
Ahmed Akakzia, C{\'e}dric Colas, Pierre-Yves Oudeyer, Mohamed Chetouani, and Olivier Sigaud.
\newblock Grounding language to autonomously-acquired skills via goal generation.
\newblock \emph{arXiv:2006.07185}, 2020.

\bibitem[Anil et~al.(2023)Anil, Dai, Firat, Johnson, Lepikhin, Passos, Shakeri, Taropa, Bailey, Chen, et~al.]{anil2023palm}
Rohan Anil, Andrew~M Dai, Orhan Firat, Melvin Johnson, Dmitry Lepikhin, Alexandre Passos, Siamak Shakeri, Emanuel Taropa, Paige Bailey, Zhifeng Chen, et~al.
\newblock Palm 2 technical report.
\newblock \emph{arXiv preprint arXiv:2305.10403}, 2023.

\bibitem[Arenas et~al.(2023)Arenas, Xiao, Singh, Jain, Ren, Vuong, Varley, Herzog, Leal, Kirmani, et~al.]{arenas2023prompt}
Montserrat~Gonzalez Arenas, Ted Xiao, Sumeet Singh, Vidhi Jain, Allen~Z Ren, Quan Vuong, Jake Varley, Alexander Herzog, Isabel Leal, Sean Kirmani, et~al.
\newblock How to prompt your robot: A promptbook for manipulation skills with code as policies.
\newblock In \emph{Towards Generalist Robots: Learning Paradigms for Scalable Skill Acquisition@ CoRL2023}, 2023.

\bibitem[Artzi and Zettlemoyer(2013)]{artzi2013weakly}
Yoav Artzi and Luke Zettlemoyer.
\newblock Weakly supervised learning of semantic parsers for mapping instructions to actions.
\newblock \emph{Transactions of the Association for Computational Linguistics (TACL)}, 1:\penalty0 49--62, 2013.

\bibitem[Ashby and Townsend(1986)]{ashby1986varieties}
F~Gregory Ashby and James~T Townsend.
\newblock Varieties of perceptual independence.
\newblock \emph{Psychological review}, 93\penalty0 (2):\penalty0 154, 1986.

\bibitem[Bai et~al.(2022)Bai, Kadavath, Kundu, Askell, Kernion, Jones, Chen, Goldie, Mirhoseini, McKinnon, et~al.]{bai2022constitutional}
Yuntao Bai, Saurav Kadavath, Sandipan Kundu, Amanda Askell, Jackson Kernion, Andy Jones, Anna Chen, Anna Goldie, Azalia Mirhoseini, Cameron McKinnon, et~al.
\newblock Constitutional ai: Harmlessness from ai feedback.
\newblock \emph{arXiv preprint arXiv:2212.08073}, 2022.

\bibitem[Brohan et~al.(2023)Brohan, Brown, Carbajal, Chebotar, Dabis, Finn, Gopalakrishnan, Hausman, Herzog, Hsu, Ibarz, Ichter, Irpan, Jackson, Jesmonth, Joshi, Julian, Kalashnikov, Kuang, Leal, Lee, Levine, Lu, Malla, Manjunath, Mordatch, Nachum, Parada, Peralta, Perez, Pertsch, Quiambao, Rao, Ryoo, Salazar, Sanketi, Sayed, Singh, Sontakke, Stone, Tan, Tran, Vanhoucke, Vega, Vuong, Xia, Xiao, Xu, Xu, Yu, and Zitkovich]{brohan2023rt1}
Anthony Brohan, Noah Brown, Justice Carbajal, Yevgen Chebotar, Joseph Dabis, Chelsea Finn, Keerthana Gopalakrishnan, Karol Hausman, Alex Herzog, Jasmine Hsu, Julian Ibarz, Brian Ichter, Alex Irpan, Tomas Jackson, Sally Jesmonth, Nikhil~J Joshi, Ryan Julian, Dmitry Kalashnikov, Yuheng Kuang, Isabel Leal, Kuang-Huei Lee, Sergey Levine, Yao Lu, Utsav Malla, Deeksha Manjunath, Igor Mordatch, Ofir Nachum, Carolina Parada, Jodilyn Peralta, Emily Perez, Karl Pertsch, Jornell Quiambao, Kanishka Rao, Michael Ryoo, Grecia Salazar, Pannag Sanketi, Kevin Sayed, Jaspiar Singh, Sumedh Sontakke, Austin Stone, Clayton Tan, Huong Tran, Vincent Vanhoucke, Steve Vega, Quan Vuong, Fei Xia, Ted Xiao, Peng Xu, Sichun Xu, Tianhe Yu, and Brianna Zitkovich.
\newblock Rt-1: Robotics transformer for real-world control at scale, 2023.

\bibitem[Brown et~al.(2020)Brown, Mann, Ryder, Subbiah, Kaplan, Dhariwal, Neelakantan, Shyam, Sastry, Askell, et~al.]{brown2020language}
Tom Brown, Benjamin Mann, Nick Ryder, Melanie Subbiah, Jared~D Kaplan, Prafulla Dhariwal, Arvind Neelakantan, Pranav Shyam, Girish Sastry, Amanda Askell, et~al.
\newblock Language models are few-shot learners.
\newblock \emph{Advances in neural information processing systems}, 33:\penalty0 1877--1901, 2020.

\bibitem[Caluwaerts et~al.(2023)Caluwaerts, Iscen, Kew, Yu, Zhang, Freeman, Lee, Lee, Saliceti, Zhuang, et~al.]{caluwaerts2023barkour}
Ken Caluwaerts, Atil Iscen, J~Chase Kew, Wenhao Yu, Tingnan Zhang, Daniel Freeman, Kuang-Huei Lee, Lisa Lee, Stefano Saliceti, Vincent Zhuang, et~al.
\newblock Barkour: Benchmarking animal-level agility with quadruped robots.
\newblock \emph{arXiv preprint arXiv:2305.14654}, 2023.

\bibitem[Chan et~al.(2022)Chan, Santoro, Lampinen, Wang, Singh, Richemond, McClelland, and Hill]{chan2022data}
Stephanie Chan, Adam Santoro, Andrew Lampinen, Jane Wang, Aaditya Singh, Pierre Richemond, James McClelland, and Felix Hill.
\newblock Data distributional properties drive emergent in-context learning in transformers.
\newblock \emph{Advances in Neural Information Processing Systems}, 35:\penalty0 18878--18891, 2022.

\bibitem[Chen et~al.(2021)Chen, Lu, Rajeswaran, Lee, Grover, Laskin, Abbeel, Srinivas, and Mordatch]{chen2021decision}
Lili Chen, Kevin Lu, Aravind Rajeswaran, Kimin Lee, Aditya Grover, Misha Laskin, Pieter Abbeel, Aravind Srinivas, and Igor Mordatch.
\newblock Decision transformer: Reinforcement learning via sequence modeling.
\newblock \emph{Advances in neural information processing systems}, 34:\penalty0 15084--15097, 2021.

\bibitem[Christiano et~al.(2023)Christiano, Leike, Brown, Martic, Legg, and Amodei]{christiano2023deep}
Paul Christiano, Jan Leike, Tom~B. Brown, Miljan Martic, Shane Legg, and Dario Amodei.
\newblock Deep reinforcement learning from human preferences, 2023.

\bibitem[Cideron et~al.(2019)Cideron, Seurin, Strub, and Pietquin]{cideron2019self}
Geoffrey Cideron, Mathieu Seurin, Florian Strub, and Olivier Pietquin.
\newblock Self-educated language agent with hindsight experience replay for instruction following.
\newblock \emph{DeepMind}, 2019.

\bibitem[Co-Reyes et~al.(2019)Co-Reyes, Gupta, Sanjeev, Altieri, DeNero, Abbeel, and Levine]{coreyes2019guiding}
John~D. Co-Reyes, Abhishek Gupta, Suvansh Sanjeev, Nick Altieri, John DeNero, Pieter Abbeel, and Sergey Levine.
\newblock Guiding policies with language via meta-learning.
\newblock In \emph{International Conference on Learning Representations (ICLR)}, 2019.

\bibitem[Cui et~al.(2023)Cui, Karamcheti, Palleti, Shivakumar, Liang, and Sadigh]{cui2023lilac}
Yuchen Cui, Siddharth Karamcheti, Raj Palleti, Nidhya Shivakumar, Percy Liang, and Dorsa Sadigh.
\newblock No, to the right -- online language corrections for robotic manipulation via shared autonomy.
\newblock In \emph{Proceedings of the 2023 ACM/IEEE Conference on Human-Robot Interaction (HRI)}, 2023.

\bibitem[Ding et~al.(2023)Ding, Zhang, Paxton, and Zhang]{ding2023task}
Yan Ding, Xiaohan Zhang, Chris Paxton, and Shiqi Zhang.
\newblock Task and motion planning with large language models for object rearrangement.
\newblock \emph{arXiv preprint arXiv:2303.06247}, 2023.

\bibitem[Driess et~al.(2023)Driess, Xia, Sajjadi, Lynch, Chowdhery, Ichter, Wahid, Tompson, Vuong, Yu, Huang, Chebotar, Sermanet, Duckworth, Levine, Vanhoucke, Hausman, Toussaint, Greff, Zeng, Mordatch, and Florence]{driess2023palme}
Danny Driess, Fei Xia, Mehdi S.~M. Sajjadi, Corey Lynch, Aakanksha Chowdhery, Brian Ichter, Ayzaan Wahid, Jonathan Tompson, Quan Vuong, Tianhe Yu, Wenlong Huang, Yevgen Chebotar, Pierre Sermanet, Daniel Duckworth, Sergey Levine, Vincent Vanhoucke, Karol Hausman, Marc Toussaint, Klaus Greff, Andy Zeng, Igor Mordatch, and Pete Florence.
\newblock Palm-e: An embodied multimodal language model, 2023.

\bibitem[Freitag and Al-Onaizan(2017)]{freitag2017beam}
Markus Freitag and Yaser Al-Onaizan.
\newblock Beam search strategies for neural machine translation.
\newblock \emph{arXiv preprint arXiv:1702.01806}, 2017.

\bibitem[Galy et~al.(2018)Galy, Paxion, and Berthelon]{galy2018measuring}
Edith Galy, Julie Paxion, and Catherine Berthelon.
\newblock Measuring mental workload with the nasa-tlx needs to examine each dimension rather than relying on the global score: an example with driving.
\newblock \emph{Ergonomics}, 61\penalty0 (4):\penalty0 517--527, 2018.

\bibitem[Goyal et~al.(2020)Goyal, Niekum, and Mooney]{goyal2020pixl2r}
Prasoon Goyal, Scott Niekum, and Raymond~J Mooney.
\newblock Pixl2r: Guiding reinforcement learning using natural language by mapping pixels to rewards.
\newblock \emph{arXiv:2007.15543}, 2020.

\bibitem[Gulcehre et~al.(2023)Gulcehre, Paine, Srinivasan, Konyushkova, Weerts, Sharma, Siddhant, Ahern, Wang, Gu, et~al.]{gulcehre2023reinforced}
Caglar Gulcehre, Tom~Le Paine, Srivatsan Srinivasan, Ksenia Konyushkova, Lotte Weerts, Abhishek Sharma, Aditya Siddhant, Alex Ahern, Miaosen Wang, Chenjie Gu, et~al.
\newblock Reinforced self-training (rest) for language modeling.
\newblock \emph{arXiv preprint arXiv:2308.08998}, 2023.

\bibitem[Hart(2006)]{hart2006TLX}
S.~G. Hart.
\newblock Nasa-task load index (nasa-tlx); 20 years later.
\newblock \emph{Proceedings of the 50th HFES Conference}, pages 904--908, 2006.

\bibitem[Herzog et~al.(2023)Herzog, Rao, Hausman, Lu, Wohlhart, Yan, Lin, Arenas, Xiao, Kappler, Ho, Rettinghouse, Chebotar, Lee, Gopalakrishnan, Julian, Li, Fu, Wei, Ramesh, Holden, Kleiven, Rendleman, Kirmani, Bingham, Weisz, Xu, Lu, Bennice, Fong, Do, Lam, Bai, Holson, Quinlan, Brown, Kalakrishnan, Ibarz, Pastor, and Levine]{herzog2023deep}
Alexander Herzog, Kanishka Rao, Karol Hausman, Yao Lu, Paul Wohlhart, Mengyuan Yan, Jessica Lin, Montserrat~Gonzalez Arenas, Ted Xiao, Daniel Kappler, Daniel Ho, Jarek Rettinghouse, Yevgen Chebotar, Kuang-Huei Lee, Keerthana Gopalakrishnan, Ryan Julian, Adrian Li, Chuyuan~Kelly Fu, Bob Wei, Sangeetha Ramesh, Khem Holden, Kim Kleiven, David Rendleman, Sean Kirmani, Jeff Bingham, Jon Weisz, Ying Xu, Wenlong Lu, Matthew Bennice, Cody Fong, David Do, Jessica Lam, Yunfei Bai, Benjie Holson, Michael Quinlan, Noah Brown, Mrinal Kalakrishnan, Julian Ibarz, Peter Pastor, and Sergey Levine.
\newblock Deep rl at scale: Sorting waste in office buildings with a fleet of mobile manipulators, 2023.

\bibitem[Honovich et~al.(2022)Honovich, Scialom, Levy, and Schick]{honovich2022unnatural}
Or~Honovich, Thomas Scialom, Omer Levy, and Timo Schick.
\newblock Unnatural instructions: Tuning language models with (almost) no human labor.
\newblock \emph{arXiv preprint arXiv:2212.09689}, 2022.

\bibitem[Hospedales et~al.(2021)Hospedales, Antoniou, Micaelli, and Storkey]{hospedales2021meta}
Timothy Hospedales, Antreas Antoniou, Paul Micaelli, and Amos Storkey.
\newblock Meta-learning in neural networks: A survey.
\newblock \emph{IEEE transactions on pattern analysis and machine intelligence}, 44\penalty0 (9):\penalty0 5149--5169, 2021.

\bibitem[Howell et~al.(2022)Howell, Gileadi, Tunyasuvunakool, Zakka, Erez, and Tassa]{howell2022predictive}
Taylor Howell, Nimrod Gileadi, Saran Tunyasuvunakool, Kevin Zakka, Tom Erez, and Yuval Tassa.
\newblock Predictive sampling: Real-time behaviour synthesis with mujoco.
\newblock \emph{arXiv preprint arXiv:2212.00541}, 2022.

\bibitem[Hu et~al.(2021)Hu, Shen, Wallis, Allen-Zhu, Li, Wang, Wang, and Chen]{hu2021lora}
Edward~J Hu, Yelong Shen, Phillip Wallis, Zeyuan Allen-Zhu, Yuanzhi Li, Shean Wang, Lu~Wang, and Weizhu Chen.
\newblock Lora: Low-rank adaptation of large language models.
\newblock \emph{arXiv preprint arXiv:2106.09685}, 2021.

\bibitem[Hu and Sadigh(2023)]{hu2023language}
Hengyuan Hu and Dorsa Sadigh.
\newblock Language instructed reinforcement learning for human-ai coordination.
\newblock In \emph{40th International Conference on Machine Learning (ICML)}, 2023.

\bibitem[Huang et~al.(2022{\natexlab{a}})Huang, Abbeel, Pathak, and Mordatch]{huang2022language}
Wenlong Huang, Pieter Abbeel, Deepak Pathak, and Igor Mordatch.
\newblock Language models as zero-shot planners: Extracting actionable knowledge for embodied agents.
\newblock In \emph{International Conference on Machine Learning}, pages 9118--9147. PMLR, 2022{\natexlab{a}}.

\bibitem[Huang et~al.(2022{\natexlab{b}})Huang, Xia, Xiao, Chan, Liang, Florence, Zeng, Tompson, Mordatch, Chebotar, et~al.]{huang2022inner}
Wenlong Huang, Fei Xia, Ted Xiao, Harris Chan, Jacky Liang, Pete Florence, Andy Zeng, Jonathan Tompson, Igor Mordatch, Yevgen Chebotar, et~al.
\newblock Inner monologue: Embodied reasoning through planning with language models.
\newblock \emph{arXiv preprint arXiv:2207.05608}, 2022{\natexlab{b}}.

\bibitem[Hupkes et~al.(2020)Hupkes, Dankers, Mul, and Bruni]{hupkes2020compositionality}
Dieuwke Hupkes, Verna Dankers, Mathijs Mul, and Elia Bruni.
\newblock Compositionality decomposed: How do neural networks generalise?
\newblock \emph{Journal of Artificial Intelligence Research}, 67:\penalty0 757--795, 2020.

\bibitem[Jang et~al.(2022)Jang, Irpan, Khansari, Kappler, Ebert, Lynch, Levine, and Finn]{jang2022bcz}
Eric Jang, Alex Irpan, Mohi Khansari, Daniel Kappler, Frederik Ebert, Corey Lynch, Sergey Levine, and Chelsea Finn.
\newblock Bc-z: Zero-shot task generalization with robotic imitation learning, 2022.

\bibitem[Jiang et~al.(2019)Jiang, Gu, Murphy, and Finn]{jiang2019language}
Yiding Jiang, Shixiang~Shane Gu, Kevin~P Murphy, and Chelsea Finn.
\newblock Language as an abstraction for hierarchical deep reinforcement learning.
\newblock \emph{NeurIPS}, 2019.

\bibitem[Karamcheti et~al.(2017)Karamcheti, Williams, Arumugam, Rhee, Gopalan, Wong, and Tellex]{karamcheti2017draggns}
Siddharth Karamcheti, Edward~C. Williams, Dilip Arumugam, Mina Rhee, Nakul Gopalan, Lawson L.~S. Wong, and Stefanie Tellex.
\newblock A tale of two draggns: A hybrid approach for interpreting action-oriented and goal-oriented instructions.
\newblock In \emph{First Workshop on Language Grounding for Robotics @ ACL}, 2017.

\bibitem[Kirillov et~al.(2023)Kirillov, Mintun, Ravi, Mao, Rolland, Gustafson, Xiao, Whitehead, Berg, Lo, et~al.]{kirillov2023segment}
Alexander Kirillov, Eric Mintun, Nikhila Ravi, Hanzi Mao, Chloe Rolland, Laura Gustafson, Tete Xiao, Spencer Whitehead, Alexander~C Berg, Wan-Yen Lo, et~al.
\newblock Segment anything.
\newblock \emph{arXiv preprint arXiv:2304.02643}, 2023.

\bibitem[Kollar et~al.(2010)Kollar, Tellex, Roy, and Roy]{kollar10directions}
Thomas Kollar, Stefanie Tellex, Deb Roy, and Nicholas Roy.
\newblock Toward understanding natural language directions.
\newblock In \emph{Human-Robot Interaction}, pages 259--266, 2010.

\bibitem[Kwon et~al.(2023)Kwon, Xie, Bullard, and Sadigh]{kwon2023reward}
Minae Kwon, Sang~Michael Xie, Kalesha Bullard, and Dorsa Sadigh.
\newblock Reward design with language models.
\newblock In \emph{International Conference on Learning Representations (ICLR)}, 2023.

\bibitem[Kwon et~al.(2024)Kwon, Hu, Myers, Karamcheti, Dragan, and Sadigh]{kwon2023toward}
Minae Kwon, Hengyuan Hu, Vivek Myers, Siddharth Karamcheti, Anca Dragan, and Dorsa Sadigh.
\newblock Toward grounded commonsense reasoning.
\newblock In \emph{2024 IEEE International Conference on Robotics and Automation (ICRA)}, 2024.
\newblock URL \url{arXiv preprint arXiv:2306.08651}.

\bibitem[Lewis et~al.(2020)Lewis, Perez, Piktus, Petroni, Karpukhin, Goyal, K{\"u}ttler, Lewis, Yih, Rockt{\"a}schel, et~al.]{lewis2020retrieval}
Patrick Lewis, Ethan Perez, Aleksandra Piktus, Fabio Petroni, Vladimir Karpukhin, Naman Goyal, Heinrich K{\"u}ttler, Mike Lewis, Wen-tau Yih, Tim Rockt{\"a}schel, et~al.
\newblock Retrieval-augmented generation for knowledge-intensive nlp tasks.
\newblock \emph{Advances in Neural Information Processing Systems}, 33:\penalty0 9459--9474, 2020.

\bibitem[Liang et~al.(2023)Liang, Huang, Xia, Xu, Hausman, Ichter, Florence, and Zeng]{liang2023code}
Jacky Liang, Wenlong Huang, Fei Xia, Peng Xu, Karol Hausman, Brian Ichter, Pete Florence, and Andy Zeng.
\newblock Code as policies: Language model programs for embodied control.
\newblock In \emph{2023 IEEE International Conference on Robotics and Automation (ICRA)}, pages 9493--9500. IEEE, 2023.

\bibitem[Liu et~al.(2023)Liu, Jiang, Zhang, Liu, Zhang, Biswas, and Stone]{liu2023llm+}
Bo~Liu, Yuqian Jiang, Xiaohan Zhang, Qiang Liu, Shiqi Zhang, Joydeep Biswas, and Peter Stone.
\newblock {LLM+P}: Empowering large language models with optimal planning proficiency.
\newblock \emph{arXiv preprint arXiv:2304.11477}, 2023.

\bibitem[Luketina et~al.(2019)Luketina, Nardelli, Farquhar, Foerster, Andreas, Grefenstette, Whiteson, and Rocktäschel]{luketina2019survey}
Jelena Luketina, Nantas Nardelli, Gregory Farquhar, Jakob Foerster, Jacob Andreas, Edward Grefenstette, Shimon Whiteson, and Tim Rocktäschel.
\newblock A survey of reinforcement learning informed by natural language, 2019.

\bibitem[Lynch and Sermanet(2021)]{lynch2021language}
Corey Lynch and Pierre Sermanet.
\newblock Language conditioned imitation learning over unstructured data, 2021.

\bibitem[Ma et~al.(2023)Ma, Liang, Wang, Huang, Bastani, Jayaraman, Zhu, Fan, and Anandkumar]{ma2023eureka}
Yecheng~Jason Ma, William Liang, Guanzhi Wang, De-An Huang, Osbert Bastani, Dinesh Jayaraman, Yuke Zhu, Linxi Fan, and Anima Anandkumar.
\newblock Eureka: Human-level reward design via coding large language models.
\newblock \emph{arXiv preprint arXiv:2310.12931}, 2023.

\bibitem[Matuszek et~al.(2012)Matuszek, Herbst, Zettlemoyer, and Fox]{matuszek2012learning}
C.~Matuszek, E.~Herbst, L.~Zettlemoyer, and D.~Fox.
\newblock Learning to parse natural language commands to a robot control system.
\newblock In \emph{International Symposium on Experimental Robotics (ISER)}, 2012.

\bibitem[Mees et~al.(2022)Mees, Hermann, Rosete-Beas, and Burgard]{mees2022calvin}
Oier Mees, Lukas Hermann, Erick Rosete-Beas, and Wolfram Burgard.
\newblock Calvin: A benchmark for language-conditioned policy learning for long-horizon robot manipulation tasks.
\newblock \emph{IEEE Robotics and Automation Letters}, 7\penalty0 (3):\penalty0 7327--7334, 2022.

\bibitem[Mirchandani et~al.(2021)Mirchandani, Karamcheti, and Sadigh]{mirchandani2021ella}
Suvir Mirchandani, Siddharth Karamcheti, and Dorsa Sadigh.
\newblock Ella: {{Exploration}} through learned language abstraction, October 2021.

\bibitem[Mirchandani et~al.(2023)Mirchandani, Xia, Florence, Ichter, Driess, Arenas, Rao, Sadigh, and Zeng]{mirchandani2023large}
Suvir Mirchandani, Fei Xia, Pete Florence, Brian Ichter, Danny Driess, Montserrat~Gonzalez Arenas, Kanishka Rao, Dorsa Sadigh, and Andy Zeng.
\newblock Large language models as general pattern machines.
\newblock \emph{arXiv preprint arXiv:2307.04721}, 2023.

\bibitem[Misra et~al.(2017)Misra, Langford, and Artzi]{misra2017mapping}
Dipendra Misra, John Langford, and Yoav Artzi.
\newblock Mapping instructions and visual observations to actions with reinforcement learning.
\newblock \emph{arXiv:1704.08795}, 2017.

\bibitem[Ouyang et~al.(2022)Ouyang, Wu, Jiang, Almeida, Wainwright, Mishkin, Zhang, Agarwal, Slama, Ray, et~al.]{ouyang2022training}
Long Ouyang, Jeffrey Wu, Xu~Jiang, Diogo Almeida, Carroll Wainwright, Pamela Mishkin, Chong Zhang, Sandhini Agarwal, Katarina Slama, Alex Ray, et~al.
\newblock Training language models to follow instructions with human feedback.
\newblock \emph{Advances in Neural Information Processing Systems}, 35:\penalty0 27730--27744, 2022.

\bibitem[Radford et~al.(2018)Radford, Wu, Child, Luan, Amodei, and Sutskever]{LLMs-are-multitask-learners}
Alec Radford, Jeffrey Wu, Rewon Child, David Luan, Dario Amodei, and Ilya Sutskever.
\newblock Language models are unsupervised multitask learners.
\newblock 2018.
\newblock URL \url{https://d4mucfpksywv.cloudfront.net/better-language-models/language-models.pdf}.

\bibitem[Raffel et~al.(2020)Raffel, Shazeer, Roberts, Lee, Narang, Matena, Zhou, Li, and Liu]{raffel2020exploring}
Colin Raffel, Noam Shazeer, Adam Roberts, Katherine Lee, Sharan Narang, Michael Matena, Yanqi Zhou, Wei Li, and Peter~J Liu.
\newblock Exploring the limits of transfer learning with a unified text-to-text transformer.
\newblock \emph{The Journal of Machine Learning Research}, 21\penalty0 (1):\penalty0 5485--5551, 2020.

\bibitem[Ren et~al.(2023)Ren, Dixit, Bodrova, Singh, Tu, Brown, Xu, Takayama, Xia, Varley, Xu, Sadigh, Zeng, and Majumdar]{ren2023robots}
Allen~Z. Ren, Anushri Dixit, Alexandra Bodrova, Sumeet Singh, Stephen Tu, Noah Brown, Peng Xu, Leila Takayama, Fei Xia, Jake Varley, Zhenjia Xu, Dorsa Sadigh, Andy Zeng, and Anirudha Majumdar.
\newblock Robots that ask for help: Uncertainty alignment for large language model planners, 2023.

\bibitem[Sha et~al.(2023)Sha, Mu, Jiang, Chen, Xu, Luo, Li, Tomizuka, Zhan, and Ding]{sha2023languagempc}
Hao Sha, Yao Mu, Yuxuan Jiang, Li~Chen, Chenfeng Xu, Ping Luo, Shengbo~Eben Li, Masayoshi Tomizuka, Wei Zhan, and Mingyu Ding.
\newblock Languagempc: Large language models as decision makers for autonomous driving, 2023.

\bibitem[Sharma et~al.(2022)Sharma, Sundaralingam, Blukis, Paxton, Hermans, Torralba, Andreas, and Fox]{sharma2022correcting}
Pratyusha Sharma, Balakumar Sundaralingam, Valts Blukis, Chris Paxton, Tucker Hermans, Antonio Torralba, Jacob Andreas, and Dieter Fox.
\newblock Correcting robot plans with natural language feedback.
\newblock \emph{arXiv preprint arXiv:2204.05186}, 2022.

\bibitem[Shepard and Chang(1963)]{shepard1963stimulus}
Roger~N Shepard and Jih-Jie Chang.
\newblock Stimulus generalization in the learning of classifications.
\newblock \emph{Journal of Experimental Psychology}, 65\penalty0 (1):\penalty0 94, 1963.

\bibitem[Shridhar et~al.(2021)Shridhar, Manuelli, and Fox]{shridhar2022cliport}
Mohit Shridhar, Lucas Manuelli, and Dieter Fox.
\newblock Cliport: What and where pathways for robotic manipulation.
\newblock In \emph{CoRL}, 2021.

\bibitem[Singh et~al.(2023)Singh, Blukis, Mousavian, Goyal, Xu, Tremblay, Fox, Thomason, and Garg]{singh2023progprompt}
Ishika Singh, Valts Blukis, Arsalan Mousavian, Ankit Goyal, Danfei Xu, Jonathan Tremblay, Dieter Fox, Jesse Thomason, and Animesh Garg.
\newblock Progprompt: Generating situated robot task plans using large language models.
\newblock In \emph{2023 IEEE International Conference on Robotics and Automation (ICRA)}, pages 11523--11530. IEEE, 2023.

\bibitem[Stepputtis et~al.(2020)Stepputtis, Campbell, Phielipp, Lee, Baral, and Ben~Amor]{stepputtis2020language}
Simon Stepputtis, Joseph Campbell, Mariano Phielipp, Stefan Lee, Chitta Baral, and Heni Ben~Amor.
\newblock Language-conditioned imitation learning for robot manipulation tasks.
\newblock \emph{NeurIPS}, 2020.

\bibitem[Stiennon et~al.(2022)Stiennon, Ouyang, Wu, Ziegler, Lowe, Voss, Radford, Amodei, and Christiano]{stiennon2022learning}
Nisan Stiennon, Long Ouyang, Jeff Wu, Daniel~M. Ziegler, Ryan Lowe, Chelsea Voss, Alec Radford, Dario Amodei, and Paul Christiano.
\newblock Learning to summarize from human feedback, 2022.

\bibitem[Talmor et~al.(2022)Talmor, Yoran, Bras, Bhagavatula, Goldberg, Choi, and Berant]{talmor2022commonsenseqa}
Alon Talmor, Ori Yoran, Ronan~Le Bras, Chandra Bhagavatula, Yoav Goldberg, Yejin Choi, and Jonathan Berant.
\newblock {{CommonsenseQA}} 2.0: {{Exposing}} the limits of {{AI}} through gamification.
\newblock \emph{arXiv preprint arXiv:2201.05320}, 2022.

\bibitem[Tellex et~al.(2011)Tellex, Kollar, Dickerson, Walter, Banerjee, Teller, and Roy]{tellex2011understanding}
Stefanie Tellex, Thomas Kollar, Steven Dickerson, Matthew Walter, Ashis Banerjee, Seth Teller, and Nicholas Roy.
\newblock Understanding natural language commands for robotic navigation and mobile manipulation.
\newblock In \emph{AAAI}, 2011.

\bibitem[Tellex et~al.(2020)Tellex, Gopalan, Kress-Gazit, and Matuszek]{2020tellex-robots-use-language}
Stefanie Tellex, Nakul Gopalan, Hadas Kress-Gazit, and Cynthia Matuszek.
\newblock Robots that use language.
\newblock \emph{Annual Review of Control, Robotics, and Autonomous Systems}, 3\penalty0 (1):\penalty0 25--55, 2020.
\newblock \doi{10.1146/annurev-control-101119-071628}.
\newblock URL \url{https://doi.org/10.1146/annurev-control-101119-071628}.

\bibitem[Todorov et~al.(2012)Todorov, Erez, and Tassa]{todorov2012mujoco}
Emanuel Todorov, Tom Erez, and Yuval Tassa.
\newblock Mujoco: A physics engine for model-based control.
\newblock In \emph{2012 IEEE/RSJ International Conference on Intelligent Robots and Systems}, pages 5026--5033. IEEE, 2012.
\newblock \doi{10.1109/IROS.2012.6386109}.

\bibitem[Vaswani et~al.(2017)Vaswani, Shazeer, Parmar, Uszkoreit, Jones, Gomez, Kaiser, and Polosukhin]{vaswani2017attention}
Ashish Vaswani, Noam Shazeer, Niki Parmar, Jakob Uszkoreit, Llion Jones, Aidan~N Gomez, {\L}ukasz Kaiser, and Illia Polosukhin.
\newblock Attention is all you need.
\newblock \emph{Advances in neural information processing systems}, 30, 2017.

\bibitem[Wei et~al.(2022)Wei, Wang, Schuurmans, Bosma, Xia, Chi, Le, Zhou, et~al.]{wei2022chain}
Jason Wei, Xuezhi Wang, Dale Schuurmans, Maarten Bosma, Fei Xia, Ed~Chi, Quoc~V Le, Denny Zhou, et~al.
\newblock Chain-of-thought prompting elicits reasoning in large language models.
\newblock \emph{Advances in Neural Information Processing Systems}, 35:\penalty0 24824--24837, 2022.

\bibitem[Wu et~al.(2023)Wu, Antonova, Kan, Lepert, Zeng, Song, Bohg, Rusinkiewicz, and Funkhouser]{wu2023tidybot}
Jimmy Wu, Rika Antonova, Adam Kan, Marion Lepert, Andy Zeng, Shuran Song, Jeannette Bohg, Szymon Rusinkiewicz, and Thomas Funkhouser.
\newblock Tidybot: Personalized robot assistance with large language models.
\newblock \emph{arXiv preprint arXiv:2305.05658}, 2023.

\bibitem[Xie et~al.(2023)Xie, Yu, Zhu, Bai, Gong, and Soh]{xie2023translating}
Yaqi Xie, Chen Yu, Tongyao Zhu, Jinbin Bai, Ze~Gong, and Harold Soh.
\newblock Translating natural language to planning goals with large-language models.
\newblock \emph{arXiv preprint arXiv:2302.05128}, 2023.

\bibitem[Yoneda et~al.(2023)Yoneda, Fang, Li, Zhang, Jiang, Lin, Picker, Yunis, Mei, and Walter]{yoneda2023statler}
Takuma Yoneda, Jiading Fang, Peng Li, Huanyu Zhang, Tianchong Jiang, Shengjie Lin, Ben Picker, David Yunis, Hongyuan Mei, and Matthew~R. Walter.
\newblock Statler: State-maintaining language models for embodied reasoning, 2023.

\bibitem[Yu et~al.(2023)Yu, Gileadi, Fu, Kirmani, Lee, Arenas, Chiang, Erez, Hasenclever, Humplik, et~al.]{yu2023language}
Wenhao Yu, Nimrod Gileadi, Chuyuan Fu, Sean Kirmani, Kuang-Huei Lee, Montse~Gonzalez Arenas, Hao-Tien~Lewis Chiang, Tom Erez, Leonard Hasenclever, Jan Humplik, et~al.
\newblock Language to rewards for robotic skill synthesis.
\newblock \emph{arXiv preprint arXiv:2306.08647}, 2023.

\bibitem[Yuan et~al.(2024)Yuan, Pang, Cho, Sukhbaatar, Xu, and Weston]{yuan2024self}
Weizhe Yuan, Richard~Yuanzhe Pang, Kyunghyun Cho, Sainbayar Sukhbaatar, Jing Xu, and Jason Weston.
\newblock Self-rewarding language models.
\newblock \emph{arXiv preprint arXiv:2401.10020}, 2024.

\bibitem[Zelikman et~al.(2023)Zelikman, Huang, Poesia, Goodman, and Haber]{zelikman2023parsel}
Eric Zelikman, Qian Huang, Gabriel Poesia, Noah~D Goodman, and Nick Haber.
\newblock Parsel: A (de-) compositional framework for algorithmic reasoning with language models.
\newblock \emph{arXiv preprint arXiv:2212.10561}, 2023.

\bibitem[Zeng et~al.(2022)Zeng, Attarian, Ichter, Choromanski, Wong, Welker, Tombari, Purohit, Ryoo, Sindhwani, Lee, Vanhoucke, and Florence]{zeng2022socratic}
Andy Zeng, Maria Attarian, Brian Ichter, Krzysztof Choromanski, Adrian Wong, Stefan Welker, Federico Tombari, Aveek Purohit, Michael Ryoo, Vikas Sindhwani, Johnny Lee, Vincent Vanhoucke, and Pete Florence.
\newblock Socratic models: Composing zero-shot multimodal reasoning with language.
\newblock \emph{arXiv preprint arXiv:2204.00598}, 2022.

\bibitem[Zha et~al.(2024)Zha, Cui, Lin, Kwon, Arenas, Zeng, Xia, and Sadigh]{zha2023distilling}
Lihan Zha, Yuchen Cui, Li-Heng Lin, Minae Kwon, Montserrat~G. Arenas, Andy Zeng, Fei Xia, and Dorsa Sadigh.
\newblock Distilling and retrieving generalizable knowledge for robot manipulation via language corrections.
\newblock In \emph{2024 IEEE International Conference on Robotics and Automation (ICRA)}, 2024.
\newblock URL \url{https://arxiv.org/abs/2311.10678}.

\bibitem[Zhao et~al.(2023)Zhao, Kumar, Levine, and Finn]{zhao2023learning}
Tony~Z Zhao, Vikash Kumar, Sergey Levine, and Chelsea Finn.
\newblock Learning fine-grained bimanual manipulation with low-cost hardware.
\newblock \emph{arXiv preprint arXiv:2304.13705}, 2023.

\end{thebibliography}

\clearpage
\normalsize
\section*{Authorship and Acknowledgments}

\noindent\textbf{Acknowledgements.}
We thank John Guilyard for his expert animations, Giles Ruscoe for beautiful renderings, and Anirudha Majumdar for help on writing. We thank Steven Bohez, Yuval Tassa, Tom Erez, Murilo Martins, Rugile Pevceviciute, David Rendleman, and Connor Schenck for their dedication to ensuring we had strong simulated environments. We thank Travis Armstrong, Noah Brown, Spencer Goodrich, Craig Hickman, Atil Iscen, Jerad Kirkland, Jason Powell, Stefano Saliceti, Ron Sloat, Sergey Yaroshenko, Eddie Yu, Grace Vesom, and Jake Varley for additional robot platform support and robot lab operations. Special thanks to Michael Ahn, Kendra Byrne, Aleksandra Faust, René Wagner, Yuheng Kuang, Yao Lu, Yansong Pang, and Zhuo Xu for supporting this project.

We thank all the users who volunteered to collect the robot teaching data. We also thank the Google DeepMind Visualization and Human Interaction teams for their help with the development and support of the chat interface. We also want to thank the entire Google DeepMind Robotics team whose tireless efforts can be traced to additional support on this paper. This includes Administrative, Product, Programs, and Strategy teams whose contributions impact all of the team's successes. We also want to thank our friends in Google DeepMind and Google Research for their guidance, inspirational research, and even direct contributions.

\noindent\textbf{Program Leads}

\noindent This project is part of the Google DeepMind 2023 program "ApprenticeBots," an interactive embodied AI moonshot with the mission statement: "anyone can teach a robot, and a robot that can learn from anyone."

\noindent Carolina Parada, \textit{Director}

\noindent Nik Stewart, \textit{Technical Program Manager}

\noindent Jie Tan, \textit{Team Lead}

\noindent\textbf{Technical Leads}

\noindent Andy Zeng, \textit{Research Lead}

\noindent Wenhao Yu, Fei Xia,	\textit{Data Collection \& Teaching Leads}

\noindent Jacky Liang, \textit{Model Training \& Improvement Lead}

\noindent Jasmine Hsu, \textit{Data \& Logging Lead}

\noindent Peng Xu, \textit{Infrastructure Lead}

\noindent Ben Jyenis, \textit{Operations Lead}

\noindent Erik Frey, \textit{Simulation Lead}

\noindent\textbf{Operations}

\noindent Ben Jyenis, Travis Armstrong, \textit{Head of operations}

\noindent Jasmine Hsu, Jacky Liang, \textit{Data collection monitoring}

\noindent Wenhao Yu, \textit{Pilot studies for Robot Dog}

\noindent Fei Xia, \textit{Pilot studies for Mobile Manipulator}

\noindent Baruch Tabanpour, \textit{Pilot studies for Aloha}

\noindent Maria Attarian, Jonathan Tompson, \textit{Pilot studies for Bi-arm Kuka}

\noindent Joss Moore, Maria Bauza, \textit{Pilot studies for Kuka+Hand}

\noindent \textit{Contributors}: Maria Attarian, Ken Caluwaerts, Jasmine Hsu, Jacky Liang, Assaf Hurwitz Michaely, Jonathan Tompson, Fei Xia, Wenhao Yu, Andy Zeng, Tingnan Zhang

\noindent\textbf{Data Logging Infrastructure}

\noindent Jasmine Hsu, Ken Caluwaerts, \textit{Datasets and dashboards}

\noindent Peng Xu, Assaf Hurwitz Michaely, Jacky Liang, \textit{Materialization}

\noindent \textit{Contributors}: Adil Dostmohamed, Marissa Giustina, Nikhil Joshi, Jacky Liang, Quan Vuong, Tingnan Zhang

\noindent\textbf{Model Serving Infrastructure}

\noindent Assaf Hurwitz Michaely, Ying Xu, \textit{Core contributors}

\noindent Jasmine Hsu, Ken Caluwaerts, Adil Dostmohamed, \textit{LLM Chat UI}

\noindent\textit{Contributors}:
\noindent Jacky Liang, Allen Ren, Andy Zeng, Tingnan Zhang

\noindent\textbf{Model Training Infrastructure}

\noindent\textit{Core contributors}:
Assaf Hurwitz Michaely, Jacky Liang, Peng Xu, Andy Zeng, Jasmine Hsu, Edward Lee 

\noindent\textit{Contributors}:
Quan Vuong, Tingnan Zhang

\noindent\textbf{Evaluations \& Analysis}

\noindent Jacky Liang, \textit{Technical Lead}

\noindent Leila Takayama, \textit{Human-Robot Interaction Lead}

\noindent\textit{Contributors}:
Alex Bewley, Keerthana Gopalakrishnan, Jasmine Hsu, Jacky Liang, Assaf Hurwitz Michaely, Dorsa Sadigh, Fei Xia, Ted Xiao, Andy Zeng, Tingnan Zhang

\noindent\textbf{Prompt Engineering}

\noindent Maria Bauza, Marissa Giustina, Kuang-Huei Lee, Jacky Liang, Joss Moore, Dushyant Rao, Baruch Tabanpour, Fei Xia, Wenhao Yu, Andy Zeng

\noindent\textbf{Simulation \& MJPC}

\noindent Maria Attarian, Ken Caluwaerts, Erik Frey, Chuyuan Kelly Fu, Nimrod Gileadi, Leonard Hasenclever, Jan Humplik, Nikhil Joshi, Ben Jyenis, Joss Moore, Dushyant Rao, Baruch Tabanpour, Fei Xia, Ted Xiao, Wenhao Yu, Tingnan Zhang

\noindent\textbf{Robot-Specific Infrastructure}

\noindent\textit{Robot Dog}: Ken Caluwaerts, Marissa Giustina, Chase Kew, Ken Oslund, Wenhao Yu Tingnan Zhang,

\noindent\textit{Mobile Manipulator}: Fei Xia, Chuyuan Kelly Fu

\noindent\textit{Aloha}: Baruch Tabanpour, Jonathan Tompson, Erik Frey

\noindent\textit{Bi-arm Kuka}: Maria Attarian

\noindent\textit{Kuka+Hand}: Maria Bauza, Joss Moore, Dushyant Rao, Nimrod Gileadi

\noindent\textbf{Real Robot Deployment \& Policy Distillation}

\noindent Ken Caluwaerts, Chuyuan Kelly Fu, Leonard Hasenclever, Jan Humplik, Chase Kew, Sean Kirmani, Kuang-Huei Lee, Ken Oslund, Allen Ren, Jonathan Tompson, Quan Vuong, Fei Xia, Ted Xiao, Zhuo Xu, Wenhao Yu, Tingnan Zhang

\noindent\textbf{Advising}

\noindent Alex Bewley, Erik Frey, Leonard Hasenclever, Jasmine Hsu, Jan Humplik, Brian Ichter, Kuang-Huei Lee, Jacky Liang, Carolina Parada, Dushyant Rao, Dorsa Sadigh, Nik Stewart, Leila Takayama, Jie Tan, Fei Xia, Ted Xiao, Peng Xu, Wenhao Yu, Andy Zeng, Tingnan Zhang

\noindent\textbf{Additional Contributions}

\noindent\textit{Authorship and Acknowledgments}: Nik Stewart				

\noindent\textit{Paper Content and Web Posts}: Carolina Parada, Andy Zeng, Wenhao Yu, Jacky Liang, Fei Xia, Tingnan Zhang

\noindent\textit{Steering}: Carolina Parada, Nik Stewart, Izhak Shafran, Vincent Vanhoucke, Maja Mataric, Leila Takayama, Jie Tan, Dorsa Sadigh, Andy Zeng, Wenhao Yu, Jacky Liang, Fei Xia, Tingnan Zhang

\clearpage
\section{Appendix}

We organize the appendix as follows:
\begin{itemize}[leftmargin=11pt]
    \item Details on data collection (\eg chat UI), and evaluation protocol (\eg task sampling). \cref{app:data_collect}
    \item Additional results and evaluations in \cref{app:additional_results}.
    \item Details on top-users conditioning (\cref{app:top_users}): how they are autonomously selected, quantitative and qualitative analysis on how top-user teaching data differs from other users.
    \item Retrieval baseline details (\cref{app:rag}) and data augmentation (\cref{app:data_augmentation}).
    \item Quantitative analysis of chat feedback embeddings (\cref{app:feedback_embeddings}).
    \item Real robot experiments details (\cref{app:real_robots}) including model training and deployment (\cref{app:finetune}).
    \item User studies and performance drift \cref{app:user_drift}.
    \item Failure mode analysis \cref{app:failure_modes}.
    \item Model performance on existing code-writing benchmarks \cref{app:coding_benchmarks}.
    \item Robot-specific embodiment details (\cref{app:MJPC}), tasks (\cref{app:tasks}), and prompts (\cref{app:prompts}).
\end{itemize}

\subsection{Data Collection and Evaluation Details}
\label{app:data_collect}

During data collection, non-expert users interact with the robot using natural language through a browser-based chat UI (shown in \cref{fig:data_collection}). 
%
The chat UI displays a user input box, the message history, and a visualization of the simulated robot and its surroundings using MuJoCo \cite{todorov2012mujoco}.
The human provides textual input and the LLM replies to each subsequent user query with executable code. 
The user can then select a button to either run the code in the simulator to observe the resulting motion, or run it on a real robot. 
The user can continue to provide feedback (which can be multi-turn contextual) and continue modifying the behavior through text inputs in the chat UI until the desired robot behavior is achieved. Each user is remotely connected (via Remote Desktop) to one machine, drawn from a shared pool of high performance machines (128 cores) in the cloud. Machines with high core counts are necessary for Mujoco's MJPC~\cite{howell2022predictive} to synthesize robot motion at an interactive rate -- leading to better low-level robot behaviors and subsequently user feedback data.

For each chat session, the user teaches the robot one task specified via language \eg teach the robot-dog to ``sit down and give a high-five.''
For each chat turn (user-input, LLM-output pair), the user has the option to rate the individual robot response as `good' or `bad'. 
These single turn good rating rates (while not used during training) help us evaluate responsiveness to feedback across individual responses, and we find they are strongly correlated to task success (see~\cref{fig:data_collection}).
Finally, users can label the entire chat session as ``success'' by clicking a success button if the robot succeeded at the task during the conversation, or ``failure'' if the task does not succeed within 7 rounds of human input. 
Variation in success is expected, and users are encouraged to rate success based on the observed behaviors of the robot (as opposed to the accuracy of the code).
After labeling a chat session, the chat history UI refreshes, the robot simulator is reset, and a new sampled task and embodiment is presented to the user.
Users are able to flag chat sessions in case of technical difficulties.

Our UI backend uses a Task Sampler, which is configured to (i) randomly sample tasks from the set of 78 tasks across 5 embodiments (platforms illustrated in~\cref{fig:teaser}), and (ii) randomly sample an LLM model to connect to. Users do not know which model they speak to, which allows us to perform fair blind A/B evaluations. All experiment numbers are computed with data collected using this sampler.


From the perspective of users, our data collection protocol is equivalent to our evaluation protocol -- we train on data collected from users interacting with the model(s) through the UI, and we measure whether users believe the model(s) to have improved (via statistics on good rating rates and session success labels) through the UI with blind A/B evaluations. This deviates from the standard norm in robot learning pipelines (\eg 3-stage pipeline of collect data, train, and evaluate), and presents practical infrastructure/operations advantages (predominantly around simplicity).

To operationalize data collection, we started off with running multiple pilot sessions with the users for each embodiment. These pilot sessions were focused on introducing these 35 non-expert users to the the chat UI, the type of tasks they are expected to teach, and the MuJoCo simulation environment.
After the pilot sessions conclude, the users were tasked to contribute 10 chat sessions per day on all embodiments through the task sampler, amounting to 350 chat sessions every day.
The users were also asked to fill out a brief questionnaire for a feedback after each day about their experience on the data collection and the overall teaching session. 
To meet the daily target of 350 chat sessions per day, it was important to maintain participation from all of the users equally to obtain the expected level of diversity in the data. We maintained consistent distribution of number of chat sessions across multiple embodiments i.e. Robot Dog, Mobile Manipulator, Aloha, Bi-arm Kuka, Kuka+Hand. 

The users who participated in these experiments were 23-43 years of age (\textit{M}=30.5, \textit{SD}=5.6), including 11 who identified as cisgender women, 17 who identified as cisgender men, and 1 as non-binary. They had a range of educational degrees (9 Associates degrees or some college, 6 Bachelor of Arts, 11 Bachelor of Sciences, and 3 Masters degrees) -- 14 non-technical and 15 technical. When asked about their familiarity with the ML models on a scale of 1 (zero familiarity) to 5 (most familiar), 15 users reported 1 (zero familiarity), 11 users reported 2, and 3 users reported 3; none of the users reported to have more familiarity with ML models (4 or 5).

\begin{figure}[!th]
    \vspace{-1em}
    \centering
    \includegraphics[width=0.6\linewidth]{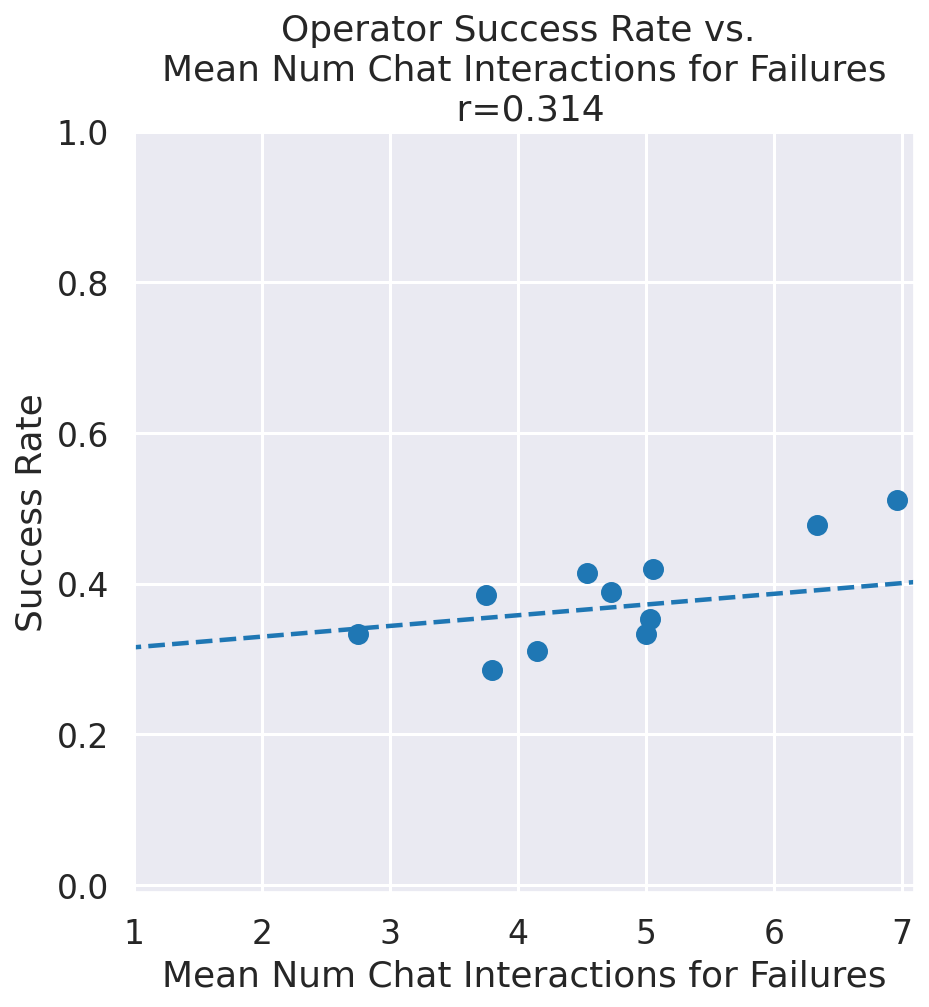}
    \caption{Correlation of Operator Success Rate and Num Chat Turns until Failures}
    \label{fig:op_success_rate_nct_failures}
\end{figure}

\textbf{User Persistence Analysis.}
We plot the success rate of each user against the mean number of chat turns in failed sessions across each user in~\cref{fig:op_success_rate_nct_failures}.
The higher the mean number of chat turns for failure, the more persistant the user was in teaching the robot (i.e. the user did not give up early).
These two quantities exhibit a slight positive correlation, suggesting that on average, users who were more persistant at teaching achieved slightly higher success rates.

\subsection{Additional Results}
\label{app:additional_results}

\begin{figure*}[!t]
    \centering
    \includegraphics[width=0.9\linewidth]{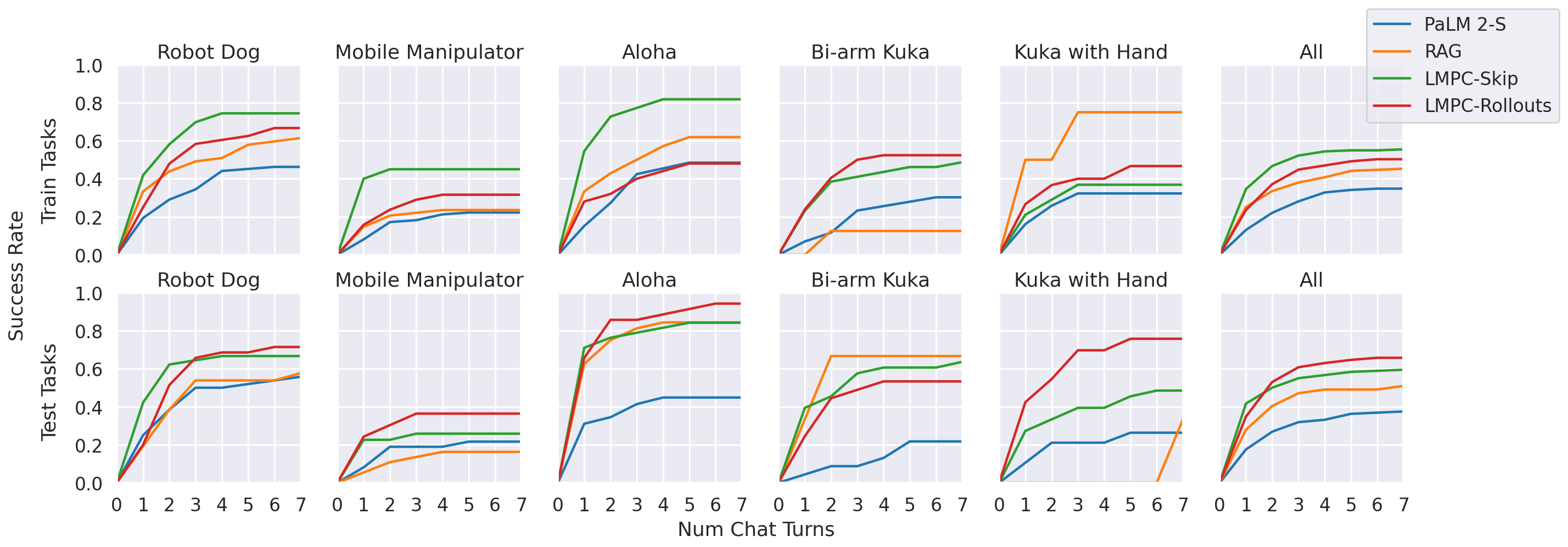}
    \caption{Task Success vs. Number of Chat Turns. across embodiments}
    \label{fig:task_success_nci_embodiment}
    \vspace{-1em}
\end{figure*}

\textbf{Per Embodiment Evaluation.}
\cref{fig:task_success_nci_embodiment} shows our main teachability result (\cref{fig:task_success_nci_all}) separated by embodiments.
On test tasks, models improved upon the base model the most in Aloha and Bi-arm Kuka, while LMPC-Rollouts improved much higher on Kuka with Hand than the other model.

\begin{figure}[th]
    \centering
    \includegraphics[width=0.49\linewidth]{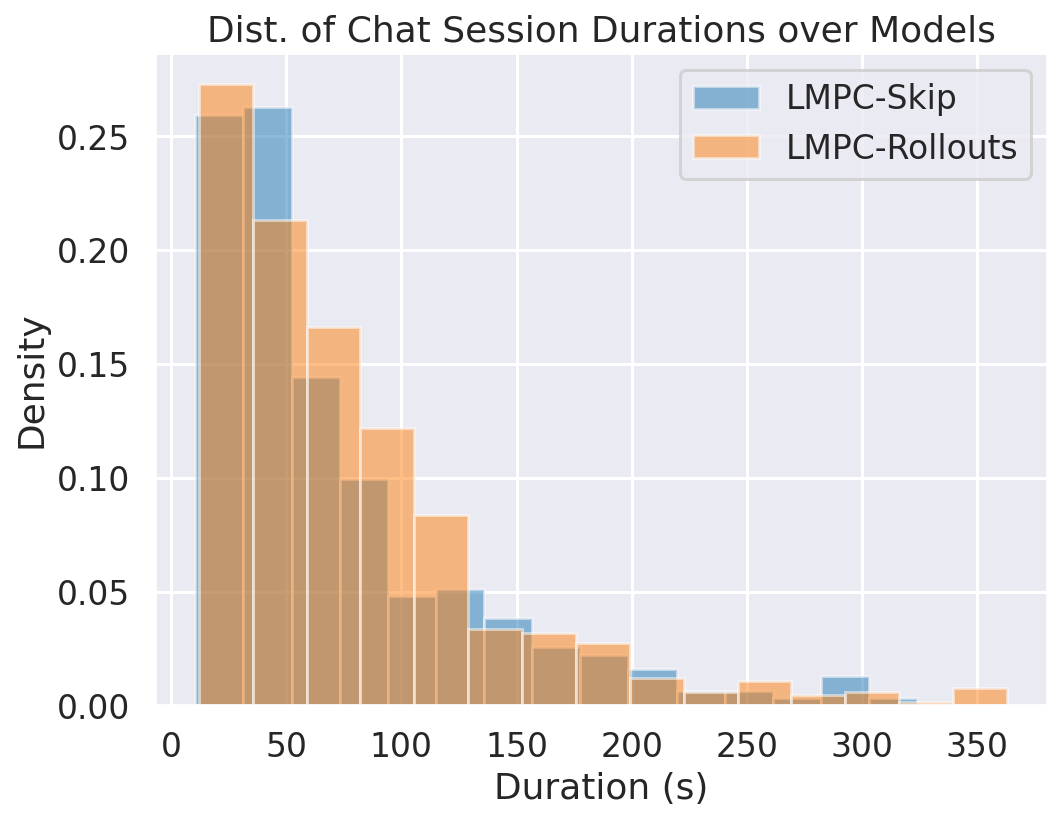}
    \includegraphics[width=0.49\linewidth]{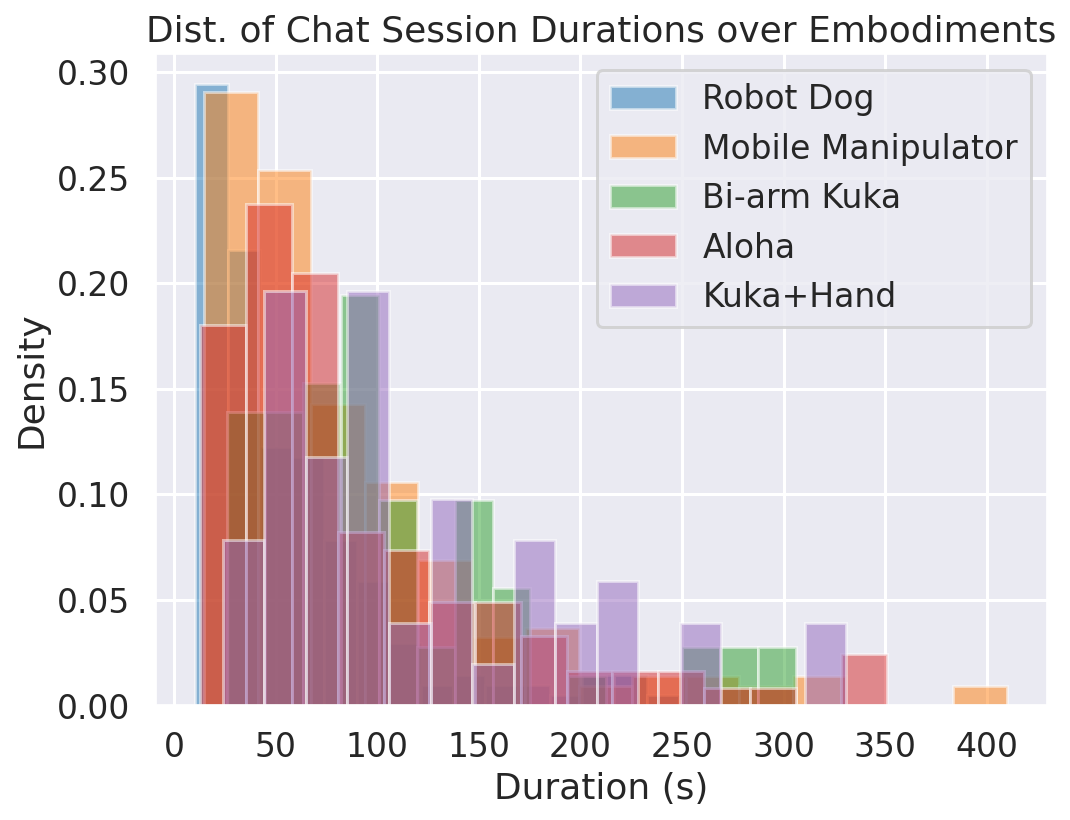}
    \caption{Distribution of Chat Turn Duration over Models and Embodiments}
    \label{fig:dist_duration}
\end{figure}

\begin{table}[th]
    \centering
    \begin{tabular}{lcc}
    \toprule
    Embodiment & Chat Session Duration (s) & Chat Turn Duration (s) \\
    \midrule
    Kuka+Hand & 429 & 97 \\
    Bi-arm Kuka & 406 & 88 \\
    Aloha & 200 & 66 \\
    Mobile Manipulator & 238 & 65 \\
    Robot Dog & 138 & 41 \\
    \bottomrule
    \end{tabular}
    \caption{Median Chat Session and Chat Turn Durations across Embodiments}
    \label{tab:durations_embodiment}
\end{table}

\begin{table}[th]
    \centering
    \begin{tabular}{lcc}
    \toprule
    Model & Chat Session Duration (s) & Chat Turn Duration (s) \\
    \midrule
    LMPC-Rollouts & 187 & 60 \\
    LMPC-Skip & 158 & 49 \\
    \bottomrule
    \end{tabular}
    \caption{Median Chat Session and Chat Turn Durations across Models}
    \label{tab:durations_model}
\end{table}

\begin{figure}[th]
    \centering
    \includegraphics[width=0.5\linewidth]{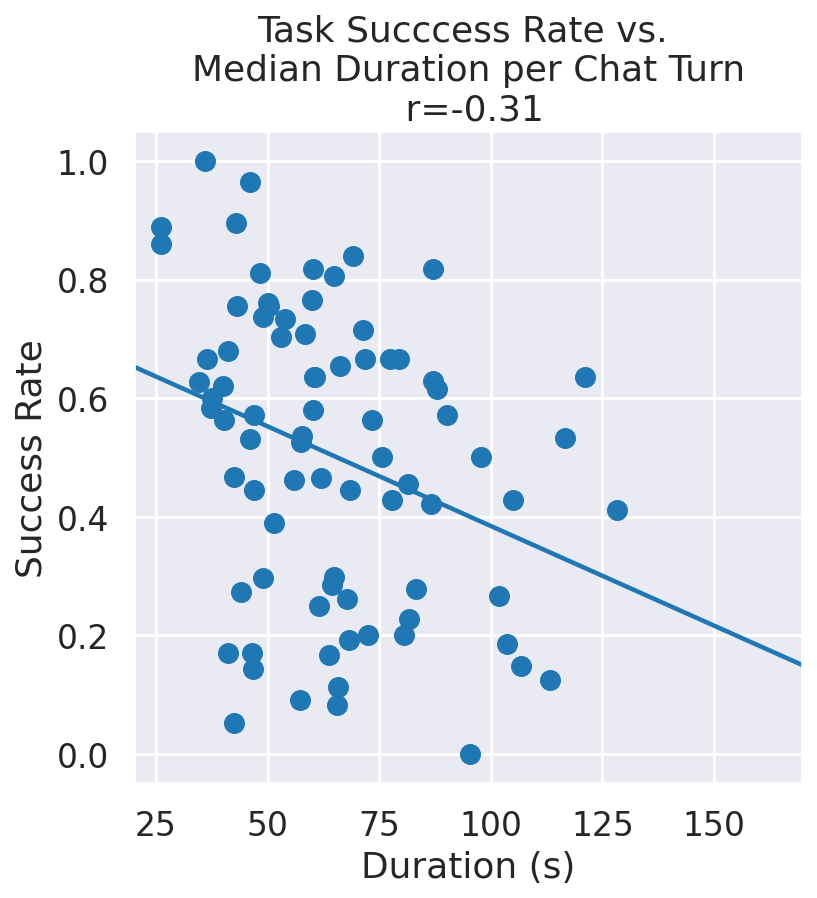}
    \caption{Correlation of Success Rate vs. Median Chat Turn Durations across Tasks.}
    \label{fig:success_vs_duration}
\end{figure}

\textbf{Chat Duration Analysis.}
We measured and analyzed the duration of chat sessions and chat turns and compared them across different embodiments and models.
Chat turn duration measures the total time it took for the model to respond, for the user to run the robot code in simulation, for the user to observe the resulting robot behavior, and for the user to input the subsequent language feedback.
\cref{fig:dist_duration} shows the distribution of chat turn durations across both models and embodiments.
While there are no obvious differences in these distributions, some are more long-tailed than others, and we see this in the median statistics.
In~\cref{tab:durations_embodiment}, we show the median durations for chat sessions and chat turns across different embodiments.
Kuka+Hand and Bi-arm Kuka have significantly higher durations than other embodiments.
This reflects that these embodiments were likely more difficult to teach (it took longer for users to respond) as well as taking longer to simulate (they had tasks that had longer horizons than the other embodiments).
In~\cref{tab:durations_model}, we compare the median durations for LMPC-Rollouts and LMPC-Skip.
LMPC-Rollouts has slightly higher chat turn and chat session durations, and this difference reflects how inference (decoding the LLM for entire chat sessions) for LMPC-Rollouts takes slightly longer than inference for LMPC-Skip.
Lastly, in~\cref{fig:success_vs_duration}, we show a small negative correlation between task success rate and chat duration --- the longer it takes for users to complete a chat turn, the less likely it is for that task to be successful.

\begin{table}[!th]
    \centering
    \begin{tabular}{lcc}
    \toprule
    & Train Tasks & Test Tasks \\
    \midrule
    LMPC-Rollouts-with-Failures &  $-11.5\%$  & $-14.0\%$  \\
    \bottomrule
    \end{tabular}
    \caption{Success Rates of Training LMPC-Rollouts on both Success and Failure chat sessions.}
    \label{fig:task_success_nci}
\end{table}

\textbf{Training LMPC on Both Success and Failures.}
In principle, LMPC-Rollouts (when viewed as a dynamics model) can be trained on both success and failure data (since all chat turns are valid transitions, regardless of whether the session ended in task success). In this version, LMPC-rollouts also predicts (on trajectory termination) whether the predicted rollout would lead to a success or failure. Inference-time search would then be adjusted accordingly to disregard sampled rollouts that ended in predicted failure. While this remains an interesting aspect of LMPC-Rollouts, our main experiments report results from training LMPC-Rollouts on success data only (as a fair comparison with LMPC-Skip, which can only be trained on success data), with which we do observe performance improvements over mixing failure sessions into the training data (results in \cref{fig:task_success_nci}). We hypothesize that training LMPC-Rollouts only on sessions that ended in task success yields more efficient inference-time search, since the alternative of training on both success and failure sessions leads to more unused predicted rollouts that terminate with failure.

\textbf{Task-Level Analysis}
We performed task-level analyses on which tasks saw improvements, degradations, and shared task patterns across the three main compared variants (RAG, LMPC-Skip, and LMPC-Rollouts). However, the median count for each (model variant, task) tuple is only 4, and the median count for each (model variant, task, user) tuple is only 1. These low sample sizes mean that comparing performances on a task level is likely noisy, and some of the following observations may exhibit stronger signals if we had the resources to collect additional human teaching data.

Through model training, some tasks saw improved performance, while others saw degradations. On Test tasks, LMPC-Rollouts had the highest rate of task improved (69\%), while RAG had the highest rate of task degredations (31\%).
See \cref{tab:task-better-same-worse}

\begin{table}[h]
    \centering
    \begin{tabular}{lcccc} 
    \toprule
        Model Name & {Better} & {Same} & {Worse} & Split \\ 
    \midrule
        RAG & 58\% & 10\% & 32\% & train \\
        RAG & 44\% & 25\% & 31\% & test \\
        LMPC-Skip & 68\% & 7\% & 24\% & train \\
        LMPC-Skip & 62\% & 15\% & 23\% & test \\
        LMPC-Rollouts & 68\% & 16\% & 16\% & train \\
        LMPC-Rollouts & 69\% & 8\% & 23\% & test \\
    \bottomrule
    \end{tabular}
    \caption{Percent of tasks that saw improvements and degradations over the base model}
    \label{tab:task-better-same-worse}
\end{table}

There is little overlap among the improved, same, and worsened tasks across the different model variants. Taking the intersection over union (IOU) of the set of tasks that have better, same, and worse success rates across the 3 model variants - the highest IOU is 0.28 for training tasks that did better, while tasks the had the same performance, as well as test tasks that became worse, have no overlap among the different model variants. With manual inspection, given the sample sizes we have, we could not identify task-level patterns that reliably predict if the given task would improve or degrade. See Table \cref{tab:task-iou}

\begin{table}[h]
    \centering
    \begin{tabular}{lcc}
    \toprule
    Mode & Split & IOU \\
    \midrule
    better & train & 0.28 \\
    better & test & 0.14 \\
    same & train & 0.00 \\
    same & test & 0.00 \\
    worse & train & 0.11 \\
    worse & test & 0.00 \\
    \bottomrule
    \end{tabular}
    \caption{Overlap (intersection over union) among tasks that were better/same/worse over the base model across train/test splits.}
    \label{tab:task-iou}
\end{table}

In addition, there are no tasks that consistently fail. There are 0 tasks who had no successes across the base and the trained models.
However, we emphasize that these task-level analyses were performed with very low sample sizes per task, so we are hesitant to use them to draw any concrete conclusions. As an example, one of the Bi-Arm Kuka tasks that saw a degradation in finetuning performance was “bring the red cube 20cm to the left of the green goal.” For this task, the base model achieved a success rate of 33\% (1 success out of 3 trials), while the fine-tuned models and RAG all achieved 0\% (0 out of 2 trials each). This is insufficient data to really conclude that performance on this task has degraded (especially given that these 9 trials were all performed by different users). While we have sufficient data to compare model performances in aggregate, we do not have sufficient data to accurately compare task-level performances.

\subsection{Top-Users and Details on Autonomous Top-Users Selection}
\label{app:top_users}

\begin{table}[!th]
    \centering
    \begin{tabular}{lcc}
    \toprule
    Model & Top Users & Other Users \\
    \midrule
    LMPC-Skip & $+15.1\%$ & $+14.2\%$ \\
    LMPC-Rollouts & $+26.3\%$ & $+18.9\%$ \\
    \bottomrule
    \end{tabular}
    \caption{Success rate improvements by user group for test tasks.}
    \label{tab:success_top_other_user}
\end{table}

We identify top-users by evaluating how well they perform on training tasks (Appendix \cref{app:tasks}), weighted by task difficulty.
Let there be $N$ tasks and $K$ users.
Let $s(n,k)$ denote the self-reported success rate of the $n$th task for the $k$th user, $c(n,k)$ denote the number of times the $k$th user taught the $n$th task, and $\bar{c}(n,k) = \mathbbm{1}(c(n,k) \ge 1)$ to indicate whether or not the $k$th user has taught the $n$th task.
Due to practical constraints, $\bar{c}(n,k) = 0$ for many user-task pairs.
We define the task difficulty rating $d(n)$ as the average task failure rate across all users:
$d(n) = 1 - \frac{1}{K_n} \sum_{k=1}^{K} s(n,k) \bar{c}(n,k)$, where $K_n = \sum_{k=1}^K \bar{c}(n, k)$.
Then, we define a user performance score as a user's average success rate weighted by the task difficulty rating: $h(k) = \sum_{n=1}^{N_k} d(n) s(n,k) \bar{c}(n,k)$, where $N_k = \sum_{n=1}^N \bar{c}(n, k)$.
We define top-users as those who are in the top $75$th percentile by this performance score. We refer to the remaining users as ``other users".
Data from top-users only account for $10.7\%$ of the training data. 
Since we do not over-index on top-user data during training (we only modify the prompt to use the user id “top user” instead of the actual user id), the small proportion of top user data limits the degree to which the model is biased toward the top users. 
As evidence that the model is not just learning from top-user data, in \cref{tab:expert_conditioning}, model performance drops significantly if we train only on top user data. 
The combination of high-quality data from top-users, along with a much bigger (~9x) set of more diverse data from other users, proves to be essential for finetuned model performance.

\cref{tab:success_top_other_user} shows the average performance improvements of user-conditioned LMPC over the base model split by top users and other users. We observe largest performance improvements when LMPC-Rollouts (conditioned on top-users) is served to top users directly, and this is less evident with LMPC-Skip, suggesting that inference-time search (via MPC) over future interactions performs better at catering to improving the teachability of top users (\eg satisfying their criterion for success).


Our experiments in the main paper (\cref{tab:expert_conditioning}) demonstrate that conditioning LMPC on top-users can drive performance improvements for all users -- but what makes top-user teaching data different from other users? To explore this question, we define 4 axes to categorize the feedback: (1) Quantitative, (2) Related to Code, (3) Detailed, and (4) Kind. Classification is done via GPT-4 with a few-shot prompt. 
See \cref{app:trait_prompts} for the prompts used to do the trait classification.
Each message of feedback is classified with these traits. If a given chat session has a trait for any message in the session, we say the entire session had that trait. For example, if one question is ``detailed", we say the session had ``detailed" feedback.

\begin{figure}[ht]
    \centering
    \includegraphics[width=0.55\linewidth]{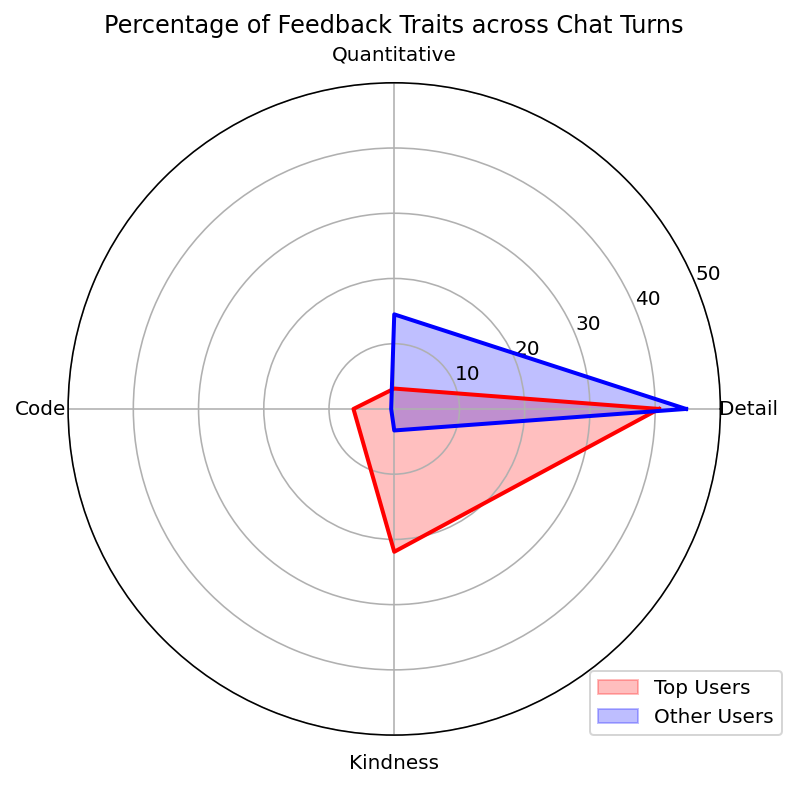}
    \caption{Analysis of feedback traits across all chat turns for Top Users and Other Users}
    \label{fig:top_user_feedback_analysis}
\end{figure}

As can be seen in the analysis of feedback for all users in \cref{fig:top_user_feedback_analysis}, both the top-users and other users provide quite detailed feedback. One interesting trait that stood out is that top users are more ``kind", which might imply that they have more patience for errors in the code generated. This may lead to more thoughtful responses that end up in successful policies. Another surprising finding was that the other-users were more quantitative in their responses than the top users, indicating that it's preferred to give softer feedback signals than precise numbers. These insights are preliminary and further investigation is needed to understand what makes top users successful.


To better understand how top-users teach, we asked them about their teaching strategies and what advice they would give to others. 
Many top-users started out with simple, natural language instructions. 
Then they would review the robot's performance. 
If that performance was not satisfactory, then they would provide more detailed feedback.
One top-user summarized this approach quite well: ``Think of talking to a toddler, sentences are a couple words long and are easy to understand; however, this toddler knows words or terms from a university physics textbook (e.g. rotational velocity, perpendicular, yaw, pitch, roll).''


\subsubsection{Trait Classification Prompts}
\label{app:trait_prompts}

\textbf{Code Feedback Prompt}
\begin{tcolorbox}[
    colframe=darkgray, 
    boxrule=0.1pt, 
    colback=lightgray!10, %
    arc=3pt, 
    fontupper=\tiny,
    breakable,
    halign=left,
    ]
\# INSTRUCTIONS:
Given an instruction teacher gives to a student, rate this instruction based on "Code Feedback".
After considering the instruction carefully, output one of three following ratings along with a justification:

* NEGATIVE: There is *no* feedback related to the code in the instruction.

* NEUTRAL: There is a *fuzzy* feedback related to the code in the instruction.

* POSITIVE: There is *clear* feedback related to the code in the instruction.

\# EXAMPLE 1:

QUERY: grasp apple and place it on top of the cube

JUSTIFICATION: There is *no* feedback related to the code in the instruction.

RATING: NEGATIVE

\# EXAMPLE 2:

QUERY: set your turning\_speed equal 0

JUSTIFICATION: There is *clear* feedback related to the code in the instruction.

RATING: POSITIVE

\# IMPORTANT: Always output justification first, then the rating.

\# INPUT

QUERY: \{query\}

JUSTIFICATION:
\end{tcolorbox}

\textbf{Quantitative Feedback Prompt}
\begin{tcolorbox}[
    colframe=darkgray, 
    boxrule=0.2pt, 
    colback=lightgray!10, %
    arc=3pt, 
    fontupper=\tiny,
    halign=left
    ]
\# INSTRUCTIONS:
Given an instruction teacher gives to a student, rate this instruction based on "Quantitative Feedback".
After considering the instruction carefully, output one of three following ratings along with a justification:

* NEGATIVE: There are *no* numerical and quantitative information in the instruction.

* NEUTRAL: There is a *fuzzy* indication of numerical and quantitative information in the instruction.

* POSITIVE: There is *clear* numerical and quantitative information in the instruction.

\# EXAMPLE 1:

QUERY: move the left arm towards green cube and push it to the right 20cm

JUSTIFICATION: There is *clear* numerical and quantitative information in the instruction.

RATING: POSITIVE

\# EXAMPLE 2:

QUERY: pick up the connector

JUSTIFICATION: There are *no* numerical and quantitative information in the instruction.

RATING: NEGATIVE

\# IMPORTANT: Always output justification first, then the rating.

\# INPUT

QUERY: \{query\}

JUSTIFICATION:
\end{tcolorbox}

\textbf{Kindness Prompt}

\begin{tcolorbox}[
    colframe=darkgray, 
    boxrule=0.2pt, 
    colback=lightgray!10, %
    arc=3pt, 
    fontupper=\tiny,
    breakable,
    halign=left
    ]
\# INSTRUCTIONS: 
Given an instruction teacher gives to a student, rate this instruction based on "Kindness".
After considering the instruction carefully, output one of three following ratings along with a justification:

* NEGATIVE: This instruction is *not* kind.

* NEUTRAL: This instruction is neither kind nor unkind.

* POSITIVE: This instruction is kind.

\# EXAMPLE 1: QUERY: move the cube a little bit to the left please

JUSTIFICATION: This instruction is kind.

RATING: POSITIVE

\# EXAMPLE 2:

QUERY: did you forget how to walk? Please reposition yourself heading south

JUSTIFICATION: This instruction is *not* kind.

RATING: NEGATIVE

\# EXAMPLE 3:

QUERY: close the door by pushing it.

JUSTIFICATION: This instruction is neither kind nor unkind.

RATING: NEUTRAL

\# IMPORTANT: Always output justification first, then the rating.

\# INPUT

QUERY: \{query\}

JUSTIFICATION:
\end{tcolorbox}

\textbf{Detail Prompt}

\begin{tcolorbox}[
    colframe=darkgray, 
    boxrule=0.2pt, 
    colback=lightgray!10, %
    arc=3pt, 
    fontupper=\tiny,
    breakable,
    halign=left
    ]

\# INSTRUCTIONS: 
Given an instruction teacher gives to a student, rate this instruction based on "Detail".
After considering the instruction carefully, output one of three following ratings along with a justification:

* NEGATIVE: This instruction is *not* detailed.

* NEUTRAL: This instruction is neither detailed nor undetailed.

* POSITIVE: This instruction is quite detailed.

\# EXAMPLE 1:

QUERY: very good, now extend your front left paw as far forward as possible without losing balance on the rest of your legs. Change the angle of your torso as needed to maintain balance

JUSTIFICATION: This instruction is quite detailed.

RATING: POSITIVE

\# EXAMPLE 2:

QUERY: stand up

JUSTIFICATION: This instruction is *not* detailed.

RATING: NEGATIVE

\# IMPORTANT: Always output justification first, then the rating.

\# INPUT

QUERY: \{query\}

JUSTIFICATION:
\end{tcolorbox}
\subsection{RAG Implementation Details}
\label{app:rag}

To implement our RAG baseline, we first construct an embedding dataset from the same data used to train LMPC-Skip.
This dataset includes each data point's initial user instruction, its Gecko embedding (obtained via an embedding model based on PaLM 2), and the final successful response code.
During inference, we use the embedding of the initial user instruction of the current chat session to find the $5$ most relevant data points of the same robot embodiment from the dataset.
This is done by first selecting the closest $30\%$ of data by cosine similarity (embeddings are normalized), then applying the farthest point sampling algorithm among this set to ensure diversity of the selected data points.
Finally, the retrieved data points are re-ordered from lowest to highest relevancy, such that the most relevant example is closest to the current instruction.
This ordered list of (instruction, code) pairs is then inserted into the context of the LLM prompt.
\subsection{Data Augmentation Details}
\label{app:data_augmentation}

\begin{table}[!th]
    \centering
    \begin{tabular}{lcc}
    \toprule
    & \multicolumn{2}{c}{Success Rate Diff w/o Data Augmentation} \\
    \cmidrule(lr){2-3}
    Model & Train Tasks & Test Tasks \\
    \midrule
    LMPC-Skip w/o Aug & -7.1\% & +0.6\% \\
    LMPC-Rollouts w/o Aug & +2.8\% & -7.0\% \\
    \bottomrule
    \end{tabular}
    \caption{Success rate differences between models that do not use data augmentation and models that do.}
    \label{tab:data_aug}
\end{table}

We augment user inputs, including 1st input (task request) and subsequent feedback and corrections with PaLM 2-L. The prompt for augmentation asks PaLM 2-L to rewrite original text in $K$ different ways by replacing words with synonyms, rephrasing, changing grammatical structure, sentence lengths, punctuation, etc. We also specifically prompt the model to output the $K$ ways in one batch with variations in the batch and use relatively high generation temperature (0.8) to ensure the output is sufficiently diverse.

For example, a user's request of ``pick up the cube" is rewritten into ``grab the cube and raise it ", ``lift up the cube", ``raise the cube", etc; user's correction ``wrong direction, keep hopping but turn the opposite direction" for the ``hop while turning counterclockwise" task is rewritten into ``that is the incorrect direction, maintain hopping but go the opposite way", ``you are going the wrong way, keep hopping but turn in the opposite direction", ``wrong direction, maintain hopping but turn the opposite way", etc.

The augmented data is used for training both LMPC-Skip and LMPC-Rollout models.
We report the success rate differences when training models without data augmentation in \cref{tab:data_aug}. 
Without data augmentation, LMPC-Skip performs worse on train tasks, but on par on test tasks.
By contrast, without data augmentation, LMPC-Rollouts performs on par on train tasks, but much worse on test tasks.
This suggests that the generalization capabilities of LMPC-Rollouts benefits more from data augmentation than does LMPC-Skip.
We hypothesize this is due to that data augmentation makes LMPC-Rollouts' chat session predictions more robust to compounding errors, leading to better predictions of feedback dialogue.

\subsection{Analysis of Chat Feedback Embeddings}
\label{app:feedback_embeddings}

\begin{figure}[h]
    \centering
    \includegraphics[width=\linewidth]{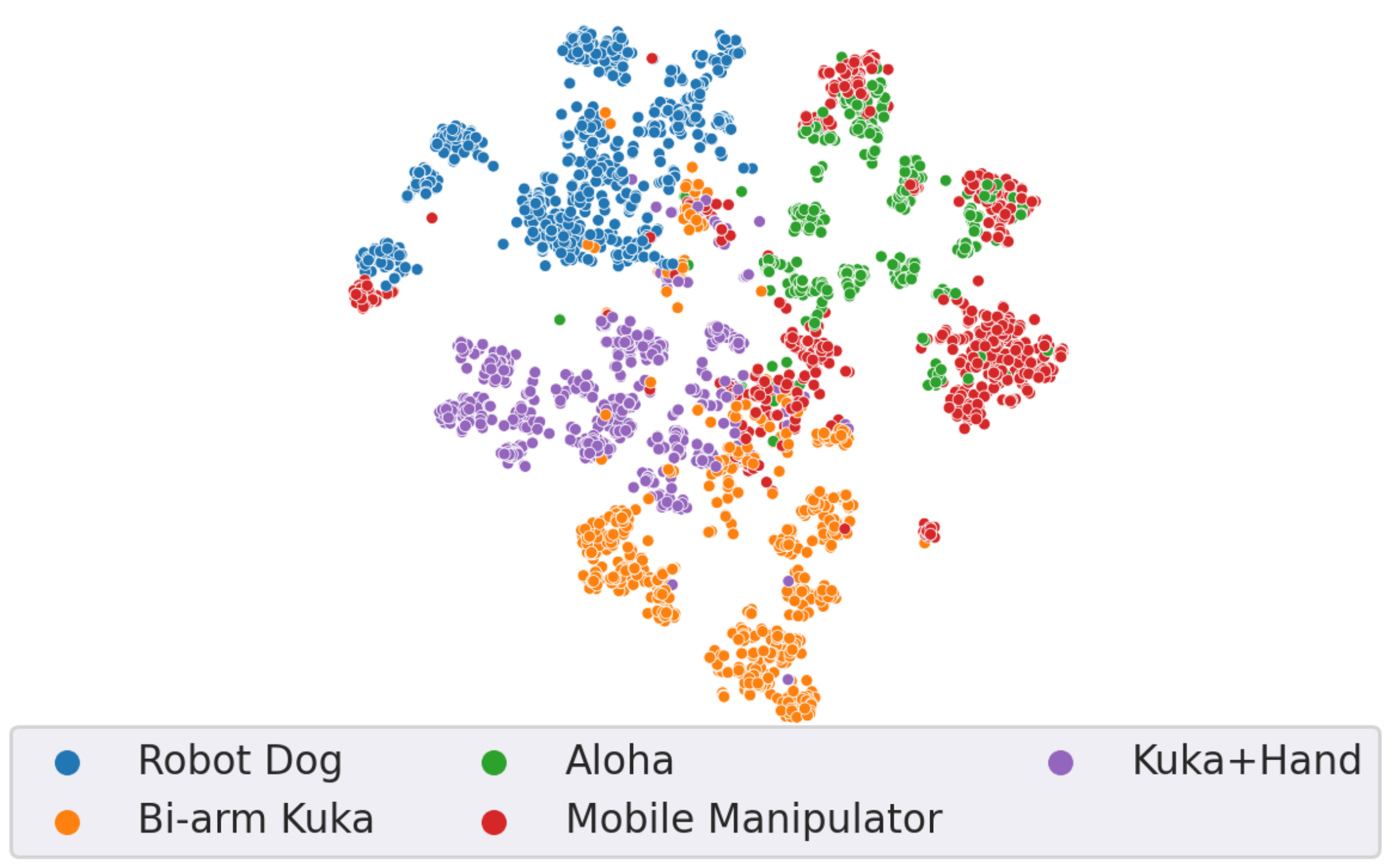}
    \caption{T-SNE plot of embeddings of human feedback across embodiment.}
    \label{fig:tsne_embodiments}
\end{figure}

\begin{figure}[h]
    \centering
    \includegraphics[width=\linewidth]{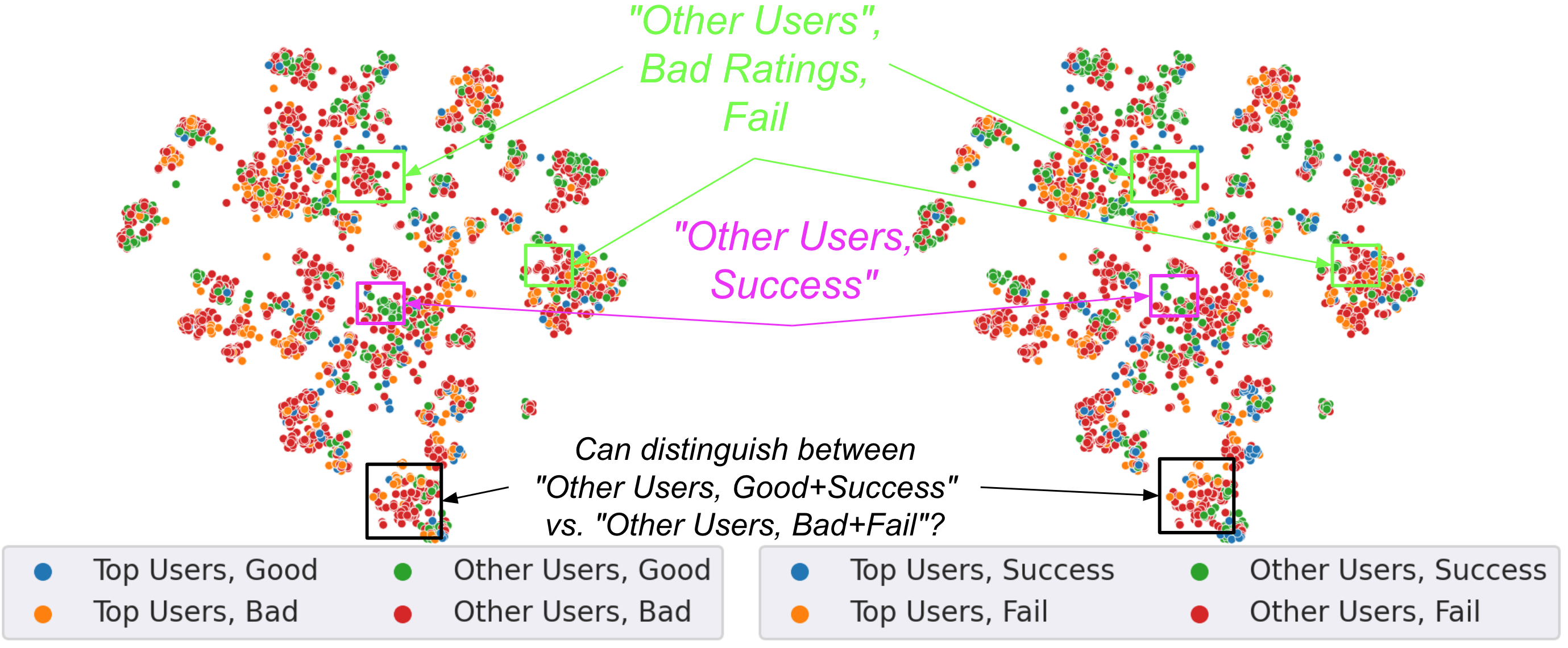}
    \caption{T-SNE plot of embeddings of human feedback across experts and non-experts, and across good/bad chat ratings (left) and whether or not that feedback belongs to a chat session that was eventually a success/failure (right).}
    \label{fig:tsne_expert_rating_success}
    \vspace{-1em}
\end{figure}

What kinds of feedback do users provide to steer robot behaviors? To study this, we compute language embeddings on all individual chat turn user queries, using a finetuned T5 XL model~\citep{raffel2020exploring}.
Then, we compute a T-SNE embedding vectors, mapping each embedding with associated features for: whether the query was from a ``Top User'' or not, whether the user rated the LLM response to the query as ``Good'' or ``Bad'', whether the query was from a session which resulted in a ``Success'' or ``Fail'', and which robot embodiment the session used. 
First, we find that user queries are indeed highly correlated with specific embodiments, as shown in Figure~\ref{fig:tsne_embodiments}.
This intuitively makes sense since language embeddings will consider semantic details like specific syntax or verbal suggestions that are specific to tasks or robot physics that are only present on a specific embodiment (for example, ``raise your paw higher'' is only relevant for the Robot Dog embodiment). 
Second, we find that there are clear cases where for even the same embodiment, ``Top User'' semantic language embeddings are clearly clustered separately from ``Other Users'', as shown in Figure~\ref{fig:tsne_expert_rating_success}.
Additionally, we also find other interesting clusters, such as where ``Other Users'' seem to be more pessimistic about LLM responses by giving clusters of ``Bad Ratings'', which result in either ``Success'' or ``Fail''.
\begin{figure*}[h]
    \centering
    \includegraphics[width=0.85\linewidth]{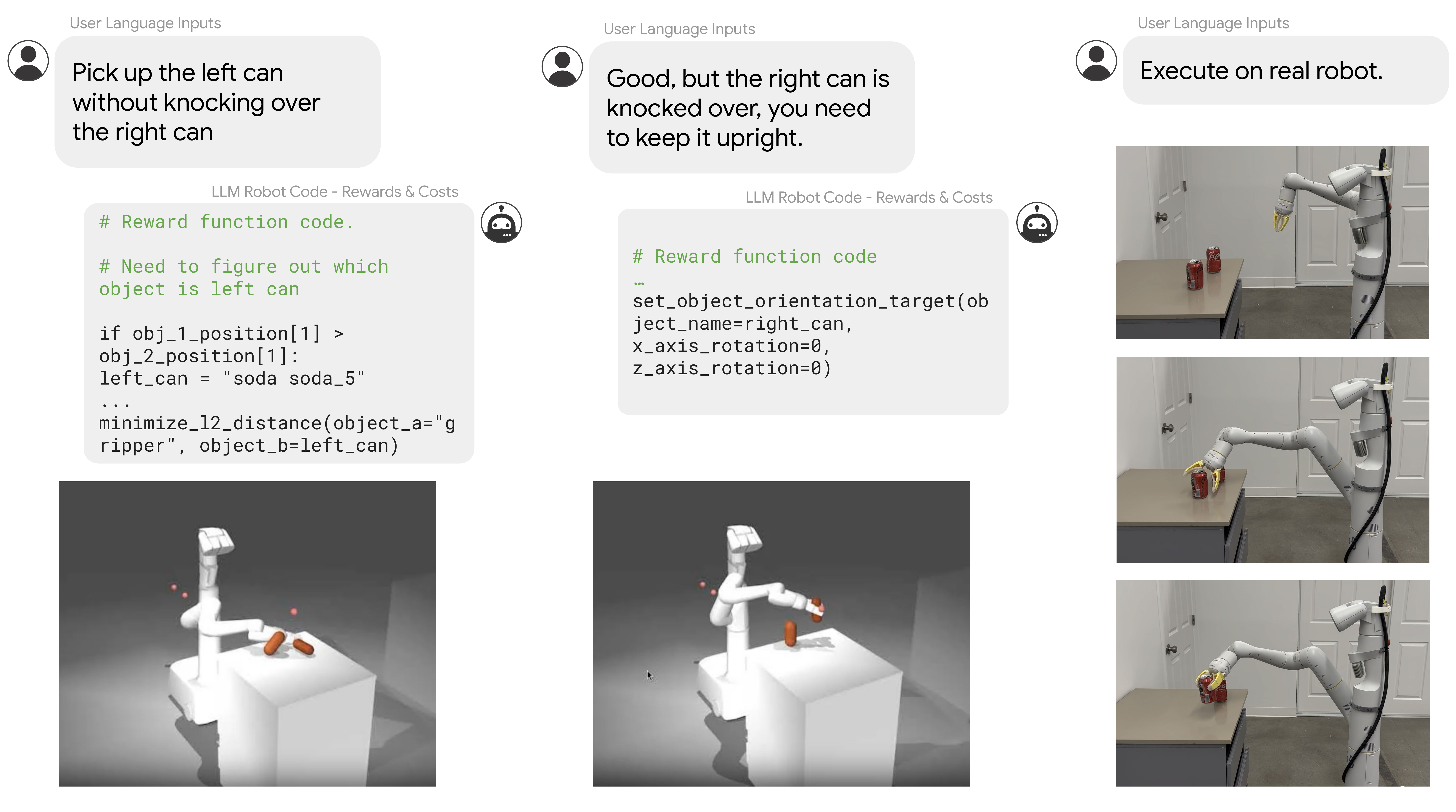}
    \caption{Real world teaching example on mobile manipulator using the MJPC-as-Planner approach.}
    \label{fig:real_world_teaching_meta}
\end{figure*}

\begin{table*}[th]
\centering
\begin{tabular}{llcccccc}
\toprule
\textbf{Embodiment} & \textbf{Task} & \textbf{Ours-Sim} & \textbf{Ours-Real} & \textbf{N Chat Turns} & \textbf{PaLM 2-S Sim} & \textbf{PaLM 2-S Real} & \textbf{N Chat Turns}  \\ 
 \midrule 
 Robot Dog & High-Five with left hand  & $100\%$ & $100\%$ & $3.0$ & $100\%$ & $75\%$ & $2.3$ \\ 
 & Downward Dog  & $100\%$ & $100\%$ & $2.8$ & $100\%$ & $100\%$ & $1.3$ \\ 
 & Walk forward in a trotting gait  &  $100\%$ & $100\%$ & $2.8$ & $25\%$ & $25\%$ & $2.0$ \\
 & Hop  & $100\%$ & $75\%$ & $2.3$ & $50\%$ & $25\%$ & $2.0$ \\ 
 & Hop while turning counterclockwise & $100\%$ & $25\%$ & $4.0$ & $100\%$ & $25\%$  &$5.0$ \\
\midrule
 Mobile Manipulator & Open top drawer half-way & $100\%$ &  $100\%$ & $3.2$ & $100\%$ & $100\%$ & $3.4$  \\ 
 & Push coke can from right to left & $100\%$ & $80\%$ & $2.0$ & $100\%$ & $60\%$ & $2.0$ \\ 
 & Knock over coke can & $100\%$ & $20\%$ & $3.0$ & $80\%$ & $20\%$ & $5.0$  \\ 
 \bottomrule
\end{tabular}
\caption{Sim vs. Real Results}
\label{table:sim_vs_real}
\end{table*}

\subsection{Real Robot Experiments}
\label{app:real_robots}

\textbf{Distillation for Robot Dog.}
Our robot dog distilled policy is based on the Locomotion-Transformer model, which uses a Transformer to map sequences of velocity commands, proprioceptive observations, and past actions to next actions \cite{caluwaerts2023barkour}. We generalize the original velocity command formulation to MJPC cost weights and parameters as the objective tokens. Our final policy consists of a transformer with $d=256$ and four layers, totaling roughly 3.2 million parameters.

To train the policy, we used online imitation learning (DAgger) against an expert MJPC policy over a distribution of tasks encompassing both static posing and locomotion behaviors. This task distribution was constructed by uniformly randomizing key target parameter values, including robot velocity, torso height and pitch, foot positions, and foot stepping. Due to the diversity of the task distribution and domain randomization, we found offline imitation (BC) to be unsuccessful. We also smooth all actions by applying an exponential filter with strength 0.9.

\textbf{MJPC-as-Planner for real Mobile Manipulator.}
For the main experiment results in deploying the taught skills in simulation to the real mobile manipulator robot, we extend the MJPC-as-Planner approach from prior work by Yu et al \cite{yu2023language}. In particular, to obtain a simulated replica of the real scene, Yu et al. used an open-vocabulary object detector to detect and segment objects in the scene and fit known mesh models to the corresponding point clouds. The reconstructed simulation scene is used in MJPC to generate a trajectory plan, which is then executed on the robot.

Though the prior work showed good results in real-world, it required knowing the list of objects in the scene and their corresponding meshes in order to query the object detection model and recreate the scene. In this work, we improve the perception pipeline on both fronts: 1) we use a large visual language model (VLM) to identify all the objects seen by the robot in the environment, each of which is then segmented using the Segment Anything (SAM) model \cite{kirillov2023segment} to achieve precise object localization, 2) we opt to use generic primitive shapes consisting of capsules and boxes to represent the objects, which enables us to represent a wide range of objects without having to obtain detailed meshes. As a result, we can apply our approach to more diverse environments with unknown objects and be able to teach the robot manipulation skills on them. An example can be seen in
\cref{fig:real_world_teaching_meta}.


\textbf{Sim-to-Real Gap.}
\cref{table:sim_vs_real} shows a comparison between sim and real performances on the set of tasks we evaluated in the real world.
For open drawer task, we achieve $100\%$ success rate in real world, likely because this task is quasi-static thus there is very little physical domain gap. 
By contrast, knock over coke can only achieved $20\%$ success rate in the real world, due to the velocity of the end effector not being fast enough. 
This is caused by physical modelling domain gap, which allows the robot to knock over the coke can with a lower end-effector velocity.
For the hop while turning task we observe a large discrepancy between simulation and real world. 
While we are able to teach the robot to hop, it often trips and falls after a few hops.
This is due to that a highly agile hopping while turning behavior is outside the training distribution of the distilled policy.
Adapting the distillation training distribution to the teaching data is a promising direction for future research.

\begin{figure*}[!t]
    \centering
    \includegraphics[width=0.9\linewidth]{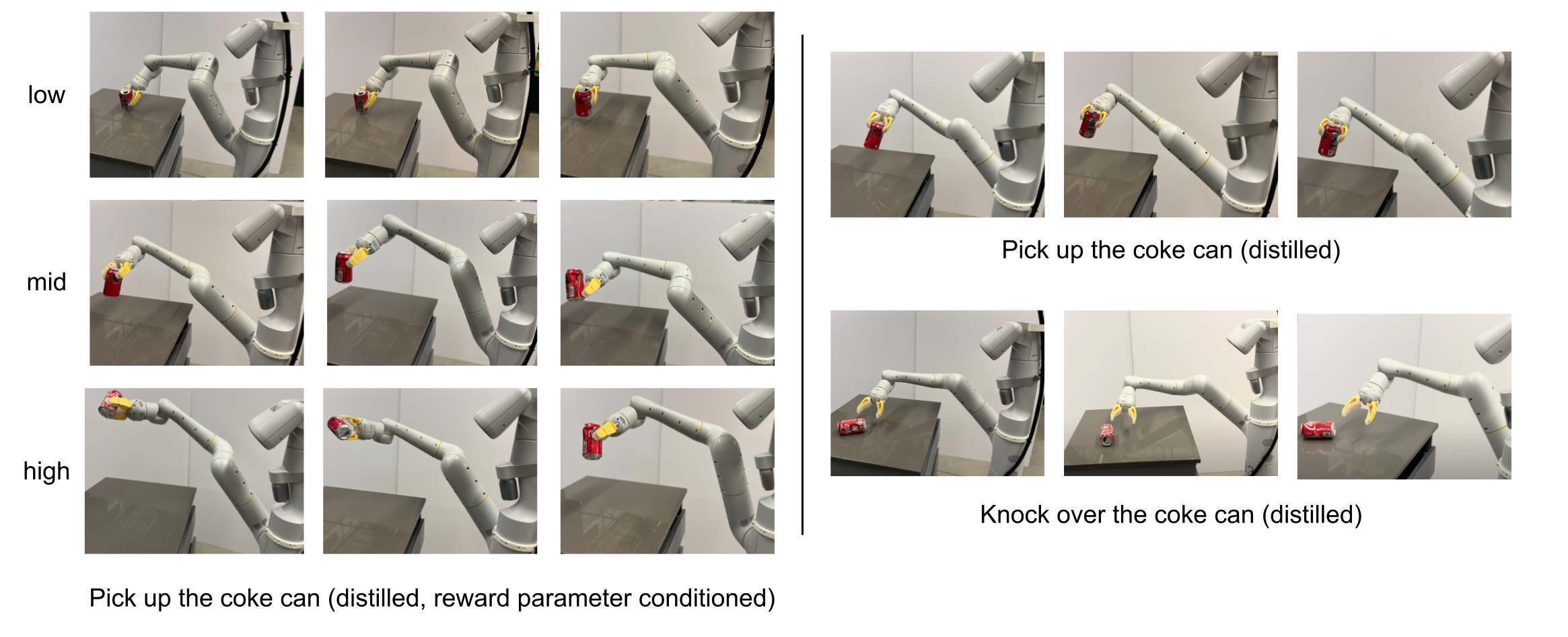}
    \caption{\textbf{Example Rollouts of reward conditioned distilled policy on mobile manipulator.} Apart from using MPJC-as-Planner for real world deployment, we also explored distilling the behavior into a policy using imitation learning following the robot dog example. This no longer requires accurate state estimation.}
    \label{fig:reward_conditioned_behavior}
\end{figure*}

\textbf{Distillation for Mobile Manipulator.}
There are a few limitations with the MJPC-as-Planner approach: 1) generating the plan with MJPC is not feasible for onboard computing due to computation requirements, 2) it needs multiple models to identify and segment the objects, adding additional complexities to the system, 3) it does not respond to changes in the environment during execution. To make a step towards addressing these issues, we explore the reward-conditioned policy distillation approach used for the Robot Dog on the Mobile Manipulator. Specifically, we validate the idea in two settings: 1) use the reward to condition final object height for the picking task, 2) use reward to condition picking up or knocking over a can. We use MJPC to generate 30k and 10k trajectories respectively with maximally 150 steps. The robot observation in each step consists of the simulated depth image from robot camera and the reward parameters. We train the policies based on the RT-1 model \cite{brohan2023rt1} using the generated dataset and deployed the policies on a real mobile manipulator robot. Distilling a reward conditioned policy allows us to deploy the policy with onboard computing and achieve more robust behavior with closed-loop control.
Although we have yet to perform quantitative evaluations of the distilled Mobile Manipulator policy, we demonstrate example policy rollouts in \cref{fig:reward_conditioned_behavior}.

\subsection{Language Model Training Details}
\label{app:finetune}

For finetuning models, we set the number of training steps to cover $10$ epochs of the available training data, apply Adam with a learning rate of $5\times10^{-3}$, a linear ramp up and cosine decay learning rate scheduler, a batch size of 4, and a context length of 4096 tokens.

Because the LMPC-Rollouts model needs to predict the entire remaining chat session, it is much slower than LMPC-Skip at inference time.
According to user feedback, the slowed inference time degrades the teaching experience, making the chat session less engaging, potentially reducing data quality.
To address this issue, we performed 8-bit quantization on the LMPC-Rollout models after finetuning, and we serve the quantized LMPC-Rollout models.
We did not observe noticeably performance drops with the quantized model.
See \cref{tab:inference_times} for measured inference times for these models --- LMPC-Rollouts without quantization is much slower than LMPC-Skip, while with quantization the inference times are similar.

\begin{table}[!th]
    \centering
    \begin{tabular}{ccc}
    \toprule
    LMPC-Skip & LMPC-Rollouts & LMPC-Rollouts-No-Quantization  \\
    \midrule
     $1.1\pm0.2$    &  $1.0\pm0.4$ & $7.4\pm4.7$\\
    \bottomrule
    \end{tabular}
    \caption{Model inference times in seconds.}
    \label{tab:inference_times}
\end{table}
\subsection{User Performance Drift Analysis}
\label{app:user_drift}
Evaluating models over an extended period of time introduces the concern that as users obtain more practice teaching the robot, they become more proficient, and model performance improvements may actually be caused by users' improved teaching skills, instead of improved model capability.
We took three measures to mitigate this concern.
First, we conducted pilot data collection sessions for each robot embodiment, so users could acquire a base level of familiarity with each embodiment before conducting official data collection.
Second, we evaluated LMPC variants during the same data collection days, so their differences are unlikely to be caused by user performance drift.
Third, we explicitly compared user performance with the base LLM during the first half and the second half of our experiments.
From the first to the second period, mean success rate for each user changed by $-0.6\%$ across all tasks, with a standard deviation of $9.3\%$, and the two periods show a Pearson correlation coefficient of $0.87$.

Another way to gauge potential changes in the users' teaching experience is by measuring the self-reported cognitive load of the teachers at the end of each day of data. 
Because different subsets of users taught robots on different days, we analyzed our data in terms of how each user experienced the cognitive load of teaching the robots on their first day vs. on their last day. 
We used a subset of the NASA-TLX measure of cognitive load~\cite{hart2006TLX} and analyzed the perceived mental demand, effort, performance, and frustration dimensions~\cite{galy2018measuring}. 
There were 13 users who completed our end-of-day questionnaires so we ran pair-wise t-tests (2-sided) on their data (N=13). 
We found no statistically significant differences in teachers' first vs. last days of teaching robots in terms of mental demand (\textit{p}=0.26), effort (\textit{p}=0.47), performance (\textit{p}=0.22) or frustration (\textit{p}=0.54). These are all well above the cut-off p-value for statistical significance of .05; with Bonferroni corrections, the cut-off value would be even lower at .0125.

These results suggest minimal user performance or user experience change over time, so differences among models are more likely the result of changes in model capability, not user teaching proficiency.
\subsection{Failure mode analysis}
\label{app:failure_modes}

\begin{table*}[!ht]
    \centering
    \begin{tabular}{lccccc}
    \toprule
    & \multicolumn{5}{c}{Failure Modes} \\
    \cmidrule{2-6}
    Model & Invalid Code & Repeated Code & Non-responsive Code & Incomplete Code & All Failures \\
    \midrule
    PaLM 2-S & 17.4\% & 10.9\% & 16.8\% & 7.6\% & 35.3\% \\
    RAG & 6.4\% & \textbf{6.7\%} & 19.8\% & 6.4\% & 38.5\% \\
    LMPC-Skip & 9.5\% & 7.6\% & 11.9\% & \textbf{3.8\%} & \textbf{23.0\%} \\
    LMPC-Rollout & \textbf{7.8\%} & 7.0\% & \textbf{11.3\%} & 4.0\% & 24.7\% \\
    \bottomrule
    \end{tabular}
    \caption{Failure Mode as percentage of all chat sessions.}
    \label{tab:failure_modes}
\end{table*}

We categorized the following failure modes across our compared models.
Failure mode 1) is outputting code with errors or executable code. 
In instances where it was outputting executable code, we further checked if 2) the failure was from repeated code outputs, 3) from incomplete plans, or 4) from the LLM not responding to the user's feedback. 
In order to identify these failure modes, we prompted an LLM to classify them from chat session data.

See \cref{tab:failure_modes}, where it shows the percentage of chat sessions that resulted in each failure mode across all chat sessions (the denominator is the total number of chat sessions for that model, not the number of failures for that model).
Please note that a particular chat session may appear in multiple failure modes, so these failure mode sets are not disjoint.
As expected, LMPC models have overall fewer failures than baselines.
The most frequent failure mode is outputting code that is not responsive to user feedback.
The least frequent failure mode is outputting incomplete code.

\subsection{Fine-tuned Models on Code-writing Benchmarks}
\label{app:coding_benchmarks}

\begin{table}[!th]
    \centering
    \begin{tabular}{ccc}
    \toprule
     & \multicolumn{2}{c}{Pass@1} \\
     Model & Iteration 1 & Iteration 2 \\
    \midrule
    PaLM 2-S & \multicolumn{2}{c}{51\%} \\
    LMPC-Rollouts & 51\% & 51\% \\
    LMPC-Skip & 49\% & 49\% \\
    \bottomrule
    \end{tabular}
    \caption{Performance on RoboCodeGen on finetuned models.}
    \label{tab:robocodegen}
\end{table}

One concern with model finetuning is that the finetuned model may forget some of its original capabilities.
In our case, we are specifically concerned about whether or not finetuning our model degrades general code-writing capabilities of the LLM.
To test this, we evaluated our models after one and two iterations of finetuning on the RoboCodeGen benchmark~\cite{liang2023code}. 
As seen in \cref{tab:robocodegen}, there is relatively no degradation between the first iteration and the baseline model or between first and second iteration.
We attribute this to our training data being in the distribution of the base LLM as well as using code, therefore not biasing the model away from code generations.

\subsection{Robot Embodiment Details}
\label{app:MJPC}

\noindent\textbf{Robot Dog.}
The robot dog is a small quadruped robot with an onboard computer and battery power. The robot was developed in-house based on the design from~\cite{caluwaerts2023barkour}. The robot has a standing height of approximately 0.4\,m. Each of the robot's 4 legs has 3 DoFs with a peak joint torque of 18\,Nm.

The distilled policy (Section~\ref{app:real_robots}) runs on the onboard computer (Intel NUC11TNBv7) and provides joint position commands at 50\,Hz to a low-level PD controller that outputs joint torques at 1\,kHz. We set the P-gain to 50\,Nm/rad and D-gain to 1.1\,Nm\,s/rad. We use ROS2 over a Wi-Fi connection to transmit the model output (the reward function code) from a desktop computer to the robot when a user clicks the \emph{Run on Robot} command in the chat UI.

The simulated environment for the robot dog contains a door without latch and a three-level kitchen drawer.
In the MJPC implementation, we place position and orientation sensors for the robot torso and end-effectors, as well as joint sensors for the articulated objects (e.g. door hinge angle). We then design a set of APIs that the LLM can use to modulate the desired absolute and relative sensor values (see more details in prompts).
By coordinating the movements of different legs in MJPC simulation, the robot dog can achieve a rich set of skills from locomotion to posing. Furthermore, although the robot dog does not have any form of gripper, it can interact with the external world using its torso and limbs to perform tasks such as open the door or close the drawer.


\noindent\textbf{Mobile Manipulator.}
The mobile manipulator~\cite{herzog2023deep} consists of a 7-DoF arm and a parallel jaw gripper. The simulated environment contains the simulated robot in front of household objects (apple, coke can, and cube) placed on a counter with drawers.  

For the MJPC implementation, we place gripper and joint sensors on the robot, position and orientation sensors on household objects, and joint sensors on articulated objects (e.g. drawer hinges).

With the MJPC planner, the robot is able to execute a rich set of tasks with various constraints, including opening the drawer half way, picking/pushing object A to a certain location without knocking over object B, lifting objects up to a certain position and orientation, etc. We further test the skills on the real robot.

We use a gRPC connection over Wi-Fi to transmit the reward function code output by the model from a desktop computer to the robot when a user clicks the \emph{Run on Robot} command in the chat UI.


\noindent\textbf{Aloha.}
The Aloha bi-manual robot \cite{zhao2023learning} consists of two 6-DoF arms fixed to a table, each with a 1-DoF parallel gripper. Household objects (e.g. apple, soda can, cube, and bowl) are placed on the table.

The MJPC implementation includes sensors for the position and orientation of each arm, and an API to get the position and orientation of all objects in the workspace. With MJPC, Aloha is able to execute a wide variety of tasks, such as picking/placing objects, using both arms to re-orient objects, and handing over objects from one arm to the other.

To allow MJPC to successfully plan in the 14-DoF action space, we run the simulation at 25\% real-time speed. MJPC is able to find dynamic behaviors to solve tasks, such as rolling an apple on the table to move it closer to another object, or using one arm as leverage to flip a bowl upside down with the other arm. These policies are only tested in simulation and are not tuned for transfer to real.

\noindent\textbf{Bi-arm Kuka.}
The Kuka bi-arm robot is comprised by two 7-DoF Kuka LBR IIWA14 arms fixed to the ground. For the scope of this work, no end-effector was attached to them. Arrow indicators along with the words "left" and "right" were added to facilitate data collection with regards to natural language to orientation mappnings in the scene. For this embodiment, we have developed two distinct scenes: 

\begin{enumerate}
    \item[a)] \textbf{Single large cube}: This scene contains a single large cube in the center of the arms' workspace. 
    \item[b)] \textbf{Particle manipulation}: This scene contains 5 small cubes (particles) of various colors - red, green, blue, yellow, purple - initialized in random positions within the workspace of the arms. More specifically, their x and y coordinates at initialization are each sampled randomly from a uniform distribution between $(-0.5, 0.5)$.
\end{enumerate}

Finally both scenes contain four separate goal points, two in the air (blue goal and red gaol) and two on the ground (green goal and purple goal) that are stationary and can be used as target positions for moving objects.

The MJPC implementation for this embodiment includes sensors for the position and orientation of each arm, each goal and each movable object. It also includes an API that allows obtaining and setting the poses of all objects. For this platform, MJPC is leveraged to perform singular tasks that involve moving the large cube towards goal positions or in relative locations, pick up cubes, sweep cubes, as well as sequential tasks such as moving one block towards another followed a subsequent move towards a third block or goal position.

A current limitation of this embodiment is that it is only evaluated in simulation. The setup as well as the arm control are not yet realistic and would hinder any sim2real transfer. In addition, for all results presented in this work, bi-arm Kuka was considered an unseen embodiment used only for testing and not training. This causes some domain shift in terms of produced code (e.g. minor code mistakes such as APIs from other embodiments can be generated on occasion) which can increase the number of chat UI interactions. 

\noindent\textbf{Kuka+Hand.}
This embodiment comprises a 7-DoF Kuka LBR IIWA14 arm attached with a custom hand with three fingers (4-DoF each).
The arm is fixed within a basket containing four objects: red and green blocks, and a connector that can be inserted into a plug base.

The simulation and MJPC implementation for this embodiment includes sensors providing the positions of all finger and arm joints and pose/orientation of the hand and all objects in the scene.

With MJPC, the Kuka+Hand can execute a number of interesting behaviors, such as dexterously using the fingers to rearrange objects. Since there are no constraints imposed on maintaining contact with the objects, we observe that the fingers can sometimes leverage dynamic manipulations \eg flicking objects from one part of the workspace to another, or juggling to re-orient object mid-air before grasping them to place them down. The caveat of course, is that these behaviors are optimized in simulation and require object pose information during predictive control rollouts (which may struggle to transfer to the real world via sim2real distillation).

In terms of limitations, the predictive control search space for dexterous manipulation with all $7+4\times 3=19$ degrees of freedom is large and can be challenging to do sampling-based control over. Thus the simulator runs at only 15\% of real-time speeds (to allow for compute-bound MJPC with 128 cores) to discover manipulation solutions. Chat turn durations (shown in \cref{tab:durations_embodiment}) suggest that the Kuka+Hand platform takes the longest time for users to teach -- each chat turn takes on average 1.5 minutes, while chat sessions around 7 minutes, much of the time is spent watching the robot ``figure out'' online how to do the task.

\subsection{Tasks}
\label{app:tasks}

\noindent\textbf{Robot Dog.}
We design 19 tasks for the robot dog embodiment, among which 12 are used in training the LLM:

\definecolor{lightgray}{gray}{0.9}
\begin{table}[H]
\small
\ContinuedFloat
\rowcolors{1}{white}{lightgray}
\small
\begin{center}
\renewcommand{\arraystretch}{1.2}
\begin{subfigure}[b]{0.5\textwidth}
         \centering
\begin{tabular}{|p{0.8\textwidth}|}
\hline
\textbf{Robot Dog Train Tasks} \\ \hline
Sit. \\
High-five with the front right paw. \\
Downward dog. \\
Walk to the left. \\
Walk forward. \\
Hop. \\
Turn around clockwise. \\
Walk backward. \\
Walk backward while turning to face right. \\
Walk forward while turning left. \\
Close the middle drawer. \\
Open the door by pushing it. \\
\hline
\end{tabular}
\end{subfigure}
\end{center}
\end{table}
and 7 are held out for testing the LLM performance:

\definecolor{lightgray}{gray}{0.9}
\begin{table}[H]
\small
\ContinuedFloat
\rowcolors{1}{white}{lightgray}
\small
\begin{center}
\renewcommand{\arraystretch}{1.2}
\begin{subfigure}[b]{0.5\textwidth}
         \centering
\begin{tabular}{|p{0.8\textwidth}|}
\hline
\textbf{Robot Dog Test Tasks} \\ \hline
   High-five with the front left paw. \\
   Walk to the right. \\
   Turn around counterclockwise. \\
   Hop while turning clockwise. \\
   Hop while turning counterclockwise. \\
   Close the bottom drawer. \\
   Close the door by pushing it. \\
\hline
\end{tabular}
\end{subfigure}
\end{center}
\end{table}

\noindent\textbf{Mobile Manipulator.}
We task the users to teach the mobile manipulator 14 tasks, among which 11 are used for training:

\definecolor{lightgray}{gray}{0.9}
\begin{table}[H]
\small
\ContinuedFloat
\rowcolors{1}{white}{lightgray}
\small
\begin{center}
\renewcommand{\arraystretch}{1.2}
\begin{subfigure}[b]{0.5\textwidth}
         \centering
\begin{tabular}{|p{0.8\textwidth}|}
\hline
\textbf{Mobile Manipulator Train Tasks} \\ \hline
  Grasp the apple. \\
   Knock over coke can. \\
   Lift the apple high. \\
   Place the apple next to the cube. \\
   Push the apple toward the cube. \\
  Move the cube further away from the robot. \\
   Move the cube a little bit to the left. \\
   Open the top drawer. \\
   Place the cube behind the apple. \\
   Flip the cube upside down. \\
   Place the apple on the cube. \\
\hline
\end{tabular}
\end{subfigure}
\end{center}
\end{table}
and 3 are held out for testing:

\definecolor{lightgray}{gray}{0.9}
\begin{table}[H]
\small
\ContinuedFloat
\rowcolors{1}{white}{lightgray}
\small
\begin{center}
\renewcommand{\arraystretch}{1.2}
\begin{subfigure}[b]{0.5\textwidth}
         \centering
\begin{tabular}{|p{0.8\textwidth}|}
\hline
\textbf{Mobile Manipulator Test Tasks} \\ \hline
   Pick up the cube. \\
   Place the apple in front of the cube. \\
   Upright the coke can. \\
\hline
\end{tabular}
\end{subfigure}
\end{center}
\end{table}

\noindent\textbf{Aloha.}
The robot was instructed by users to perform 16 tasks in total, with the following 10 used in training the LLM:

\definecolor{lightgray}{gray}{0.9}
\begin{table}[H]
\small
\ContinuedFloat
\rowcolors{1}{white}{lightgray}
\small
\begin{center}
\renewcommand{\arraystretch}{1.2}
\begin{subfigure}[b]{0.5\textwidth}
         \centering
\begin{tabular}{|p{0.8\textwidth}|}
\hline
\textbf{Aloha Train Tasks} \\ \hline
Grasp the apple and lift it up. \\
Grasp the coke can and lift it up. \\
Pick up the cube and lift it above the apple. \\
Pick up the box and lift it above the coke can. \\
Flip the box upside down. \\
Flip the apple upside down. \\
Flip the drink upside down. \\
Flip the apple upside down and move the apple to the center of the table. \\
Move the box and the apple close to each other. \\
Push the box and the bowl close to each other. \\
\hline
\end{tabular}
\end{subfigure}
\end{center}
\end{table}
and the following 6 held out for testing:

\definecolor{lightgray}{gray}{0.9}
\begin{table}[H]
\small
\ContinuedFloat
\rowcolors{1}{white}{lightgray}
\small
\begin{center}
\renewcommand{\arraystretch}{1.2}
\begin{subfigure}[b]{0.5\textwidth}
         \centering
\begin{tabular}{|p{0.8\textwidth}|}
\hline
\textbf{Aloha Test Tasks} \\ \hline
Grasp the box and lift it up. \\
Pick up the coke can and lift it above the apple. \\
Flip the bowl upside down. \\
Move the apple and the bowl closer to each other. \\
Move the foods closer to each other. \\
Flip the box upside down and move the box to the center of the table. \\
\hline
\end{tabular}
\end{subfigure}
\end{center}
\end{table}

\noindent\textbf{Bi-arm Kuka.}
\noindent The robot was instructed by users to perform 16 different tasks (test only) across the two available scenes:

\definecolor{lightgray}{gray}{0.9}
\begin{table}[H]
\small
\ContinuedFloat
\rowcolors{1}{white}{lightgray}
\small
\begin{center}
\renewcommand{\arraystretch}{1.2}
\begin{subfigure}[b]{0.5\textwidth}
         \centering
\begin{tabular}{|p{0.8\textwidth}|}
\hline
\textbf{Bi-arm Kuka Single Large Cube Scene Test Tasks} \\ \hline
 Pick up the cube and lift it up to the blue goal. \\
 Pick up the cube and lift it up by 20cm. \\
 Move the cube to the green goal on the floor. \\
 Move the cube 20cm to the right without rotating it. \\
 Pick up the cube and lift it up to the red goal. \\
 Move the cube 20cm to the left of the purple goal on the floor and rotate it 90 degrees. \\
\hline
\end{tabular}
\end{subfigure}
\end{center}
\end{table}

\definecolor{lightgray}{gray}{0.9}
\begin{table}[H]
\small
\ContinuedFloat
\rowcolors{1}{white}{lightgray}
\small
\begin{center}
\renewcommand{\arraystretch}{1.2}
\begin{subfigure}[b]{0.5\textwidth}
         \centering
\begin{tabular}{|p{0.8\textwidth}|}
\hline
\textbf{Bi-arm Kuka Particle Manipulation Scene Test Tasks} \\ \hline
     Move the blue cube to the green goal on the floor. \\
     Move the green cube 20cm to the right. \\
     Move the red cube to the green goal, then to the purple goal. \\
     Sweep the red cube and the blue cube towards the green goal.\\
     Sweep all the cubes to the purple goal.\\
     Bring the red cube 20cm to the left of the green goal.\\
     Move the blue cube 20cm in front of the green cube.\\
     Sweep the yellow cube to the blue cube, then to the red cube.\\
     Move the purple cube to the yellow cube, then to the green cube, then to the blue cube.\\
     Move the purple cube 10cm to the right of the yellow cube, then 20cm behind the blue cube.\\
\hline
\end{tabular}
\end{subfigure}
\end{center}
\end{table}

\textbf{Kuka+Hand.}
\noindent The robot was instructed by users to perform 18 different tasks (test only):
\definecolor{lightgray}{gray}{0.9}
\begin{table}[H]
\small
\ContinuedFloat
\rowcolors{1}{white}{lightgray}
\small
\begin{center}
\renewcommand{\arraystretch}{1.2}
\begin{subfigure}[b]{0.5\textwidth}
         \centering
\begin{tabular}{|p{0.8\textwidth}|}
\hline
\textbf{Kuka+Hand Test Tasks} \\ \hline
     Move the gripper to reach the red block. \\
     Lift the connector in the air. \\
     Grasp the green object, hold it for a while in the air, and then drop it. \\
     Insert the connector into the socket. \\
     Stack the red block on the green block. \\
     Move the green cube to the far right corner. \\
     Move the red thing and the plug base to the far left corner. \\
     Stack the red block on the the base. \\
     Move the four objects into different corners. \\
     Move all objects into near left corner. \\
     Insert the plug into the base and stack the red cube on the green cube. \\
     Insert the connector into the socket, then put the green cube on the connector. \\
     Lift both cubes in the air. \\
     Disconnect the connector from the base. \\
     Separate the red block from the green block. \\
     Separate the red block away from the other objects. \\
     Move all objects into near left corner. \\
     Move the four objects into different corners. \\
\hline
\end{tabular}
\end{subfigure}
\end{center}
\end{table}
\subsection{Prompts}
\label{app:prompts}

\begin{figure*}[h]
    \centering
    \includegraphics[width=0.7\linewidth]{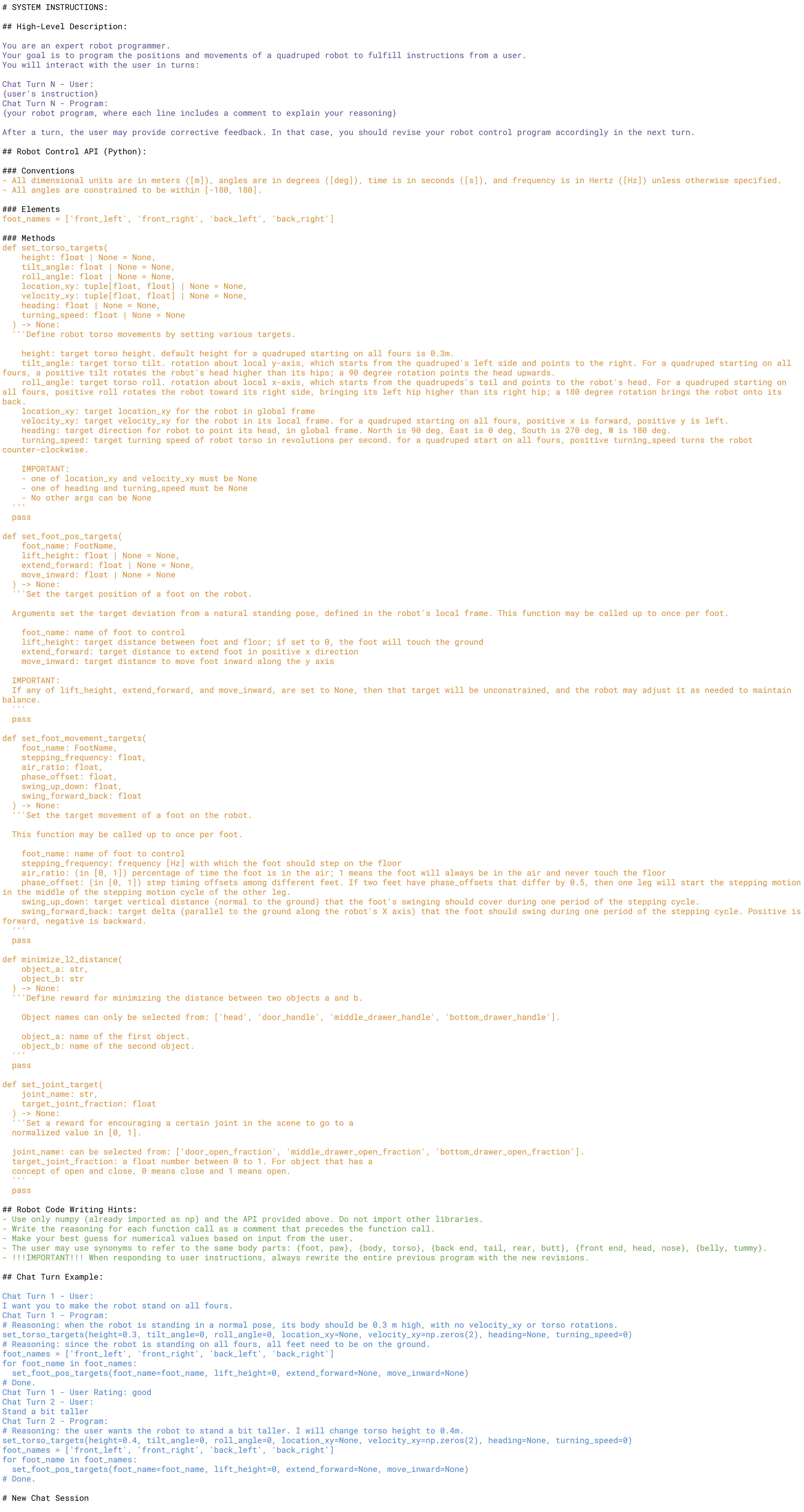}
    \caption{Robot dog prompt consists of a high level description of the goals and format (\textcolor{prompt-purple}{purple}), robot reward code API (\textcolor{prompt-orange}{orange}), code-writing hints (\textcolor{prompt-green}{green}), and chat turn examples (\textcolor{prompt-blue}{blue}).}
    \label{fig:prompt_adog}
\end{figure*}

\begin{figure*}[h]
    \centering
    \includegraphics[width=0.7\linewidth]{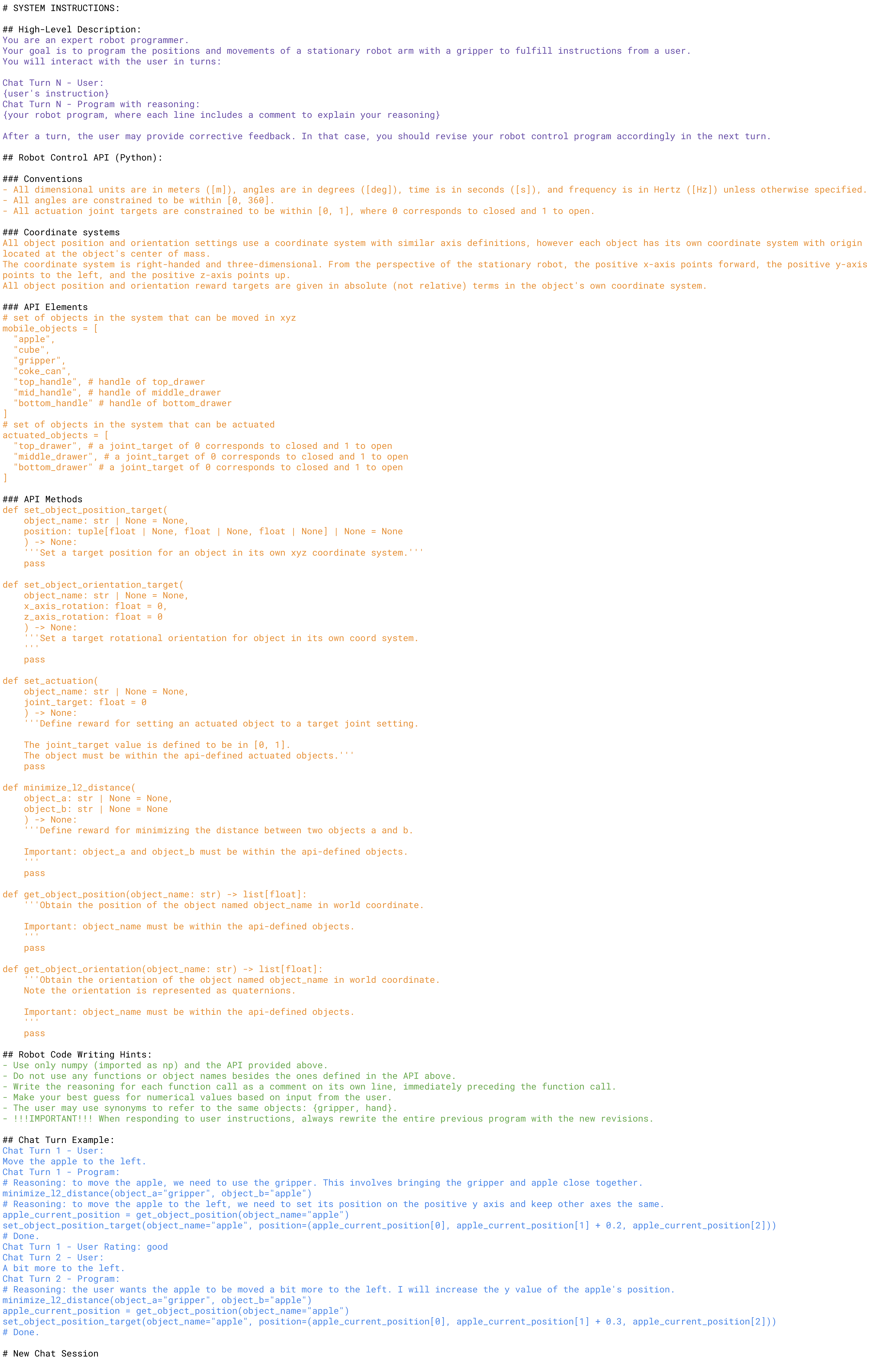}
    \caption{Mobile manipulator prompt consists of a high level description of the goals and format (\textcolor{prompt-purple}{purple}), robot reward code API (\textcolor{prompt-orange}{orange}), code-writing hints (\textcolor{prompt-green}{green}), and chat turn examples (\textcolor{prompt-blue}{blue}).}
    \label{fig:prompt_meta}
\end{figure*}

\begin{figure*}[h]
    \centering
    \includegraphics[width=0.7\linewidth]{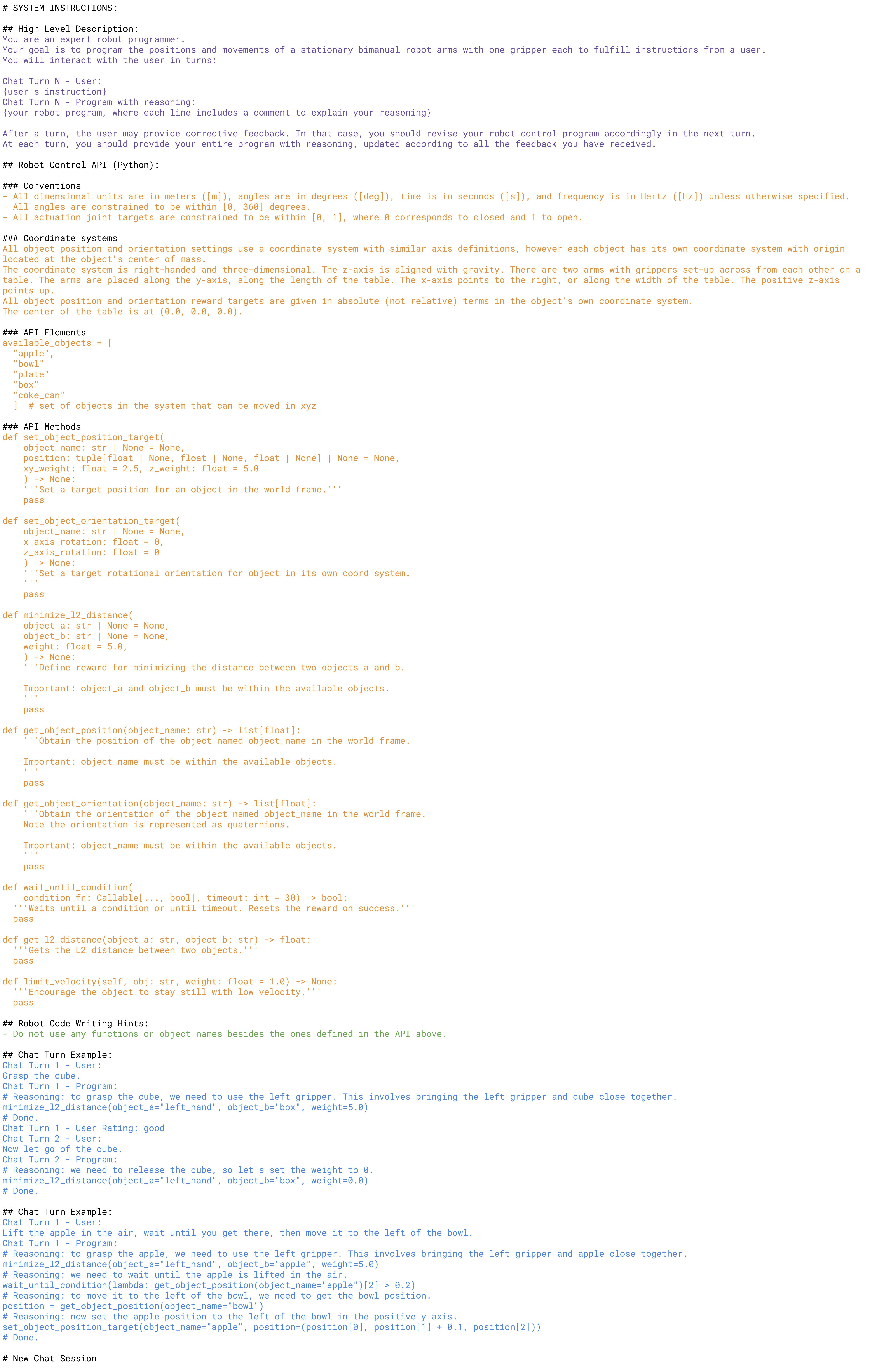}
    \caption{ALOHA prompt consists of a high level description of the goals and format (\textcolor{prompt-purple}{purple}), robot reward code API (\textcolor{prompt-orange}{orange}), code-writing hints (\textcolor{prompt-green}{green}), and chat turn examples (\textcolor{prompt-blue}{blue}).}
    \label{fig:prompt_aloha}
\end{figure*}

\begin{figure*}[h]
    \centering
    \includegraphics[width=0.7\linewidth]{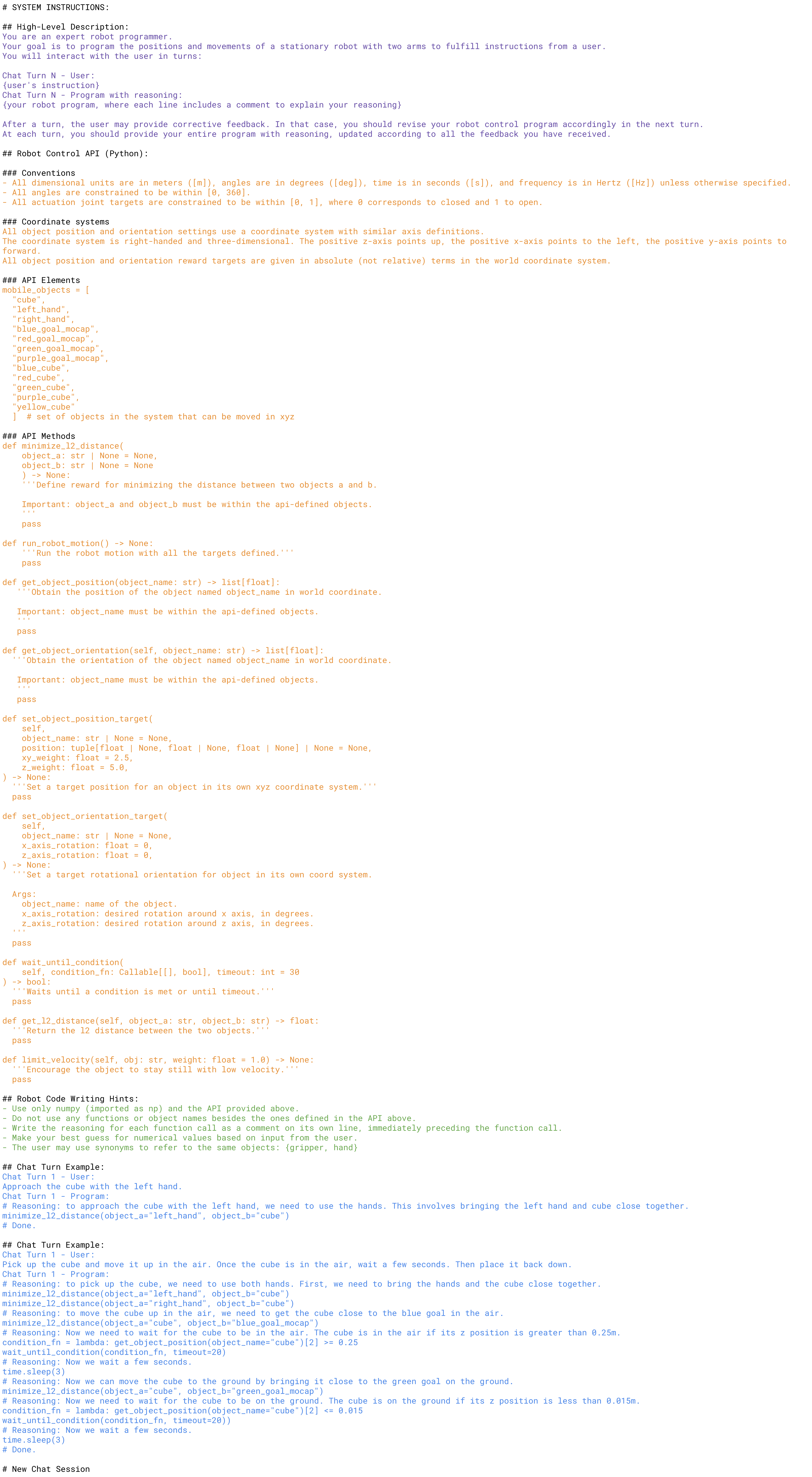}
    \caption{Bi-arm Kuka prompt consists of a high level description of the goals and format (\textcolor{prompt-purple}{purple}), robot reward code API (\textcolor{prompt-orange}{orange}), code-writing hints (\textcolor{prompt-green}{green}), and chat turn examples (\textcolor{prompt-blue}{blue}).}
    \label{fig:prompt_biarm_kuka}
\end{figure*}

\begin{figure*}[h]
    \centering
    \includegraphics[width=0.7\linewidth]{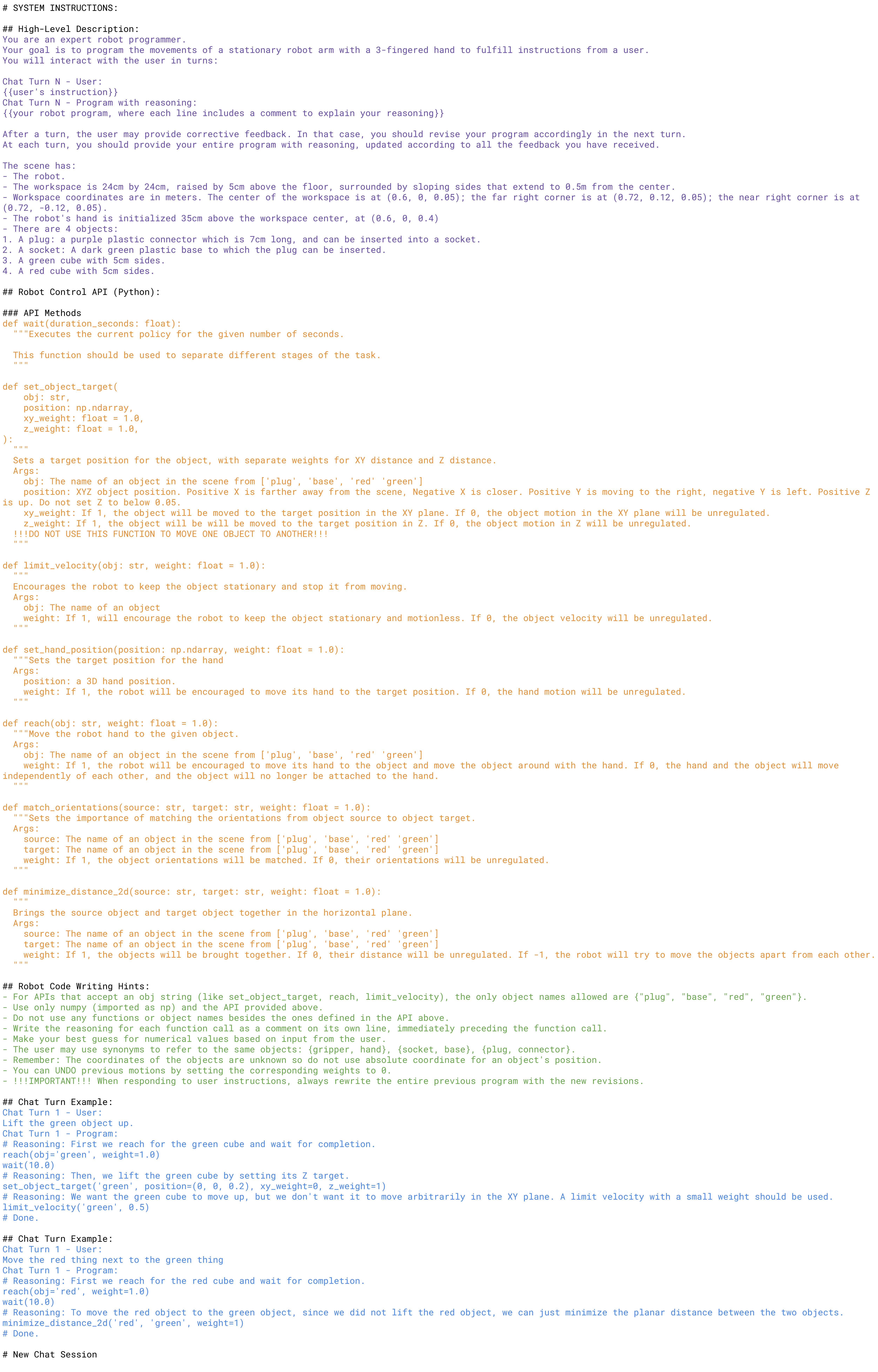}
    \caption{Kuka+Hand prompt consists of a high level description of the goals and format (\textcolor{prompt-purple}{purple}), robot reward code API (\textcolor{prompt-orange}{orange}), code-writing hints (\textcolor{prompt-green}{green}), and chat turn examples (\textcolor{prompt-blue}{blue}).}
    \label{fig:prompt_kuka_hand}
\end{figure*}

Prompts for each of the embodiments are shown in \cref{fig:prompt_adog}, \cref{fig:prompt_meta}, \cref{fig:prompt_aloha}, \cref{fig:prompt_biarm_kuka}, and \cref{fig:prompt_kuka_hand} respectively. The LLMs are trained to complete session data  with the input prompts prepended for each robot embodiment.

\end{document}